\documentclass[lettersize,journal]{IEEEtran}
\usepackage{amsmath,amsfonts}
\usepackage{algorithmic}
\usepackage{algorithm}
\usepackage{array}
\usepackage{textcomp}
\usepackage{stfloats}
\usepackage{url}
\usepackage{verbatim}
\usepackage{graphicx}
\usepackage{cite}
\usepackage{multirow}
\usepackage{multicol}
\usepackage{rotating}
\usepackage{wrapfig}
\usepackage{tabularray}
\usepackage{float}
\usepackage{xcolor}
\usepackage{caption}
\usepackage{subcaption}
\usepackage{xcolor}
\usepackage{colortbl}
\usepackage[super]{nth}
\definecolor{mygray}{gray}{.92}
\definecolor{customviolet}{RGB}{148, 0, 211}
\definecolor{cvprblue}{rgb}{0.21,0.49,0.74}
\definecolor{Green}{rgb}{0.85882353, 0.90980392, 0.84705882}

\begin{document}

\title{Implicit to Explicit Entropy Regularization: Benchmarking ViT Fine-tuning under Noisy Labels}
%\author{Anonymous Submission
\author{Maria Marrium, Arif Mahmood, Mohammed Bennamoun
% ~\IEEEmembership{Member,~IEEE,}
        % <-this % stops a space
\thanks{M. Marrium and A. Mahmood are with the Center for Artificial Intelligence and Robot Vision, Information Technology University, Lahore, Pakistan. Corresponding author email: arif.mahmood@itu.edu.pk}% <-this % stops a space
\thanks{M. Bennamoun is with the University of the Western Australia.  }% <-this % stops a space
%\thanks{Manuscript received August 21, 2024; revised Dec 21, 2024.}
}

\markboth{}%
{Marrium \MakeLowercase{\textit{et al.}}: 
Benchmarking ViT Fine-tuning under Noisy Labels}

% \IEEEpubidadjcol
% Remember, if you use this you must call  

\maketitle

\begin{abstract}
Automatic annotation of large-scale datasets can introduce noisy training data labels, which adversely affect the learning process of deep neural networks (DNNs). Consequently, Noisy Labels Learning (NLL) has become a critical research field for Convolutional Neural Networks (CNNs), though it remains less explored for Vision Transformers (ViTs). In this study, we evaluate the vulnerability of ViT fine-tuning to noisy labels and compare its robustness with CNNs. We also investigate whether NLL methods developed for CNNs are equally effective for ViTs. Using linear probing and MLP-K fine-tuning, we benchmark two ViT backbones (ViT-B/16 and ViT-L/16) using three commonly used classification losses: Cross Entropy (CE), Focal Loss (FL), and Mean Absolute Error (MAE), alongside six robust NLL methods: GCE, SCE, NLNL, APL, NCE+AGCE, and ANL-CE. The evaluation is conducted across six datasets including MNIST, CIFAR-10/100, WebVision, Clothing1M, and Food-101N. Furthermore, we explore whether implicit prediction entropy minimization contributes to ViT robustness against noisy labels, noting a general trend of prediction entropy reduction across most NLL methods. Building on this observation, we examine whether explicit entropy minimization could enhance ViT resilience to noisy labels. Our findings indicate that incorporating entropy regularization enhances the performance of established loss functions such as CE and FL, as well as the robustness of the six studied NLL methods across both ViT backbones.
\end{abstract}

\begin{IEEEkeywords}
Vision Transformers (ViTs); Noisy Label Learning (NLL); Fine-tuning
Performance; Entropy Regularization; Robust Classification Methods.\end{IEEEkeywords}

\section{Introduction}\label{sec:intro}
\IEEEPARstart{D}{eep} Neural Networks (DNNs) have transformed a variety of machine learning tasks, driven by the availability of large, high-quality annotated datasets \cite{deng2009imagenet, russakovsky2015imagenet, lin2014microsoft, shao2019objects365}. Large-scale datasets can be collected from the web via search engines or social media \cite{algan2021image}. 
Acquiring and manually annotating these datasets is both costly and time-intensive. To mitigate this, cheaper alternatives have been developed. 
One method involves crowdsourcing the labeling process through platforms like Amazon Mechanical Turk and Crowdflower, significantly reducing labeling costs. Another method employs automated systems for labeling data using deep learning techniques \cite{wang2016cnn, hu2016learning},  retrieval-based methods \cite{verma2012image, zhang2003automated}, and graph-based semi-supervised learning methods \cite{huang2012multi, wang2011image}. However, these approaches often lead to the introduction of noisy labels, which can adversely affect the learning outcomes of DNNs \cite{liang2022few, xiao2015learning}. 
Moreover, label noise can also stem from human annotators who may lack the necessary experience, or from data that is too complex to be accurately labeled even by experts \cite{7885126, wei2022deep}. This widespread issue underscores the need for developing robust algorithms capable of managing noisy labels effectively \cite{song2022learning, liu2022convergence}.

\begin{figure}[t!]
    \centering
\includegraphics[width=0.49\textwidth]{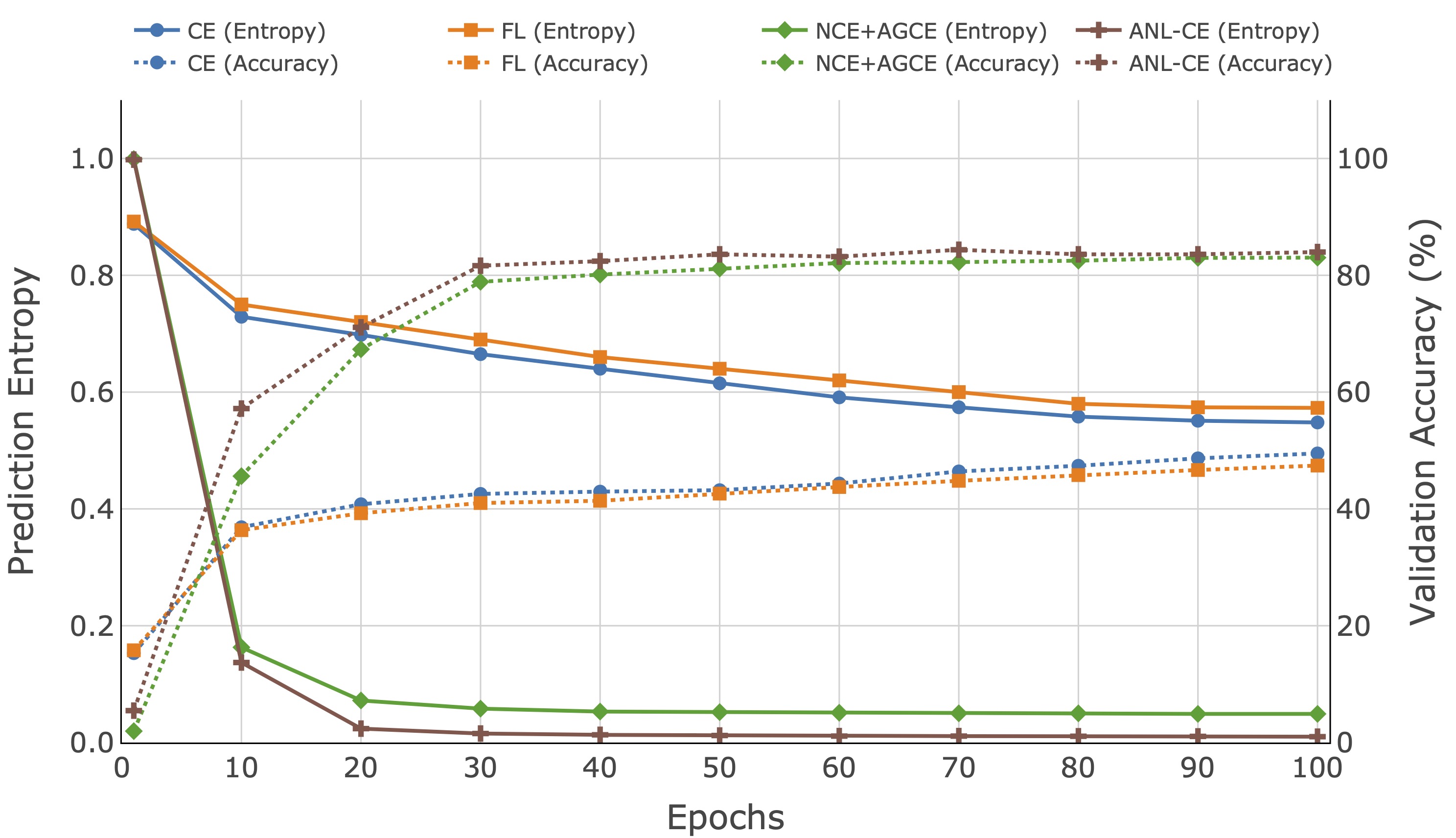}
\caption{
\textbf{Prediction Entropy and Validation Accuracy Trends during Training on Noisy CIFAR-100 Data using ViT-B/16 with MLP-3 Fine-Tuning.} Illustration of the changes in prediction entropy and the corresponding validation accuracy over 100 epochs for various classification loss functions: Cross-Entropy (CE), Focal Loss (FL) \cite{lin2017focal}, NCE+AGCE \cite{zhou2021asymmetric}, and ANL-CE \cite{ye2023active}. The graph shows that as prediction entropy decreases, there is a marked improvement in validation accuracy, indicating effective learning and adaptation to noisy data conditions. }
\label{fig:ent_reduction_acc_comp}
\end{figure}
Large-scale real-world datasets inevitably contain a significant portion of mislabeled training samples. Previous research has shown that these samples can disrupt the learning process of DNNs, impairing their effectiveness \cite{zhang2017understanding, xiao2015learning, liu2022video}. Consequently, developing strategies to learn in the presence of noisy labels has become a crucial area of research. Existing research on robust noisy label learning (NLL) can be categorized into four main approaches: \textbf{1)} Label correction methods aimed at detecting and correcting incorrect labels \cite{xiao2015learning, vahdat2017toward, li2017learning, zhang2017improving, chen2024learning}. \textbf{2)} Loss correction methods that adjust the loss function based on an estimated noise transition matrix \cite{patrini2017making, reed2014training, han2018masking,sukhbaatar2014training}. \textbf{3)} Refined training strategies designed to better accommodate incorrect labels \cite{wang2018iterative, tanaka2018joint, ma2018dimensionality,jiang2018mentornet, han2018co, kim2019nlnl, 10496838}. \textbf{4)} Robust loss functions inherently designed to withstand the impact of noisy labels \cite{ghosh2017robust, ma2020normalized, wang2019symmetric, zhang2018generalized, ye2023active}. The first three categories often suffer from inaccurate noise estimations and involve complex training procedures, whereas robust loss functions offer a simpler and more effective solution. 

\begin{figure*}[!t]
\centering
\includegraphics[width=0.99\textwidth]{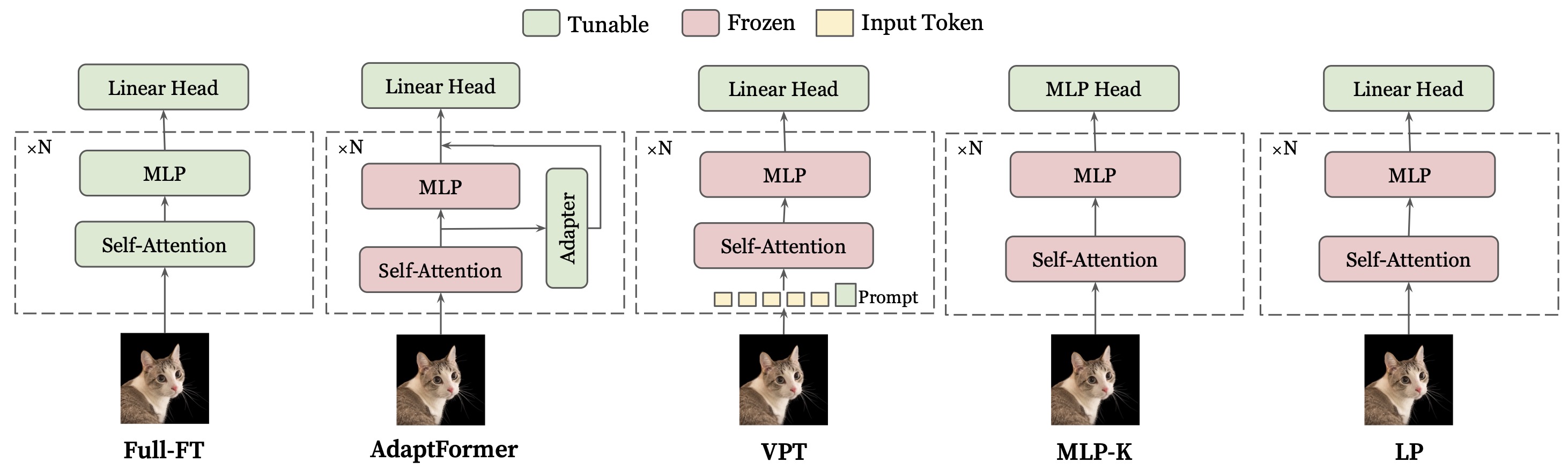}
\caption{ \textbf{Comparative Diagram of Five Fine-Tuning Techniques for Vision Transformers.} Details of the architectural modifications in ViTs when employing different fine-tuning strategies: Full Fine-Tuning, AdaptFormer \cite{chen2022adaptformer}, Visual Prompt Tuning \cite{jia2022visual}, MLP-K, and Linear Probing \cite{he2020momentum}. Each diagram shows which components of the architecture are tunable (green) versus frozen (pink) during the fine-tuning process. Specific elements such as input tokens and prompts are indicated, providing insights into how each technique modifies the standard ViT architecture to adapt to training constraints and objectives.}
\label{fig:finetune_tech}
\end{figure*}

Given the success of Vision Transformers (ViTs) \cite{dosovitskiy2020image} across various computer vision tasks \cite{kirillov2023segment, radford2021learning, yuan2021transanomaly, sharir2021image}, ViTs have established themselves as the de facto standard in the field. ViTs, pre-trained on massive datasets like ImageNet-21K, are typically employed for downstream tasks using fine-tuning methods rather than training from scratch \cite{he2022masked, kirillov2023segany}. Fine-tuning is an effective solution for overcoming challenges associated with limited training data and scarce computational resources. Recently, many fine-tuning techniques have been proposed as a trade-off between computational cost and performance \cite{zhou2022learning, touvron2022three,  chen2022adaptformer, zhang2022neural, lian2022scaling}. This work focuses on the most commonly used techniques including full fine-tuning (updating all parameters), AdaptFormer \cite{chen2022adaptformer}, VPT \cite{jia2022visual}, linear probing \cite{he2020momentum} (updating only the last linear layer), and MLP-K (updating only added K layers). While full fine-tuning often yields superior performance on clean datasets, it requires significant computational resources and is more sensitive to noisy labels, resulting in deteriorated performance compared to other fine-tuning methods (Section IV-A).

Existing research has extensively examined the robustness of Vision Transformers (ViTs) against adversarial and out-of-distribution data \cite{bai2021transformers, zhou2022understanding, paul2022vision}. However, the robustness of ViTs to noisy labels remains relatively unexplored \cite{liang2022few}. In this study, we first benchmark the robustness of ViTs to noisy label learning and propose a method to enhance this robustness. We evaluate the robustness of two ViT backbones, ViT-B/16 and ViT-L/16, using six datasets, which include three benchmark datasets (MNIST, CIFAR-10/100) and three real-world noisy datasets (WebVision, Clothing1M, and Food-101N). 
Initially, we apply six existing NLL methods, originally designed for CNNs \cite{wang2019symmetric, ma2020normalized, ye2023active, zhang2018generalized,kim2019nlnl,zhou2021asymmetric}, to both ViT backbones to test their effectiveness. For fair comparisons, we also employ standard classification losses such as Cross-Entropy (CE), Focal Loss (FL) \cite{lin2017focal}, and Mean Absolute Error (MAE). Our comprehensive benchmarking reveals that ViTs are generally less sensitive to noisy labels compared to CNNs, though their performance still declines as noise levels increase. Existing NLL methods do enhance the performance of ViTs in the presence of noisy labels; however, there remains a significant performance gap between clean and noisy data scenarios. This gap underscores the need for further development of more robust NLL methods for ViTs. Our detailed analysis indicates that NLL methods may improve performance by implicitly minimizing prediction entropy (see Fig. 1). Building on this insight, we propose the incorporation of explicit prediction entropy minimization through regularization. Extensive experimentation shows that this entropy regularization notably enhances the performance of nearly all NLL methods included in this study.

\textbf{Research Contributions:} This work aims to address the following research questions:

\begin{itemize}
    
%\item \textit{RQ1: Is ViT fine-tuning vulnerable to noisy labels}? The evaluation of various ViT fine-tuning techniques reveals significant performance reduction when fine-tuned with noisy labels, compared to their performance on clean data.

\item \textit{RQ1: How vulnerable is ViT fine-tuning to noisy labels?} We evaluate various ViT fine-tuning techniques to determine their performance stability under noisy conditions.

%\item \textit{RQ2: Is ViT fine-tuning more robust to noisy labels compared to CNNs}? Our experiments reveal that ViT fine-tuning is more robust to the label noise compared to CNNs across all datasets.

\item \textit{RQ2: How does ViT fine-tuning compare in robustness to noisy labels relative to CNNs?} We compare the robustness of ViT and CNN models to understand differences in handling label noise.

%\item \textit{RQ3: Do existing NLL methods for CNNs are effective for ViTs fine-tuning}? We observed that existing NLL methods are effective and improve the robustness of ViTs.

\item \textit{RQ3: Are existing NLL methods developed for CNNs also effective when applied to ViT fine-tuning?} We assess the transferability of established NLL methods from CNNs to ViTs.

%\item \textit{RQ4: Does implicit prediction entropy minimization relate to ViT robustness to noisy labels}? We analyze the results of existing NLL methods and observe an implicit entropy minimization trend.

\item \textit{RQ4: Is there a relationship between implicit prediction entropy minimization and ViT robustness to noisy labels?} We analyze existing NLL methods to explore potential relationships.

%\item \textit{RQ5: Does explicit entropy regularization improves ViT robustness to noisy labels}? We experimented adding entropy regularization with existing NLL methods and achieved performance improvement in majority cases.

\item \textit{RQ5: Can explicit entropy regularization enhance the robustness of ViTs to noisy labels?} We experiment with adding entropy regularization to NLL methods to examine its impact on performance.

\end{itemize}

\begin{figure*}[!t]
\centering
%\subfloat[]
\begin{subfigure}{0.3\linewidth}\includegraphics[width=\linewidth]{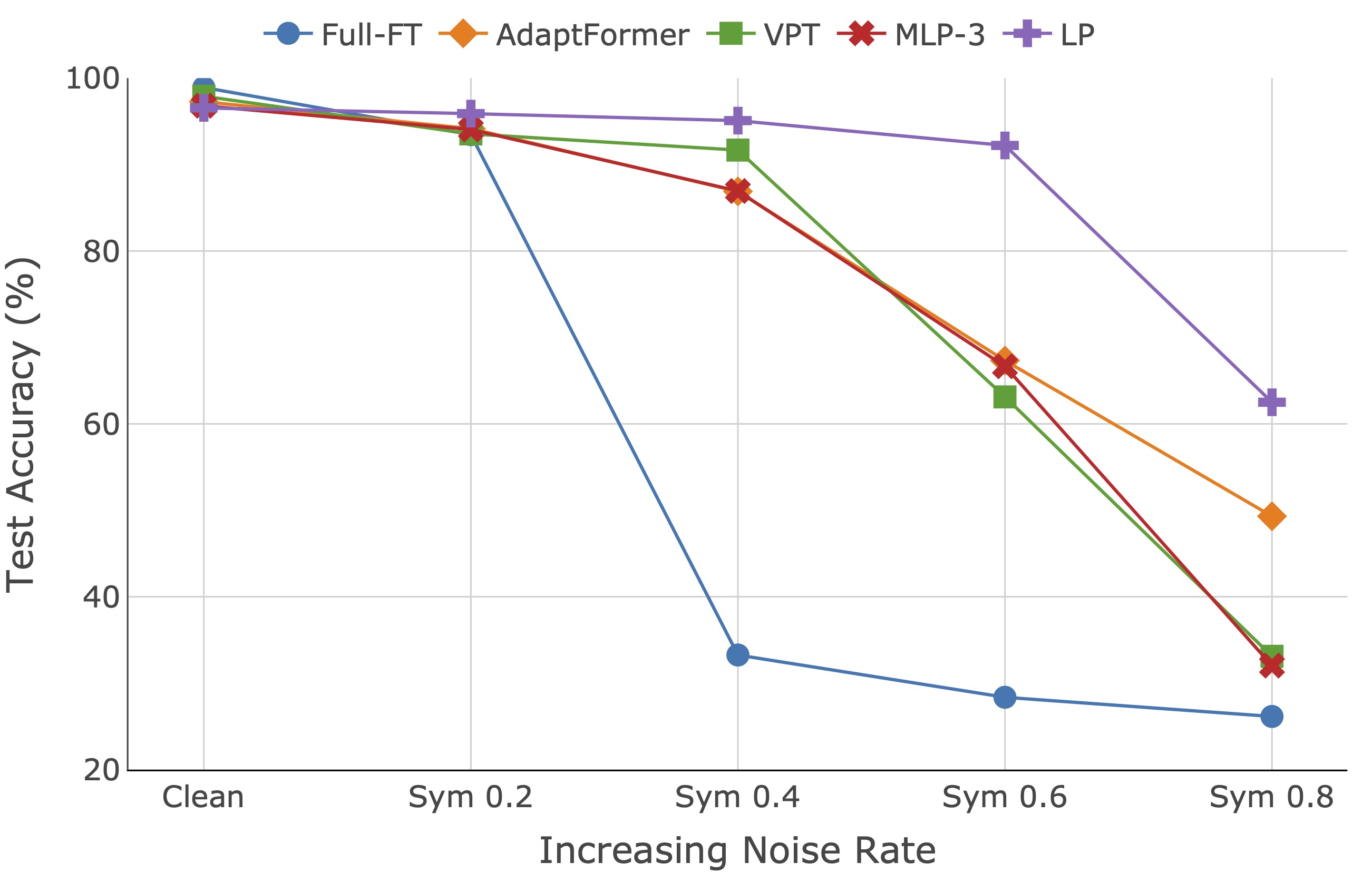}
\caption{CIFAR-10 + Symmetric Noise}
\end{subfigure}
\hfil
%\subfloat[]
\begin{subfigure}{0.3\linewidth}\includegraphics[width=\linewidth]{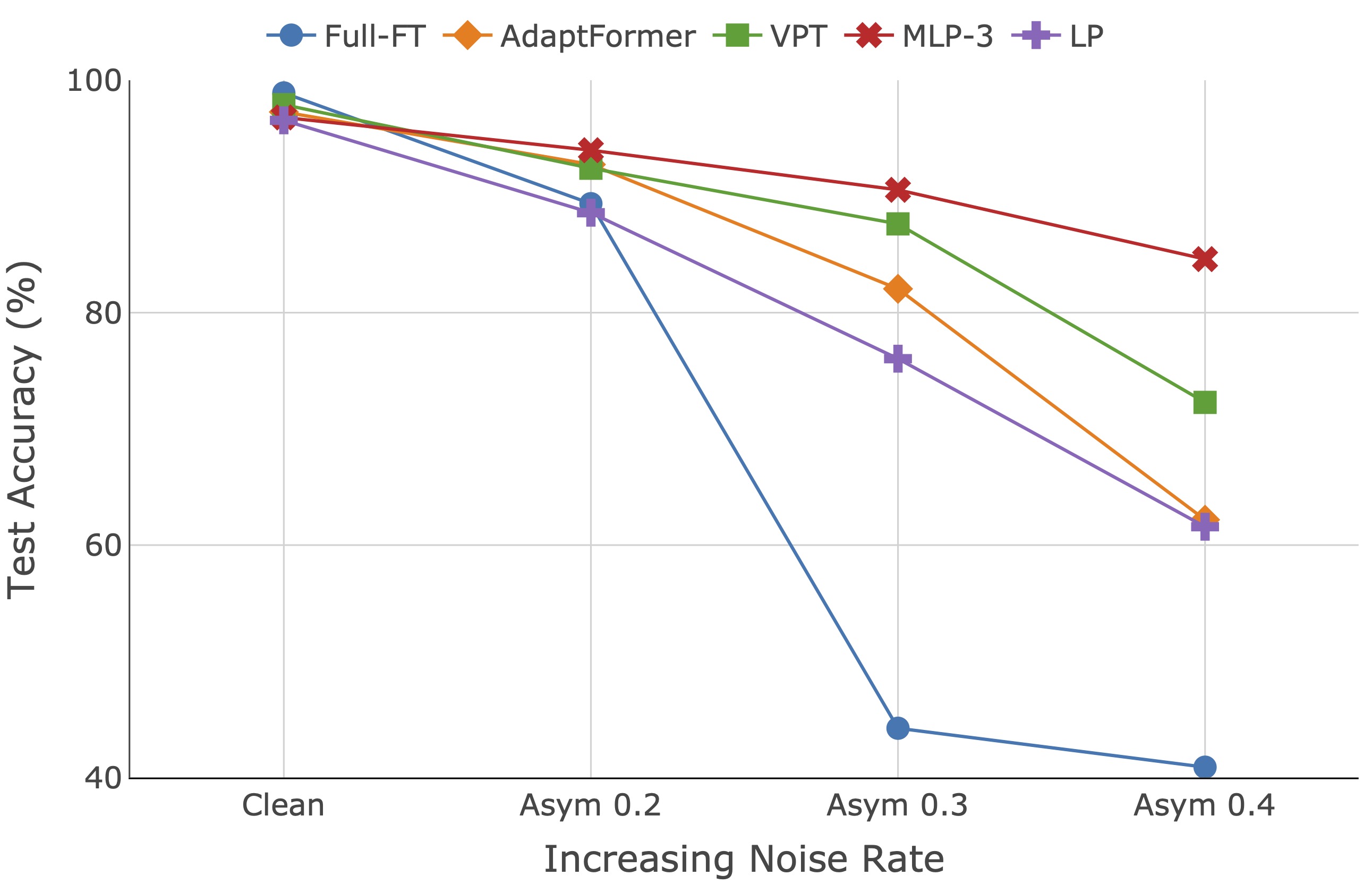} 
\caption{CIFAR-10 + Asymmetric Noise}\end{subfigure}
\hfil
%\subfloat[]
\begin{subfigure}{0.3\linewidth}\includegraphics[width=\linewidth]{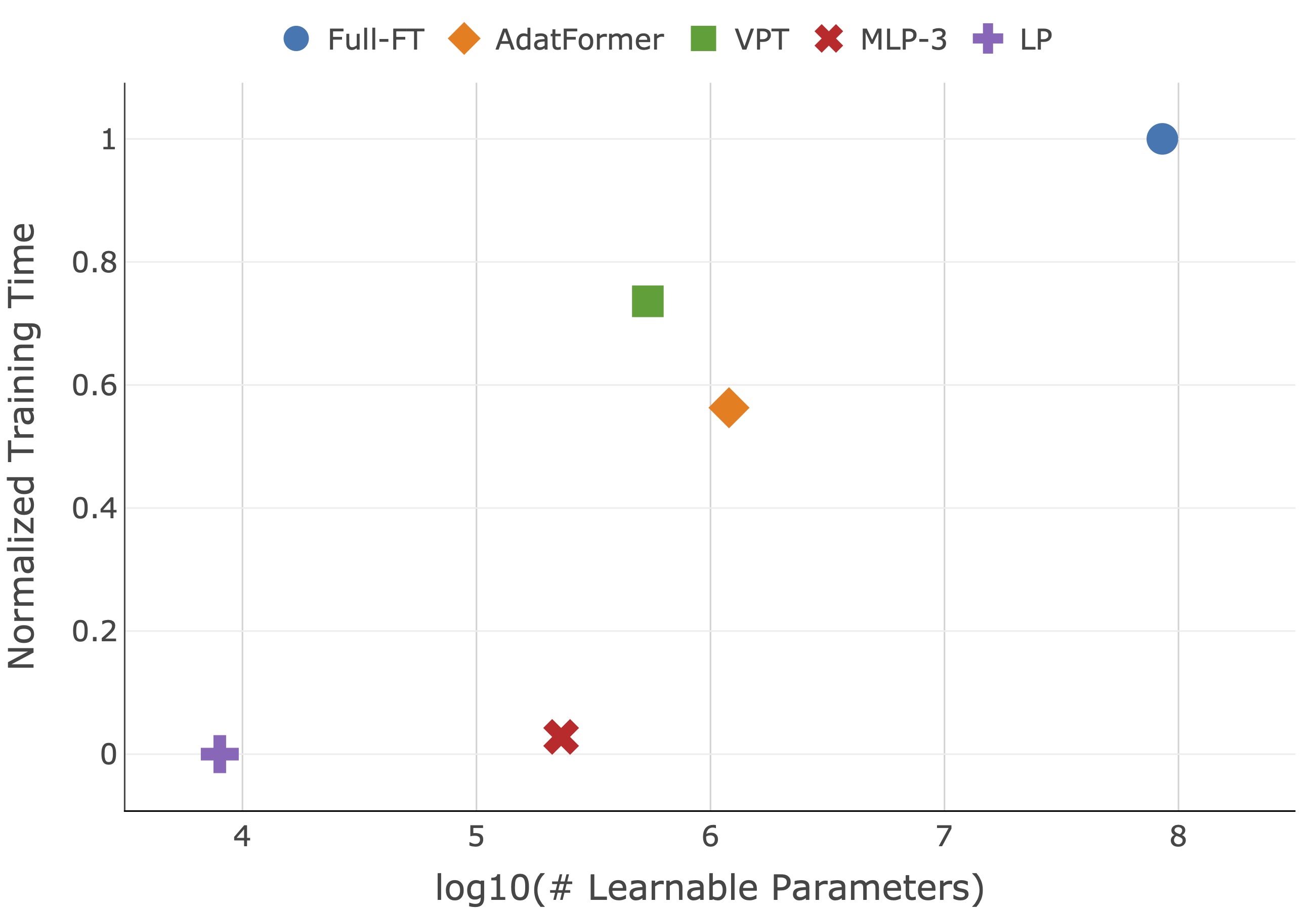}
\caption{Computational Overhead}
\end{subfigure}
\caption{\textbf{Impact of Noise Rates on Test Accuracy and Computational Overhead for Various Fine-Tuning Techniques.} \textbf{(a)} illustrates the test accuracy of five fine-tuning methods—Full Fine-Tuning, AdaptFormer \cite{chen2022adaptformer}, Visual Prompt Tuning \cite{jia2022visual}, MLP-3, and Linear Probing \cite{he2020momentum}-on the CIFAR-10 dataset under increasing symmetric noise rates from 0.2 to 0.8. \textbf{(b)} similarly depicts test accuracy as asymmetric noise levels increase from 0.2 to 0.4, demonstrating how each method copes with noise imbalance. \textbf{(c)} compares the computational overhead by showing the training time and the number of learnable parameters across these fine-tuning techniques, highlighting differences in computational efficiency and resource demands.
%Noise Rate Vs. Test Accuracy comparison of five fine-tuning techniques including Full-FT, AdaptFormer , VPT , MLP-3, and Linear Probing (LP) on CIFAR-10 dataset (a)-(b) performance reduction by increasing symmetric and asymmetric noise rates. (c)  Comparison of training time and the number of learnable parameters across different fine-tuning techniques.
}\vspace{-2mm}
\label{fig:finetune_comp}
\end{figure*}

\section{Related Work}

\subsection{Deep Learning-based NLL Methods}
Deep learning methods for Noisy Label Learning (NLL) are typically divided into four distinct categories:

\noindent\textbf{Label Cleaning Methods:} These methods aim to identify and correct mislabeled data \cite{xiao2015learning, vahdat2017toward, li2017learning, veit2017learning, zheng2021meta,yi2019probabilistic}. Xiao \textit{et al.} \cite{xiao2015learning} employ dual networks to predict the noise type and the probability of label transition. Li \textit{et al.} \cite{li2017learning} average knowledge transfer from an expert model trained on a clean dataset to enhance a target model trained with noisy data.

\noindent\textbf{Loss Correction Methods:} This category involves adjusting the loss function based on an estimated noise transition matrix \cite{patrini2017making, reed2014training, han2018masking, sukhbaatar2014training}. Patrini \textit{et al.} \cite{patrini2017making} developed loss correction techniques that are independent of the application domain and network architecture. Another approach, called `Masking' \cite{han2018masking} uses human judgment to handle improbable label transitions effectively.

\noindent\textbf{Refined Training Strategies:} These strategies are developed to adapt the training process for better handling of noisy labels \cite{wang2018iterative,tanaka2018joint, ma2018dimensionality,jiang2018mentornet, han2018co, kim2019nlnl, ma2018dimensionality}. Wang \textit{et al.} \cite{wang2018iterative}specifically refine labels within a single training iteration by identifying and correcting mislabeled examples using a local outlier factor algorithm \cite{breunig2000lof}. 
Kim \textit{et al.} \cite{kim2019nlnl} have introduced a method known as Negative Learning for Noisy Labels (NLNL).  Negative learning means an input sample does not belong to a class; instead of conventional  Positive Learning (PL) where an input sample belongs to a class. NLNL does not provide wrong information to the model as frequently as PL and hence is more robust to noisy labels. 

\begin{table*}[]
    \caption{
    \textbf{Evaluation of Existing NLL Methods for ViT Fine-Tuning Across Six Datasets.} The average test accuracy for ViT-B/16 and ViT-L/16 models using Linear Probing (LP) and MLP-3 fine-tuning techniques. Performance metrics for Common Loss Functions (CLF) are averaged over Cross-Entropy (CE), Focal Loss (FL), and Mean Absolute Error (MAE), while Noisy Label Learning (NLL) methods encompass averages from GCE, SCE, NLNL, NCE+RCE, NCE+AGCE, and ANL-CE. Results are also averaged  across various levels of noise specifically on MNIST, CIFAR-10, and CIFAR-100 datasets while WebVision, Clothing1M, and Food-101N are real-world noisy labels datasets. For further details and breakdowns, refer to Tables \ref{tab:vitb_MLP-K_res}, \ref{tab:real_world_data_res}, and supplementary Tables IX, X, and XI.}
    
    \centering
    \begin{tabular}{l|l|cc|cc|cc|c|c|c}
    \hline
    \multirow{2}{4em}{Loss} & \multirow{2}{4em}{Variants} &  \multicolumn{2}{c|}{MNIST} &  \multicolumn{2}{c|}{CIFAR-10} & \multicolumn{2}{c|}{CIFAR-100} & WebVision & Clothing1M & Food-101N\\
    \cline{3-11}
        & & Clean &  Noisy & Clean & Noisy & Clean & Noisy & Noisy & Noisy & Noisy\\
        \hline
        \multirow{5}{4em}{CLF} &ViT-B (LP) & 92.57 & 82.74&96.25& 86.23 &77.31 & 58.87&87.79& 63.96 & 75.09\\
        &ViT-L (LP) &94.01 &86.02 & 96.20 & 86.68&80.79 & 57.93&86.71 & 63.86& 81.05\\
         &ViT-B (MLP) & 94.53	&84.25 &96.52 & 75.90 &68.85 & 45.90 &88.47 & 64.64 & 74.31\\

         &ViT-L (MLP) & 98.31 & 88.93 &95.70 & 76.37&75.60& 52.11&86.81& 65.03 & 81.34\\
        
        \rowcolor{mygray}& \textbf{Average} & 94.85&85.48  &96.17 & 81.30 & 75.64 & 53.70 & 87.45& 64.37&	77.95 \\
         \hline
         \multirow{5}{4em}{NLL}& ViT-B (LP) &91.84 &80.65 &96.26 & 92.48 &83.95 & 75.36 &88.41 & 62.67& 74.59\\
         
         &ViT-L (LP)& 94.61& 80.64&95.44 & 93.58&87.99& 79.57&89.0& 63.94& 80.52\\
         
        &ViT-B (MLP)& 94.66 & 84.00&96.24 & 90.69 &83.80 & 74.62& 86.25 & 64.43 & 73.55\\
        
         &ViT-L (MLP) & 95.95 & 89.46 &95.39 & 91.08&88.41& 79.40&87.79& 65.15 & 79.74\\
         
         \rowcolor{mygray}& \textbf{Average} & 	94.26& 83.69	&95.83 &	91.96	&86.04 &77.24	&87.86 &	64.05 &77.10 \\
         \hline
    \end{tabular}
    
    \label{tab:LF_vs_NLL_comp}
\end{table*}
 
\noindent\textbf{Robust Loss Functions:} These methods are specifically designed to mitigate the effects of  noisy labels \cite{ma2020normalized, ye2023active, wang2019symmetric,zhang2018generalized, zhou2021asymmetric,amid2019two, amid2019robust, lyu2019curriculum}. 
Generalized Cross Entropy (GCE) \cite{zhang2018generalized}, for example, merges the benefits of  Mean Absolute Error (MAE) and Cross-Entropy (CE). Symmetric Cross Entropy (SCE) \cite{wang2019symmetric} addresses noisy data by combining Reverse Cross Entropy (RCE) with CE, where RCE is defined as: $-\sum_{k=1}^{k_c}\mathbf{p}(k|\mathbf{x}_i) \log \mathbf{q}(k|\mathbf{x}_i)$. 
Zhou \textit{et al.} \cite{zhou2021asymmetric} proposed Asymmetric Generalized Cross Entropy (AGCE)  fulfilling the noise tolerance condition proposed by Ghosh \textit{et al.} \cite{ghosh2017robust}. 
Ma \textit{et al.} \cite{ma2020normalized} designed Active Passive Loss (APL), which integrates an active component that assigns high probability to the ground truth class and a passive component that diminishes the likelihood of high probabilities for other classes. One implementation of APL is NCE+RCE, which has proven effective in noisy conditions.
Expanding on this concept,  Ye \textit{et al.} \cite{ye2023active}, noting that existing passive loss functions are scaled versions of MAE, proposed a new class of passive loss functions called Normalized Negative Loss Functions (NNLFs). An example of NNLF is ANL-CE loss which combines NCE with negative normalized cross entropy (NNCE). 
  
%\vspace{-6mm}
\subsection{ViT Fine-tuning Techniques}
The development of large-scale deep learning models has led to a shift towards a pre-training and fine-tuning paradigm, prominently used in fields like computer vision \cite{dosovitskiy2020image,he2022masked}  and natural language processing \cite{raffel2020exploring, vaswani2017attention}. Recent works have used large ViT models \cite{dosovitskiy2020image} trained on extensive datasets such as ImageNet-21K \cite{deng2009imagenet}, which have shown significant performance improvements and exceptional generalizability. These models provide pre-trained weights that are versatile across various downstream tasks \cite{  radford2021learning,he2022masked}. As pre-trained models become more complex, the focus of research has shifted to devising efficient fine-tuning methods that optimize performance for specific tasks, resulting in several parameter-efficient fine-tuning strategies \cite{zhou2022learning, touvron2022three, mahabadi2021parameter, karimi2021compacter, zaken2021bitfit, zhang2022neural, lian2022scaling}. Full Fine-Tuning (Full-FT) involves adjusting all parameters of a pre-trained model for a downstream task, which consumes substantial computational resources. Alternatively, techniques like linear probing, where only the final linear layer is adjusted, or MLP-K, which only fine-tunes the MLP classification head, are computationally economical due to fewer tunable parameters. Visual Prompt Tuning (VPT) \cite{jia2022visual} and AdaptFormer \cite{chen2022adaptformer} introduced an adapter module for task-specific fine-tuning while keeping the core transformer structure largely unchanged. AdaptFormer \cite{chen2022adaptformer} introduced an adapter module for task specific fine-tuning while keeping the core transformer structure largely unchanged. Both VPT and AdaptFormer are computationally expensive and incur significantly larger memory overhead compared to LP/MLP-K based fine-tuning. A visual illustration of these techniques is shown in Figure \ref{fig:finetune_tech}.

\begin{table*}[b]
\centering
\caption{\textbf{Test Accuracy for ViT-B/16 Using MLP-3 Fine-Tuning Under Varying Noise Conditions.} Comparison of test accuracies for ViT-B/16 across MNIST, CIFAR-10, and CIFAR-100 datasets, employing different NLL methods and common loss functions (CLF) under both clean and noisy scenarios. Noise levels are evaluated from 0.2 to 0.8 symmetrically and 0.2 to 0.4 asymmetrically. Best and 2nd best performances are in BOLD and underlined, respectively. 
%Effectiveness of existing NLL methods for ViTs fine-tuning: detailed performance of ViT-B/16 backbone using MLP-3 fine-tuning in terms of test accuracy (\%) on three benchmarks. The \nth{1} and \nth{2} best results are in \textbf{BOLD} and \underline{underline}.
} 
\begin{tabular}{p{0.01cm}p{0.01cm}l|l|cccc|ccc}
\hline

\hline
 &&\multirow{2}{4em}{Method} & \multirow{2}{4em}{Clean} & \multicolumn{4}{c|}{Sym Noise Rate ($\eta$)} & \multicolumn{3}{c}{Asym Noise Rate ($\eta$)}\\
\cline{5-11}
&&&&  0.2 &0.4 &  0.6 & 0.8 & 0.2 & 0.3 & 0.4\\
\hline

\hline

\parbox[t]{0.01cm}{\multirow{11}{*}{\rotatebox[origin=c]{90}{MNIST}}}&\parbox[t]{0.01cm}{\multirow{3}{*}{\rotatebox[origin=c]{90}{CLF}}} &CE & \textbf{98.83$\pm$0.05} & \textbf{97.65$\pm$0.18} &\textbf{ 97.26$\pm$0.48} & 94.53$\pm$0.08 & \underline{91.80$\pm$1.30} &\textbf{97.65$\pm$0.21}	&\textbf{96.85$\pm$0.10}	&95.31$\pm$0.07 \\

&&MAE &87.11$\pm$0.98 &77.92$\pm$0.19 &76.04$\pm$0.66	&68.09$\pm$3.51	&26.80$\pm$3.80&67.57$\pm$0.46	&59.01$\pm$0.03	&57.03$\pm$0.45	\\

&&FL & \underline{97.65$\pm$0.03} & \textbf{97.65$\pm$0.32} & 96.09$\pm$0.80  & 95.09$\pm$0.64 & 88.67$\pm$0.98&\textbf{97.65$\pm$0.05}	&95.70$\pm$0.21	&94.92$\pm$0.26  \\
\cline{2-11}

&\parbox[t]{0.01cm}{\multirow{6}{*}{\rotatebox[origin=c]{90}{NLL}}}&GCE & 97.27$\pm$0.02 & \underline{96.87$\pm$0.55} & 96.87$\pm$0.31  & 94.92$\pm$0.88 & 64.06$\pm$1.48 &   96.09$\pm$0.48	&95.31$\pm$0.18	&91.79$\pm$0.76\\

&&SCE & 97.26$\pm$0.08& \underline{97.26$\pm$0.48} & 96.48$\pm$0.21 & \underline{95.87$\pm$0.08} & \textbf{94.53$\pm$1.56} & \underline{96.48$\pm$0.21}	&\underline{96.48$\pm$0.21}	& \textbf{96.09$\pm$0.15} \\

&&NLNL& 90.01$\pm$0.02 &88.87$\pm$0.03 &79.85$\pm$0.01 &45.88$\pm$1.52 & 30.85$\pm$0.55&85.94$\pm$0.23 &77.90$\pm$0.18 &45.58$\pm$0.05\\

&&NCE+RCE & 97.27$\pm$0.05 &96.87$\pm$0.18  & 96.48$\pm$0.36  & \textbf{96.09$\pm$1.28} & 73.43$\pm$1.95  & 96.09$\pm$0.13	&89.06$\pm$0.08	&80.08$\pm$0.02\\

&&NCE+AGCE & 88.90$\pm$0.17 &80.85$\pm$0.74 & 74.61$\pm$0.28  & 60.54$\pm$0.34 & 46.87$\pm$0.83& 67.18$\pm$1.05	&57.81$\pm$0.06	&57.42$\pm$0.28\\

&&ANL-CE & 92.58$\pm$0.84& 91.01$\pm$0.48 & 83.98$\pm$0.73  & 69.14$\pm$0.73 & 66.40$\pm$0.50 &  91.02$\pm$0.18	&85.54$\pm$0.63	&70.31$\pm$0.48\\

\hline

\hline

\parbox[t]{0.01cm}{\multirow{11}{*}{\rotatebox[origin=c]{90}{CIFAR-10}}}&\parbox[t]{0.01cm}{\multirow{3}{*}{\rotatebox[origin=c]{90}{CLF}}}&CE & \textbf{96.80$\pm$0.04}& 94.05$\pm$0.05& 86.94$\pm$0.31& 66.66$\pm$0.15&32.03$\pm$0.44 & 93.98$\pm$0.03& 90.57$\pm$0.18& 84.61$\pm$0.30 \\

&&MAE & 96.27$\pm$0.12& 95.70$\pm$0.09& 87.50$\pm$0.32& 75.82$\pm$0.04&36.42$\pm$1.02 & 67.71$\pm$0.26& 58.89$\pm$0.12& 58.72$\pm$0.09\\

&&FL & \underline{96.50$\pm$0.07}& 94.60$\pm$0.09& 88.64$\pm$0.32& 70.81$\pm$0.04&33.47$\pm$0.41 & 95.27$\pm$0.03& 93.39$\pm$0.12& 88.20$\pm$0.12\\
\cline{2-11}
&\parbox[t]{0.01cm}{\multirow{6}{*}{\rotatebox[origin=c]{90}{NLL}}}&GCE & 96.40$\pm$0.03& \textbf{96.27$\pm$0.03}& \textbf{96.16$\pm$0.04}& \textbf{95.63$\pm$0.04} & \textbf{92.70$\pm$0.06}& 94.15$\pm$0.01& 94.97$\pm$0.10& \underline{88.89$\pm$0.42}\\

&&SCE & 96.36$\pm$0.04& 96.01$\pm$0.04& 94.98$\pm$0.02& 89.58$\pm$0.22&48.88$\pm$1.03 & 95.48$\pm$0.09& 92.40$\pm$0.20& 84.58$\pm$0.17\\

&&NLNL& 95.42$\pm$0.06 & 90.18$\pm$0.01 & 85.32$\pm$0.02 & 20.03$\pm$0.03 &10.00$\pm$0.01&86.37$\pm$0.17&82.05$\pm$0.01&78.07$\pm$0.07\\

&&NCE+RCE & 96.28$\pm$0.05& \underline{96.24$\pm$0.07}& \underline{95.96$\pm$0.05}& \underline{95.12$\pm$0.13}&\underline{89.66$\pm$0.07}& \underline{96.20$\pm$0.10}& \underline{95.66$\pm$0.07}& 75.19$\pm$0.59 \\

&&NCE+AGCE &96.31$\pm$0.03& 96.08$\pm$0.02& 95.81$\pm$0.08& 94.53$\pm$0.07&88.90$\pm$0.58 & 94.53$\pm$0.12& 84.37$\pm$0.09& 67.57$\pm$1.06\\

&&ANL-CE & 95.83$\pm$0.18& 95.70$\pm$0.32& 94.92$\pm$0.63& 94.27$\pm$0.48&76.17$\pm$0.16&\textbf{96.61$\pm$0.48}& \textbf{95.70$\pm$0.84}& \textbf{94.14$\pm$0.31} \\

\hline

\hline

\parbox[t]{0.01cm}{\multirow{11}{*}{\rotatebox[origin=c]{90}{CIFAR-100}}}&\parbox[t]{0.01cm}{\multirow{3}{*}{\rotatebox[origin=c]{90}{CLF}}}&CE &  \textbf{86.12$\pm$0.97}& 71.87$\pm$0.68& 58.98$\pm$0.55& 41.66$\pm$1.28&36.71$\pm$1.22& 70.17$\pm$0.12& 60.02$\pm$0.20& 48.17$\pm$1.75\\
&&MAE &37.23$\pm$0.13& 36.97$\pm$0.48& 34.63$\pm$0.48& 33.06$\pm$1.75&16.01$\pm$1.13& 29.16$\pm$0.02& 25.64$\pm$0.58& 21.74$\pm$1.02\\

&&FL &83.20$\pm$0.55& 70.56$\pm$0.07& 69.80$\pm$0.75& 42.44$\pm$0.40&22.78$\pm$0.66& 71.34$\pm$0.63& 62.23$\pm$0.29& 52.08$\pm$0.10\\
\cline{2-11}
&\parbox[t]{0.01cm}{\multirow{6}{*}{\rotatebox[origin=c]{90}{NLL}}}&GCE &83.46$\pm$0.55& 83.20$\pm$0.97& 82.42$\pm$0.80& 79.29$\pm$1.98&75.38$\pm$1.77& 82.03$\pm$0.73& 76.55$\pm$0.68& 57.80$\pm$0.18\\

&&SCE &83.20$\pm$0.48& 74.73$\pm$0.84& 61.19$\pm$0.40& 47.26$\pm$0.92&28.51$\pm$0.39& 73.56$\pm$0.80& 60.93$\pm$0.14& 51.55$\pm$0.77\\

&&NLNL& 74.33$\pm$0.63 & 65.82$\pm$0.74 & 52.92$\pm$0.36 &38.52$\pm$0.11 & 10.41$\pm$0.13&63.14$\pm$0.04&41.84$\pm$0.13&36.59$\pm$0.18 \\

&&NCE+RCE &\underline{84.42$\pm$0.76}& 82.81$\pm$0.10& 82.42$\pm$0.38& 80.07$\pm$0.55&\underline{77.34$\pm$0.14}& \underline{83.20$\pm$0.31}& 78.25$\pm$0.28& 64.71$\pm$0.73\\

&&NCE+AGCE & 84.11$\pm$0.11& \textbf{83.85$\pm$0.48}& \underline{82.81$\pm$0.84}& \underline{81.37$\pm$1.02}& \textbf{78.25$\pm$1.41}& \textbf{83.85$\pm$0.97}& \underline{81.63$\pm$0.10} & \underline{70.83$\pm$0.97}\\

&&ANL-CE & 83.79$\pm$0.70& \underline{83.78$\pm$0.58}& \textbf{83.20$\pm$0.68}& \textbf{81.50$\pm$1.21}&65.75$\pm$1.57& 82.55$\pm$0.80& \textbf{82.52$\pm$0.69} & \textbf{77.34$\pm$0.95} \\

\hline

\hline

\end{tabular}
\label{tab:vitb_MLP-K_res}
\end{table*}

\section{Proposed Methodology}
\subsection{Problem Formulation} 
Let  $D = \{(\mathbf{x}_i, y_i)\}_{i=1}^n$ represent the dataset where $\mathbf{x}_i \in \mathcal{X} \subset \mathcal{R}^d$ is a sample and $y_i \in \mathcal{Y} = \{1,..., k_c\}$ denotes its annotated labels from $k_c$ classes (which may include noise). The distribution over different labels for sample $\mathbf{x}_i$ is represented as $\mathbf{q}(k|\mathbf{x}_i)$ with $\Sigma_{k=1}^{k_c} \mathbf{q}(k|\mathbf{x}_i)=1$. In this paper, we focus on the common scenario where there is a single label $y_i$ for each $\mathbf{x}_i$, i.e., $\mathbf{q}(k=y_i|\mathbf{x}_i)=1$ and $\mathbf{q}(k \neq y_i|\mathbf{x}_i)=0$. In this case, $\mathbf{q}$ is simply the one-hot encoding of the label. 

For the classification task, the goal is to learn a function $f(\cdot): \mathcal{X} \rightarrow \mathcal{Y} $ that maps the input space to the label space. 
In this work, we model $f(\cdot)$ using a Vision Transformer (ViT) backbone, followed by one or more dense layers with a softmax applied at the output layer. 
For a sample $\mathbf{x}_i$, we denote the probability output of classifier $f(\mathbf{x}_i)$ as $\mathbf{p}(k|\mathbf{x}_i) = {e^{z_k}}/{\Sigma_{j=1}^{k_c} e^{z_j}}$, where $z_k$ represents the output from last layer before the softmax.
Training the classifier $f(\cdot)$ involves finding an optimal classifier $f^*(\cdot)$ that minimizes the empirical risk defined by a loss function: $f^*(\cdot) \equiv \text{argmin}_{\theta} \Sigma_{i=1}^n \mathcal{L}(f(\mathbf{x}_i,y_i))$, where $\theta$ represents the trainable parameters of $f(\cdot)$.

\subsection{Label Noise Generation.}
To systematically evaluate the robustness of various methods to noisy labels different noise levels are introduced into clean datasets \cite{ma2020normalized, wang2019symmetric, ye2023active, zhou2021asymmetric}.  
There are two common types of label noise: symmetric (or uniform) noise and asymmetric (or class-conditional) noise. Let the overall noise rate be denoted by $\eta \in [0, 1]$ and the class-wise noise rate from class $i$ to class $j$ be denoted by $\eta_{ij}$. Noise is called symmetric if $\eta_{ij} = \frac{\eta}{k_c-1}, \forall j \neq i$. In contrast, asymmetric noise, $\eta_{ij}$ is conditioned on both the true label $i$ and corrupted label $j$. In this case, for a given class $j$, its labels are corrupted by adding  $\eta_{ij}$ labels from a semantically similar class $i$. For Example, if class $i$ represents `cars' and class $j$  represents `trucks', class $j$ may be corrupted by $\eta_{ij}$ images of cars.

\begin{figure*}[t!]
    \centering
    \begin{subfigure}{0.9\textwidth}
        \centering
        \includegraphics[width=0.75\textwidth]{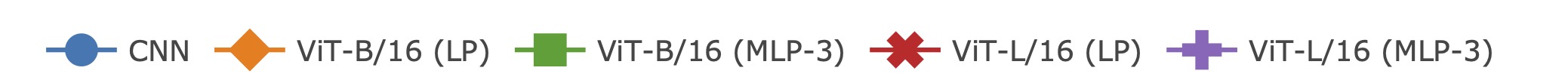}
    \end{subfigure}
    \newline\begin{subfigure}{0.24\textwidth}
        \centering
        \includegraphics[width=0.94\textwidth]{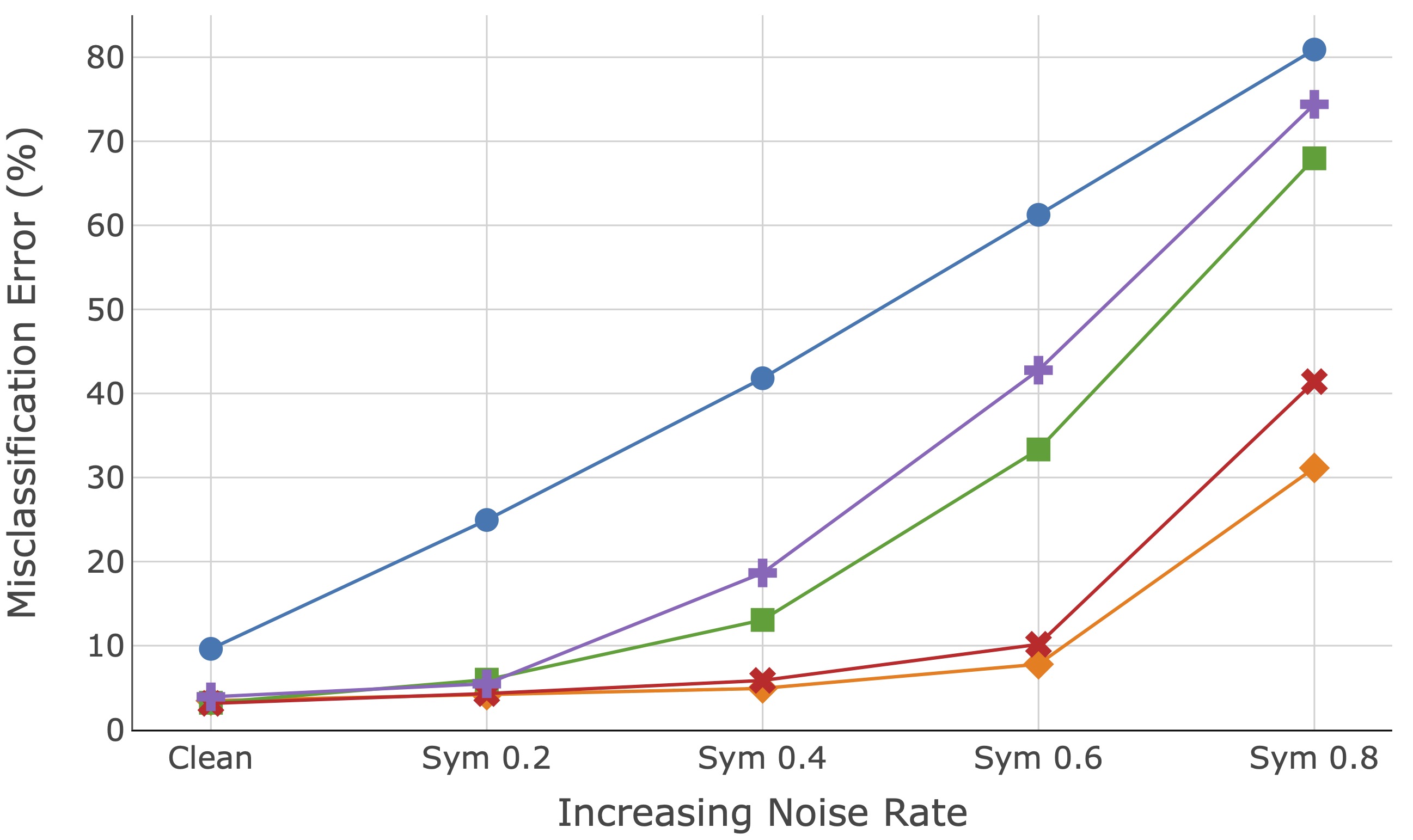}
        \caption{CIFAR-10+Symmetric Noise}
    \end{subfigure}%
    \begin{subfigure}{0.24\textwidth}
        \centering
        \includegraphics[width=0.94\textwidth]{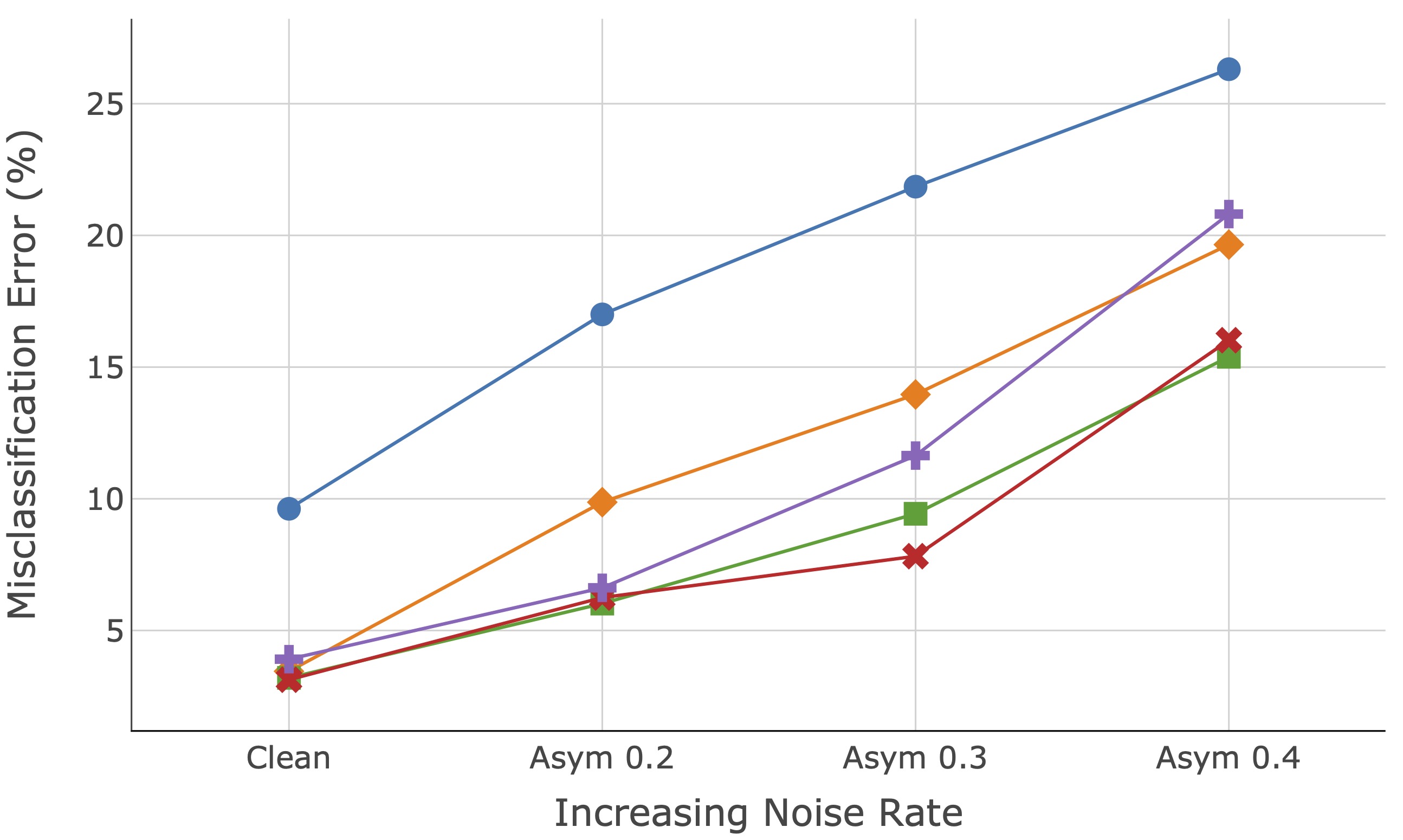}
        \caption{CIFAR-10+Asymmetric Noise}
    \end{subfigure}%
    \begin{subfigure}{0.24\textwidth}
        \centering
        \includegraphics[width=0.91\textwidth]{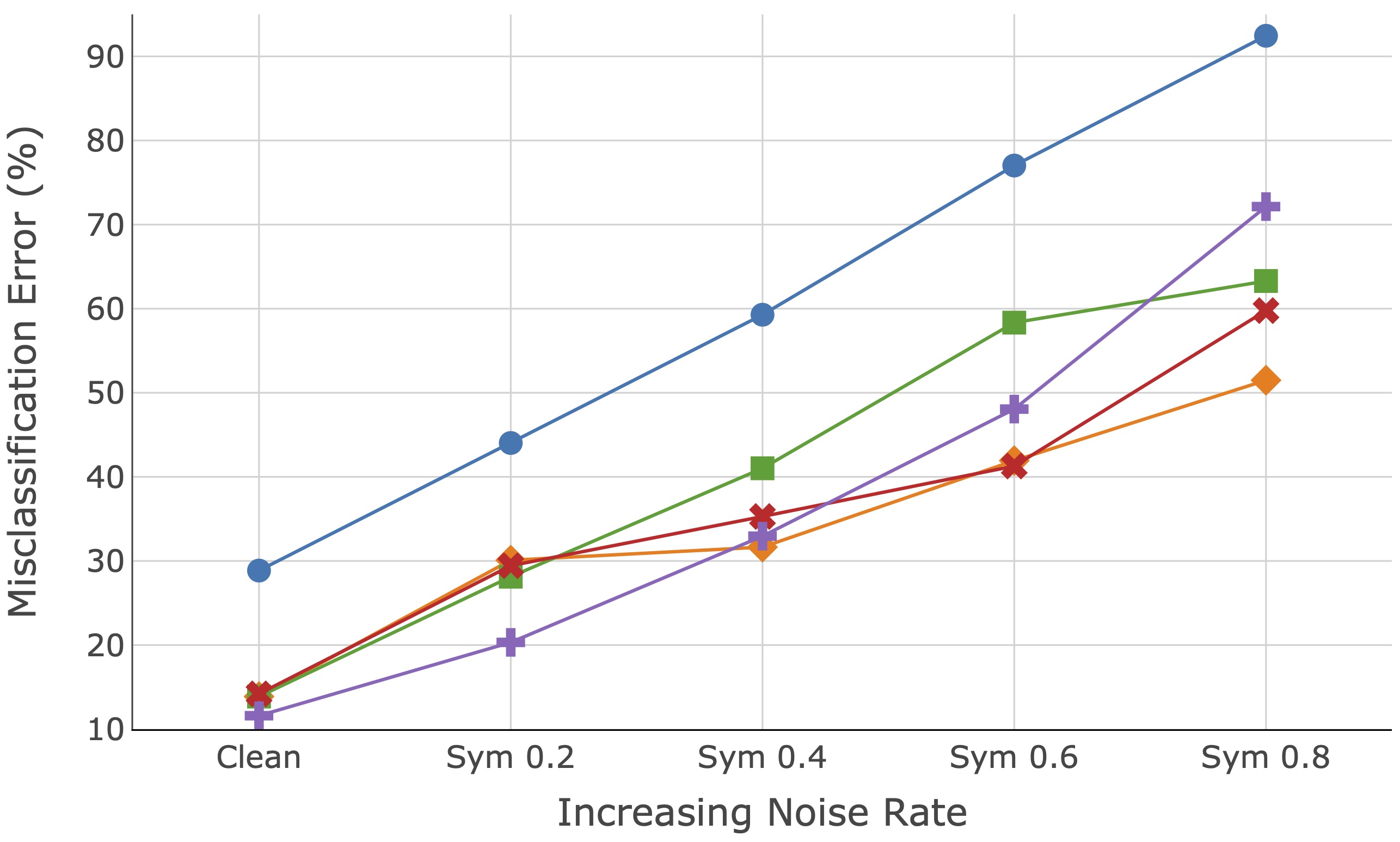}
        \caption{CIFAR-100+Sym. Noise}
    \end{subfigure}%
    \begin{subfigure}{0.24\textwidth}
        \centering
        \includegraphics[width=0.91\textwidth]{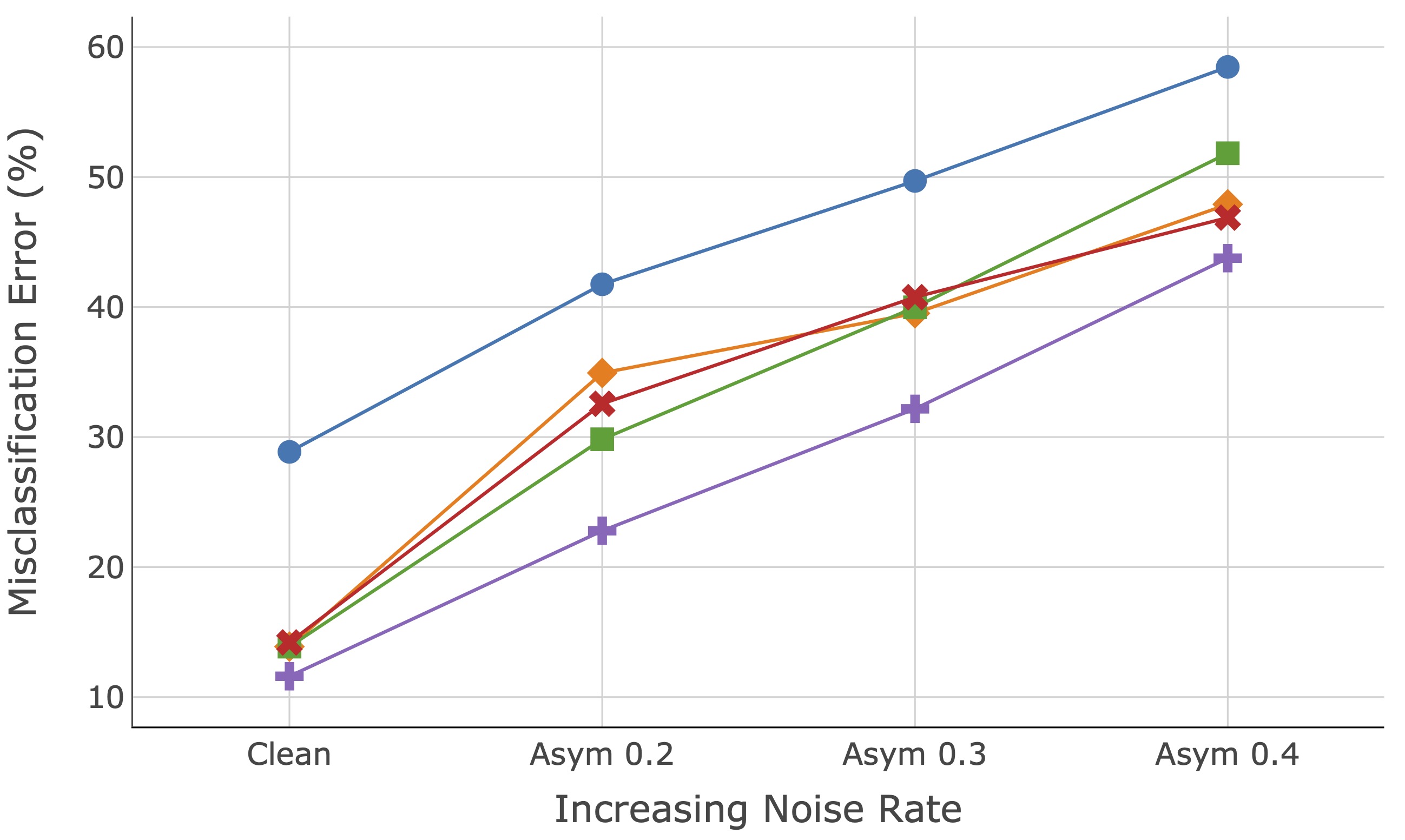}
        \caption{CIFAR-100+Asym. Noise}
    \end{subfigure}%
    
    \caption{Robustness comparison between CNNs and Vision Transformers (ViTs) across different noise types and levels on CIFAR-10 and CIFAR-100 datasets. The misclassification error is plotted against increasing noise rates for both symmetric and asymmetric noise. Results indicate that ViTs exhibit greater robustness to noisy training labels compared to CNNs, particularly as the noise rate increases. Performance is measured using the cross-entropy (CE) loss function across all model backbones.}
    % Robustness comparison between CNNs and Vision Transformers (ViTs)using cross-entropy (CE) loss function. For all backbones, misclassification error increases with increasing noise rate. These experiments demonstrate that ViTs are more robust to noisy training labels as compared to CNNs.}
    \label{fig:cnn_vs_vits}
\end{figure*}

\subsection{Entropy Regularization as a Robust Loss Function}\label{sec:ent_analysis}

\begin{table*}
    \centering
    \caption{\textbf{Comparison of implicit entropy reduction  $\Delta H$ between the 1st and last training epochs, alongside test accuracy (Acc\%).} Commonly used classification loss functions (CLF) and noisy label learning (NLL) methods are evaluated across multiple datasets with a 0.60 symmetric noise rate. The results highlight the performance differences in entropy reduction and accuracy for different Vision Transformer (ViT) variants.}
    \begin{tabular}{ll|cc|cc|cc|cc|cc|cc}
    \hline

    \hline
    &\multirow{2}{4em}{Variant} & \multicolumn{2}{c|}{MNIST} & \multicolumn{2}{c|}{CIFAR-10} & \multicolumn{2}{c}{CIFAR-100} & \multicolumn{2}{c}{WebVision} & \multicolumn{2}{c}{Clothing1M} & \multicolumn{2}{c}{Food-101N}\\
         \cline{3-14}
         &&$\Delta H$&  Acc.&  $\Delta H$&  Acc.&  $\Delta H$& Acc. &  $\Delta H$& Acc. &  $\Delta H$& Acc. &  $\Delta H$& Acc.\\
         \hline

         \hline
         \parbox[t]{0.01cm}{\multirow{4}{*}{\rotatebox[origin=c]{90}{CLF}}}&ViT-B (LP) & 0.174	&94.92	&0.419	&92.21	&0.409	&58.07
         &0.120	&87.79	&0.018	&63.96	&0.46	&75.09\\
         
         &ViT-L (LP) &0.174	&94.92	&0.39	&89.84	&0.412	&58.71	&0.101	&86.71	&0.079	&63.86	&0.521	&81.05\\
         
         &ViT-B (MLP) &0.082	&94.53	&0.153	&66.66	&0.34	&41.66	&0.345	&88.47	&0.204	&64.64	&0.420	&74.31\\
         &ViT-L (MLP) & 0.188	&96.48	&0.103	&57.23	&0.38	&51.94	&0.062	&86.81	&0.26	&65.03	&0.538	&81.34\\
         \cline{2-14}
         \rowcolor{mygray}&\textbf{Average} & 0.15	&95.21	&0.27	&76.49& 	0.39	&52.60	&0.16	&87.45	&0.14	&64.37	&0.48	&77.95\\
         
         \hline

         \hline
         \parbox[t]{0.01cm}{\multirow{5}{*}{\rotatebox[origin=c]{90}{NLL}}}&ViT-B (LP) &0.186	&95.31	&0.913	&95.79	&0.967	&84.76	&0.468	&88.96	&0.028	&63.37	&0.654	&76.60\\
         
         &ViT-L (LP) & 0.190	&95.31	&0.99	&95.96	&0.988	&87.23	&0.556	&90.82	&0.182	&64.06	&0.534	&81.73\\
         
         &ViT-B (MLP) & 0.143	&95.87	&0.985	&94.27	&0.988	&81.50	&0.286	&89.16	&0.419	&65.42	&0.528	&75.18 \\
         
         &ViT-L (MLP) & 0.445	&96.87	&0.985	&95.05	&0.982	&85.15	&0.40	&89.06	&0.434	&65.62	&0.495	&80.07\\
         \cline{2-14}
         \rowcolor{mygray}& \textbf{Average} & 0.24	&95.84	&0.97	&95.27	&0.98	&84.66	&0.43	&89.50	&0.27	&64.62	&0.55	&78.40\\
         \hline

         \hline
    \end{tabular}
    \label{tab:implicit_entropy_comp}
\end{table*}
\subsubsection{Motivation for Entropy Regularization}\label{subsec:ent_reduction}
We have observed a trend of decreasing entropy as the network converges during training. To investigate this, we analyze the entropy of predictions from ViT-B/16 backbone with MLP-3 fine-tuning across consecutive training epochs. The experiments were conducted on the CIFAR-100 dataset with a symmetric noise rate of 0.50. The analysis includes commonly used classification loss functions such as Cross Entropy (CE) and Focal Loss (FL), as well as robust loss functions like NCE+AGCE \cite{zhou2021asymmetric} and ANL-CE \cite{ye2023active}. 
As shown in Fig. \ref{fig:ent_reduction_acc_comp}, there is a consistent decrease in entropy across all loss functions. Notably, the robust loss functions exhibit a larger reduction in entropy compared to the conventional loss functions, which can be attributed to their enhanced performance. 
Throughout the training process, the continuous decrease in entropy suggests an improvement in prediction accuracy, highlighting that robust loss functions implicitly reduce entropy. Additionally, some semi-supervised learning (SSL) methods have also incorporated entropy regularization to enhance performance \cite{grandvalet2004semi, lidividemix, berthelot2019mixmatch}. However, in this work, we propose the use of entropy regularization in supervised learning to address the challenge of noisy labels.  To the best of our knowledge entropy regularization has not previously been applied to improve the robustness of vision transformers (ViTs) against noisy labels.

\subsubsection{Explicit  Entropy Regularization}
Entropy measures the uncertainty or randomness of a probability distribution \cite{renyi1961measures}. In machine learning, it is often used to quantify the uncertainty in a decision. The entropy $H(\mathcal{X})$ for all samples is defined as:
\begin{equation}
    H_l(\mathcal{X}) =  \frac{1}{n}\sum_{i=1}^n\sum_{k=1}^{k_c} \mathbf{p}_l(k|\mathbf{x}_i) \log \frac{1}{\mathbf{p}_l(k|\mathbf{x}_i)},
\end{equation}
where, $\mathbf{p}_l(k|\mathbf{x}_i)$ represents the softmax probability of the classifier for the $k$-th class in $l$-th iteration, while $\sum_{k=1}^{k_c} 
\mathbf{p}_l(k|\mathbf{x}_i) = 1$, and $k_c$ is the number of classes. Entropy reduction $\Delta H_{(l, l+\Delta l)}(\mathcal{X})$ is defined as:

\begin{equation}
\Delta H_{(l, l+\Delta l)}(\mathcal{X}) = H_l(\mathcal{X}) - H_{l+\Delta l}(\mathcal{X}), 
\end{equation}
where, $H_l(\mathcal{X})$ and $H_{l+\Delta l}(\mathcal{X})$ are the mean entropies at $l$-th and $(l+\Delta l)$-th epochs, respectively.

The investigation in the previous section showed that robust loss functions implicitly minimize prediction entropy, leading to a more significant reduction in entropy when dealing with noisy labels. Building on this observation, we propose incorporating explicit entropy minimization in addition to any baseline loss function. For example, if we use a loss function  $L_{b}$ to fine-tune a model with noisy labels, $L_{b}$ will be augmented with explicit entropy minimization, resulting in a modified training loss: 
\begin{equation}
    \mathcal{L}(f(\boldsymbol{x}), y)  =  L_{b} + \lambda_l  H_l(\mathcal{X})  
\end{equation}
where $\lambda_l$ is a hyper-parameter that controls the weight of the entropy term.

\begin{table*}[t!]
    \caption{
    \textbf{Performance comparison of ViT-B/16 and ViT-L/16 models using LP and MLP-3 fine-tuning across six datasets with explicit entropy minimization for robust handling of noisy labels.} The table shows the average test accuracy on clean and noisy data across three Common Loss Functions (CLF) and six state-of-the-art (SOTA) Noisy Label Learning (NLL) methods. Performance on noisy datasets for MNIST, CIFAR-10/100, WebVision, Clothing1M, and Food-101N is evaluated over symmetric noise levels \{0.2, 0.4, 0.6, 0.8\} and asymmetric noise levels \{0.2, 0.3, 0.4\}. For detailed results, refer to Tables \ref{tab:vitb_mlp_explicit_ent_min} and \ref{tab:vitl_mlp_explicit_ent_min}, as well as supplementary Tables XII through XVIII.}
    \centering
    \begin{tabular}{l|l|ll|ll|ll|l|l|l}
    \hline
     & \multirow{2}{4em}{Variants}  & \multicolumn{2}{c|}{MNIST} &  \multicolumn{2}{c|}{CIFAR-10} & \multicolumn{2}{c|}{CIFAR-100} & WebVision & Clothing1M & Food-101N\\
    \cline{3-11}
        & &Clean &  Noisy & Clean & Noisy & Clean & Noisy & Noisy & Noisy & Noisy\\
        \hline
        \parbox[t]{0.01cm}{\multirow{4}{*}{\rotatebox[origin=c]{90}{CLF}}} &ViT-B (LP)& 93.10  \textcolor{blue}{\tiny{($\uparrow$0.53)}} &85.39  \textcolor{blue}{\tiny{($\uparrow$2.65)}} & 96.87 \textcolor{blue}{\tiny{($\uparrow$0.62)}} &92.00  \textcolor{blue}{\tiny{($\uparrow$5.78)}} &78.64  \textcolor{blue}{\tiny{($\uparrow$1.33)}} & 68.08  \textcolor{blue}{\tiny{($\uparrow$9.22)}}& 88.18  \textcolor{blue}{\tiny{($\uparrow$0.39)}} &65.04  \textcolor{blue}{\tiny{($\uparrow$1.08)}} & 75.58 \textcolor{blue}{\tiny{($\uparrow$0.49)}}\\
        
         &ViT-L (LP) & 95.05 \textcolor{blue}{\tiny{($\uparrow$1.04)}} & 89.62 \textcolor{blue}{\tiny{($\uparrow$3.60)}} & 96.87 
         \textcolor{blue}{\tiny{($\uparrow$0.67)}} &89.51 \textcolor{blue}{\tiny{($\uparrow$2.82)}} &84.24 \textcolor{blue}{\tiny{($\uparrow$3.45)}} & 66.98 \textcolor{blue}{\tiny{($\uparrow$9.05)}} &89.16 \textcolor{blue}{\tiny{($\uparrow$2.45)}} &64.84 \textcolor{blue}{\tiny{($\uparrow$0.98)}} &81.35 \textcolor{blue}{\tiny{($\uparrow$0.30)}}\\

         &ViT-B (MLP) &95.05  \textcolor{blue}{\tiny{($\uparrow$0.52)}} &85.92 \textcolor{blue}{\tiny{($\uparrow$1.66)}}& 97.13  \textcolor{blue}{\tiny{($\uparrow$0.61)}}& 84.89 \textcolor{blue}{\tiny{($\uparrow$8.98)}}& 71.61  \textcolor{blue}{\tiny{($\uparrow$2.76)}}& 63.26 \textcolor{blue}{\tiny{($\uparrow$17.36)}} &89.35 \textcolor{blue}{\tiny{($\uparrow$0.88)}} &66.40  \textcolor{blue}{\tiny{($\uparrow$1.76)}}&75.00 \textcolor{blue}{\tiny{($\uparrow$0.69)}}\\

         &ViT-L (MLP) &98.70 \textcolor{blue}{\tiny{($\uparrow$0.39)}}&90.01 \textcolor{blue}{\tiny{($\uparrow$1.08)}} &96.87 \textcolor{blue}{\tiny{($\uparrow$1.17)}} & 85.93 \textcolor{blue}{\tiny{($\uparrow$9.57)}}& 77.13 \textcolor{blue}{\tiny{($\uparrow$1.53)}} & 61.20 \textcolor{blue}{\tiny{($\uparrow$9.09)}} &89.74 \textcolor{blue}{\tiny{($\uparrow$2.93)}} &66.30 \textcolor{blue}{\tiny{($\uparrow$1.27)}} &82.26 \textcolor{blue}{\tiny{($\uparrow$0.92)}}\\
        
        \rowcolor{mygray}&\textbf{Average} & 95.48 \textcolor{blue}{\tiny{($\uparrow$0.62)}}& 87.73 \textcolor{blue}{\tiny{($\uparrow$2.25)}} &96.94 \textcolor{blue}{\tiny{($\uparrow$0.77)}}& 88.08 \textcolor{blue}{\tiny{($\uparrow$6.79)}}& 77.90 \textcolor{blue}{\tiny{($\uparrow$2.27)}}& 64.88 \textcolor{blue}{\tiny{($\uparrow$11.18)}} &89.11 \textcolor{blue}{\tiny{($\uparrow$1.66)}} &65.65 \textcolor{blue}{\tiny{($\uparrow$1.27)}}&78.55 \textcolor{blue}{\tiny{($\uparrow$0.60)}}\\
        
         \hline
         \parbox[t]{0.01cm}{\multirow{4}{*}{\rotatebox[origin=c]{90}{NLL}}}& ViT-B (LP)& 92.50 \textcolor{blue}{\tiny{($\uparrow$0.66)}}& 83.02 \textcolor{blue}{\tiny{($\uparrow$2.37)}}&96.56 \textcolor{blue}{\tiny{($\uparrow$0.29)}}& 95.65 \textcolor{blue}{\tiny{($\uparrow$3.17)}} &85.31 \textcolor{blue}{\tiny{($\uparrow$1.35)}}& 79.96 \textcolor{blue}{\tiny{($\uparrow$4.60)}}&89.08 \textcolor{blue}{\tiny{($\uparrow$0.66)}}&63.79 \textcolor{blue}{\tiny{($\uparrow$1.12)}}&75.13 \textcolor{blue}{\tiny{($\uparrow$0.54)}}\\
         
         & ViT-L (LP)& 96.06 \textcolor{blue}{\tiny{($\uparrow$1.45)}}& 90.75 \textcolor{blue}{\tiny{($\uparrow$10.10)}}&96.01 \textcolor{blue}{\tiny{($\uparrow$0.57)}}& 95.41 \textcolor{blue}{\tiny{($\uparrow$1.83)}}&89.76 \textcolor{blue}{\tiny{($\uparrow$1.77)}}&83.42 \textcolor{blue}{\tiny{($\uparrow$3.85)}}& 90.31 \textcolor{blue}{\tiny{($\uparrow$1.31)}}&65.06 \textcolor{blue}{\tiny{($\uparrow$1.12)}}&80.90 \textcolor{blue}{\tiny{($\uparrow$0.38)}}\\
         
         &ViT-B (MLP)&95.49  \textcolor{blue}{\tiny{($\uparrow$0.84)}} &85.01 \textcolor{blue}{\tiny{($\uparrow$1.01)}} &96.71 \textcolor{blue}{\tiny{($\uparrow$0.48)}} & 93.51 \textcolor{blue}{\tiny{($\uparrow$2.83)}} &85.93 \textcolor{blue}{\tiny{($\uparrow$2.13)}} &82.00 \textcolor{blue}{\tiny{($\uparrow$7.38)}} &88.37 \textcolor{blue}{\tiny{($\uparrow$2.13)}} &65.27 \textcolor{blue}{\tiny{($\uparrow$0.85)}} &74.43 \textcolor{blue}{\tiny{($\uparrow$0.89)}}\\
         
         &ViT-L (MLP) & 96.52 \textcolor{blue}{\tiny{($\uparrow$0.57)}} &  91.37 \textcolor{blue}{\tiny{($\uparrow$1.91)}} &96.56 \textcolor{blue}{\tiny{($\uparrow$1.17)}} & 92.92 \textcolor{blue}{\tiny{($\uparrow$1.80)}} &90.39 \textcolor{blue}{\tiny{($\uparrow$1.98)}} & 85.34 \textcolor{blue}{\tiny{($\uparrow$5.94)}} & 89.62 \textcolor{blue}{\tiny{($\uparrow$1.83)}} &66.78 \textcolor{blue}{\tiny{($\uparrow$1.63)}} & 80.99 \textcolor{blue}{\tiny{($\uparrow$1.26)}}\\
         
         \rowcolor{mygray}&\textbf{Average} & 95.14 \textcolor{blue}{\tiny{($\uparrow$0.88)}}& 87.54 \textcolor{blue}{\tiny{($\uparrow$3.85)}} &96.46 \textcolor{blue}{\tiny{($\uparrow$0.63)}} & 94.37 \textcolor{blue}{\tiny{($\uparrow$2.40)}}&	87.85 \textcolor{blue}{\tiny{($\uparrow$1.81)}}& 82.68 \textcolor{blue}{\tiny{($\uparrow$5.44)}}& 89.35 \textcolor{blue}{\tiny{($\uparrow$1.48)}} &65.23 \textcolor{blue}{\tiny{($\uparrow$1.18)}} &77.87 \textcolor{blue}{\tiny{($\uparrow$0.77)}}\\
         \hline
    \end{tabular}
    
    \label{tab:explicit_ent_min}
\end{table*}

\begin{table*}
\centering
\caption{\textbf{Test accuracy of ViT-B/16 backbone with MLP-3 fine-tuning, showing the effect of explicit entropy minimization on robustness to noisy labels.} The table presents test accuracy results on clean data and across symmetric noise rates ($\eta$) \{0.2, 0.4, 0.6, 0.8\} and asymmetric noise rates ($\eta$) \{0.2, 0.3, 0.4\}. The improvement in accuracy due to the proposed entropy loss is indicated in \textcolor{blue}{blue}. Performance is evaluated across three datasets (MNIST, CIFAR-10, CIFAR-100) using different classification methods and noisy label learning (NLL) techniques.} 
\renewcommand{\arraystretch}{0.94}
\begin{tabular}{cl|l|llll|lll}
\hline

 \hline
& \multirow{2}{4em}{Method} & \multirow{2}{4em}{Clean} & \multicolumn{4}{c|}{Symm Noise Rate ($\eta$)} & \multicolumn{3}{c}{Asym Noise Rate ($\eta$)}\\
\cline{4-10}
&&& 0.2 & 0.4  & 0.6 & 0.8 & 0.2 & 0.3 & 0.4\\
\hline

 \hline

\parbox[t]{0.01cm}{\multirow{8}{*}{\rotatebox[origin=c]{90}{MNIST}}}&CE+$H_l$ & 99.22 \textcolor{blue}{\tiny{($\uparrow$0.39)}} & 98.05 \textcolor{blue}{\tiny{($\uparrow$0.40)}}  & 98.05 \textcolor{blue}{\tiny{($\uparrow$0.79)}} & 96.48 \textcolor{blue}{\tiny{($\uparrow$1.95)}}& 92.08 \textcolor{blue}{\tiny{($\uparrow$0.28)}} & 98.04 \textcolor{blue}{\tiny{($\uparrow$0.39)}}& 97.65 \textcolor{blue}{\tiny{($\uparrow$0.80)}} & 95.70 \textcolor{blue}{\tiny{($\uparrow$0.39)}} \\

&MAE+$H_l$ & 87.50 \textcolor{blue}{\tiny{($\uparrow$0.39)}}&79.68 \textcolor{blue}{\tiny{($\uparrow$1.76)}} &78.51 \textcolor{blue}{\tiny{($\uparrow$2.47)}} &76.95 \textcolor{blue}{\tiny{($\uparrow$8.86)}} &29.29 \textcolor{blue}{\tiny{($\uparrow$2.49)}} &67.96	\textcolor{blue}{\tiny{($\uparrow$0.39)}} &67.07	\textcolor{blue}{\tiny{($\uparrow$8.06)}} &57.81 \textcolor{blue}{\tiny{($\uparrow$0.78)}}\\

&FL+$H_l$ & 98.44 \textcolor{blue}{\tiny{($\uparrow$0.79)}}& 98.05 \textcolor{blue}{\tiny{($\uparrow$0.39)}}& 96.48 \textcolor{blue}{\tiny{($\uparrow$0.39)}}& 96.48 \textcolor{blue}{\tiny{($\uparrow$1.39)}}& 89.28  \textcolor{blue}{\tiny{($\uparrow$0.61)}}& 98.05 \textcolor{blue}{\tiny{($\uparrow$0.40)}} & 97.26 \textcolor{blue}{\tiny{($\uparrow$1.56)}} & 95.31 \textcolor{blue}{\tiny{($\uparrow$0.39)}} \\

&GCE+$H_l$& 97.65  \textcolor{blue}{\tiny{($\uparrow$0.38)}}& 97.27 \textcolor{blue}{\tiny{($\uparrow$0.40)}}& 97.09 \textcolor{blue}{\tiny{($\uparrow$0.22)}} & 96.09 \textcolor{blue}{\tiny{($\uparrow$1.17)}} & 66.79 \textcolor{blue}{\tiny{($\uparrow$2.73)}} & 96.87 \textcolor{blue}{\tiny{($\uparrow$0.78)}} & 96.09 \textcolor{blue}{\tiny{($\uparrow$0.78)}} & 94.92  \textcolor{blue}{\tiny{($\uparrow$3.13)}}\\

&SCE+$H_l$ & 98.05 \textcolor{blue}{\tiny{($\uparrow$0.79)}}& 98.05 \textcolor{blue}{\tiny{($\uparrow$0.79)}} & 97.26 \textcolor{blue}{\tiny{($\uparrow$0.78)}}& 96.88  \textcolor{blue}{\tiny{($\uparrow$1.01)}}& 95.70 \textcolor{blue}{\tiny{($\uparrow$1.17)}}& 97.26 \textcolor{blue}{\tiny{($\uparrow$0.78)}}& 96.87 \textcolor{blue}{\tiny{($\uparrow$0.39)}} & 96.48  \textcolor{blue}{\tiny{($\uparrow$0.39)}}\\

&NCE+RCE+$H_l$ & 98.04 \textcolor{blue}{\tiny{($\uparrow$0.77)}}& 97.27 \textcolor{blue}{\tiny{($\uparrow$0.40)}}& 96.88 \textcolor{blue}{\tiny{($\uparrow$0.40)}}& 96.48 \textcolor{blue}{\tiny{($\uparrow$0.39)}} & 75.94 \textcolor{blue}{\tiny{($\uparrow$2.51)}} & 96.48 \textcolor{blue}{\tiny{($\uparrow$0.39)}}& 88.45 \textcolor{blue}{\tiny{($\uparrow$0.39)}}& 81.11 \textcolor{blue}{\tiny{($\uparrow$1.03)}}\\

&NCE+AGCE+$H_l$ & 89.59 \textcolor{blue}{\tiny{($\uparrow$0.69)}}& 81.64 \textcolor{blue}{\tiny{($\uparrow$0.79)}}& 75.92 \textcolor{blue}{\tiny{($\uparrow$1.31)}}& 61.66 \textcolor{blue}{\tiny{($\uparrow$1.12)}} & 47.66  \textcolor{blue}{\tiny{($\uparrow$0.79)}}& 68.81 \textcolor{blue}{\tiny{($\uparrow$1.63)}}& 57.91 \textcolor{blue}{\tiny{($\uparrow$0.10)}} & 57.62  \textcolor{blue}{\tiny{($\uparrow$0.20)}}\\

&ANL-CE +$H_l$ & 94.14 \textcolor{blue}{\tiny{($\uparrow$1.56)}}& 91.80 \textcolor{blue}{\tiny{($\uparrow$0.79)}} & 85.20 \textcolor{blue}{\tiny{($\uparrow$1.22)}}& 71.48 \textcolor{blue}{\tiny{($\uparrow$2.34)}} & 68.43 \textcolor{blue}{\tiny{($\uparrow$2.03)}}& 91.65 \textcolor{blue}{\tiny{($\uparrow$0.63)}}& 86.33 \textcolor{blue}{\tiny{($\uparrow$0.79)}}& 71.88  \textcolor{blue}{\tiny{($\uparrow$1.57)}}\\

\hline

\hline
\parbox[t]{0.01cm}{\multirow{8}{*}{\rotatebox[origin=c]{90}{CIFAR-10}}}&CE+$H_l$ & 97.26 \textcolor{blue}{\tiny{($\uparrow$0.46)}}&97.26 \textcolor{blue}{\tiny{($\uparrow$3.21)}}		& 96.87 \textcolor{blue}{\tiny{($\uparrow$9.93)}} &96.35 \textcolor{blue}{\tiny{($\uparrow$29.69)}} 	& 93.22 \textcolor{blue}{\tiny{($\uparrow$61.19)}} 	&97.26 \textcolor{blue}{\tiny{($\uparrow$3.28)}}  &	96.35 \textcolor{blue}{\tiny{($\uparrow$5.78)}}	& 95.18 \textcolor{blue}{\tiny{($\uparrow$10.57)}} \\

&MAE+$H_l$ & 96.87 \textcolor{blue}{\tiny{($\uparrow$0.6)}} & 96.48 \textcolor{blue}{\tiny{($\uparrow$0.78)}}& 88.28 \textcolor{blue}{\tiny{($\uparrow$0.78)}}&76.17 \textcolor{blue}{\tiny{($\uparrow$0.35)}} & 37.89 \textcolor{blue}{\tiny{($\uparrow$1.47)}} & 69.14 \textcolor{blue}{\tiny{($\uparrow$1.43)}} & 60.15 \textcolor{blue}{\tiny{($\uparrow$1.26)}}& 58.98 \textcolor{blue}{\tiny{($\uparrow$0.26)}}\\
&FL+$H_l$ & 97.26 \textcolor{blue}{\tiny{($\uparrow$0.76)}}& 95.31 \textcolor{blue}{\tiny{($\uparrow$0.71)}}& 94.53 \textcolor{blue}{\tiny{($\uparrow$5.89)}} &90.23 \textcolor{blue}{\tiny{($\uparrow$19.42)}} & 57.03 \textcolor{blue}{\tiny{($\uparrow$23.56)}} & 97.26 \textcolor{blue}{\tiny{($\uparrow$1.99)}}& 95.31 \textcolor{blue}{\tiny{($\uparrow$1.92)}} & 93.35 \textcolor{blue}{\tiny{($\uparrow$5.15)}} \\

&GCE+$H_l$& 96.87 \textcolor{blue}{\tiny{($\uparrow$0.47)}} & 96.87 \textcolor{blue}{\tiny{($\uparrow$0.60)}}& 96.48 \textcolor{blue}{\tiny{($\uparrow$0.32)}} &96.48 \textcolor{blue}{\tiny{($\uparrow$0.85)}}& 96.09 \textcolor{blue}{\tiny{($\uparrow$3.39)}}& 96.87 \textcolor{blue}{\tiny{($\uparrow$2.72)}} & 95.70 \textcolor{blue}{\tiny{($\uparrow$0.73)}} & 91.40 \textcolor{blue}{\tiny{($\uparrow$2.51)}}\\

&SCE+$H_l$ & 96.48 \textcolor{blue}{\tiny{($\uparrow$0.12)}} & 96.48 \textcolor{blue}{\tiny{($\uparrow$0.47)}}& 96.09 \textcolor{blue}{\tiny{($\uparrow$1.11)}}&92.18 \textcolor{blue}{\tiny{($\uparrow$2.60)}}& 73.06 \textcolor{blue}{\tiny{($\uparrow$24.18)}}& 95.70 \textcolor{blue}{\tiny{($\uparrow$0.22)}}& 94.53 \textcolor{blue}{\tiny{($\uparrow$2.13)}} & 89.84 \textcolor{blue}{\tiny{($\uparrow$5.26)}}\\

&NCE+RCE+$H_l$ & 97.26 \textcolor{blue}{\tiny{($\uparrow$0.98)}}& 96.87 \textcolor{blue}{\tiny{($\uparrow$0.63)}}& 96.48 \textcolor{blue}{\tiny{($\uparrow$0.52)}}&96.09 \textcolor{blue}{\tiny{($\uparrow$0.97)}} & 94.92 \textcolor{blue}{\tiny{($\uparrow$5.26)}}& 97.26 \textcolor{blue}{\tiny{($\uparrow$1.06)}}& 96.09 \textcolor{blue}{\tiny{($\uparrow$0.43)}} & 78.12 \textcolor{blue}{\tiny{($\uparrow$2.93)}} \\

&NCE+AGCE+$H_l$ & 96.87 \textcolor{blue}{\tiny{($\uparrow$0.56)}}& 96.87 \textcolor{blue}{\tiny{($\uparrow$0.79)}} & 96.48 \textcolor{blue}{\tiny{($\uparrow$0.67)}} &96.09 \textcolor{blue}{\tiny{($\uparrow$1.56)}}& 91.40 \textcolor{blue}{\tiny{($\uparrow$2.5)}}& 95.70 \textcolor{blue}{\tiny{($\uparrow$1.17)}}& 96.09 \textcolor{blue}{\tiny{($\uparrow$11.72)}} & 68.35 \textcolor{blue}{\tiny{($\uparrow$0.78)}} \\

&ANL-CE+$H_l$& 96.09 \textcolor{blue}{\tiny{($\uparrow$0.26)}} & 96.48 \textcolor{blue}{\tiny{($\uparrow$0.78)}}  &95.70 \textcolor{blue}{\tiny{($\uparrow$0.78)}}  & 94.92 \textcolor{blue}{\tiny{($\uparrow$0.65)}} & 92.57 \textcolor{blue}{\tiny{($\uparrow$16.4)}}& 96.87 \textcolor{blue}{\tiny{($\uparrow$0.26)}} & 96.87 \textcolor{blue}{\tiny{($\uparrow$1.17)}}& 94.92 \textcolor{blue}{\tiny{($\uparrow$0.78)}} \\
\hline

\hline
\parbox[t]{0.01cm}{\multirow{8}{*}{\rotatebox[origin=c]{90}{CIFAR-100}}}&CE+$H_l$& 86.32 \textcolor{blue}{\tiny{($\uparrow$0.2)}}&84.89 \textcolor{blue}{\tiny{($\uparrow$13.02)}}	&82.68 \textcolor{blue}{\tiny{($\uparrow$23.7)}} &80.33 \textcolor{blue}{\tiny{($\uparrow$38.67)}} &70.04 \textcolor{blue}{\tiny{($\uparrow$33.33)}}	&82.89 \textcolor{blue}{\tiny{($\uparrow$12.72)}} &80.33 \textcolor{blue}{\tiny{($\uparrow$20.31)}}&73.43 \textcolor{blue}{\tiny{($\uparrow$25.26)}} \\

&MAE+$H_l$ &41.79 \textcolor{blue}{\tiny{($\uparrow$4.56)}} &40.62 \textcolor{blue}{\tiny{($\uparrow$3.65)}} 	 &40.23	 \textcolor{blue}{\tiny{($\uparrow$5.6)}}&33.98	\textcolor{blue}{\tiny{($\uparrow$0.92)}}&16.02 \textcolor{blue}{\tiny{($\uparrow$0.01)}}&33.59 \textcolor{blue}{\tiny{($\uparrow$4.43)}}	&30.07 \textcolor{blue}{\tiny{($\uparrow$4.43)}}	&25.78 \textcolor{blue}{\tiny{($\uparrow$4.04)}}\\

&FL+$H_l$ & 86.71 \textcolor{blue}{\tiny{($\uparrow$3.51)}} &83.98 \textcolor{blue}{\tiny{($\uparrow$13.42)}}	&84.37 \textcolor{blue}{\tiny{($\uparrow$14.57)}}&78.51 \textcolor{blue}{\tiny{($\uparrow$36.07)}}&66.01 \textcolor{blue}{\tiny{($\uparrow$43.23)}}	&81.64 \textcolor{blue}{\tiny{($\uparrow$10.3)}}&79.68 \textcolor{blue}{\tiny{($\uparrow$17.45)}}& 73.04 \textcolor{blue}{\tiny{($\uparrow$20.96)}}\\

&GCE+$H_l$ & 84.37 \textcolor{blue}{\tiny{($\uparrow$0.91)}}&84.37 \textcolor{blue}{\tiny{($\uparrow$1.17)}}	&85.93 \textcolor{blue}{\tiny{($\uparrow$3.51)}}	&83.59 \textcolor{blue}{\tiny{($\uparrow$4.3)}} &	82.81 \textcolor{blue}{\tiny{($\uparrow$7.43)}}&84.37 \textcolor{blue}{\tiny{($\uparrow$2.34)}}	&79.68 \textcolor{blue}{\tiny{($\uparrow$3.13)}} &	59.76 \textcolor{blue}{\tiny{($\uparrow$1.96)}}\\

&SCE+$H_l$ & 86.71 \textcolor{blue}{\tiny{($\uparrow$3.51)}} &86.71 \textcolor{blue}{\tiny{($\uparrow$11.98)}}	&83.59 \textcolor{blue}{\tiny{($\uparrow$22.4)}} &83.20 \textcolor{blue}{\tiny{($\uparrow$35.94)}}	&78.12 \textcolor{blue}{\tiny{($\uparrow$49.61)}} &84.37 \textcolor{blue}{\tiny{($\uparrow$10.81)}}&79.68 \textcolor{blue}{\tiny{($\uparrow$18.75)}} &69.92 \textcolor{blue}{\tiny{($\uparrow$18.37)}}\\

&NCE+RCE+$H_l$ &86.71 \textcolor{blue}{\tiny{($\uparrow$2.29)}}&	84.76 \textcolor{blue}{\tiny{($\uparrow$1.95)}}	&83.59 \textcolor{blue}{\tiny{($\uparrow$1.17)}}	&82.81 \textcolor{blue}{\tiny{($\uparrow$2.74)}}	&81.64 \textcolor{blue}{\tiny{($\uparrow$4.3)}}	&86.71 \textcolor{blue}{\tiny{($\uparrow$3.51)}}	&86.32 \textcolor{blue}{\tiny{($\uparrow$8.07)}}&	74.21 \textcolor{blue}{\tiny{($\uparrow$9.5)}}\\

&NCE+AGCE+$H_l$ & 85.93 \textcolor{blue}{\tiny{($\uparrow$1.82)}}&86.32 \textcolor{blue}{\tiny{($\uparrow$2.47)}}	&85.93 \textcolor{blue}{\tiny{($\uparrow$3.12)}}&84.76	 \textcolor{blue}{\tiny{($\uparrow$3.39)}}&82.03	\textcolor{blue}{\tiny{($\uparrow$3.78)}}&85.54	 \textcolor{blue}{\tiny{($\uparrow$1.69)}} &84.76 \textcolor{blue}{\tiny{($\uparrow$3.13)}}	&78.90 \textcolor{blue}{\tiny{($\uparrow$8.07)}}\\

&ANL-CE+$H_l$ & 85.93 \textcolor{blue}{\tiny{($\uparrow$2.14)}}	&85.54 \textcolor{blue}{\tiny{($\uparrow$1.76)}}	&84.37 \textcolor{blue}{\tiny{($\uparrow$1.17)}} &82.42	\textcolor{blue}{\tiny{($\uparrow$0.92)}}&68.35	\textcolor{blue}{\tiny{($\uparrow$2.6)}} &83.98	\textcolor{blue}{\tiny{($\uparrow$1.42)}}&82.81 \textcolor{blue}{\tiny{($\uparrow$0.29)}}	 &77.73 \textcolor{blue}{\tiny{($\uparrow$0.39)}}\\
\hline

\hline

\end{tabular}
\label{tab:vitb_mlp_explicit_ent_min}
\end{table*}

\section{Experiments and Results}\label{sec:benchmark_NLL}
\noindent \textbf{Datasets.} We use six benchmark datasets, including MNIST, CIFAR-10/100, as well as real-world noisy datasets such as WebVision \cite{li2017webvision}, Clothing1M \cite{xiao2015learning}, and Food-101N \cite{lee2017cleannet}, to assess and compare the performance of various NLL methods.   

\noindent \textbf{Baselines.} We consider three commonly used classification losses: CE, FL, and MAE alongside six state-of-the-art (SOTA)NLL methods, including  GCE \cite{zhang2018generalized}, SCE \cite{wang2019symmetric}, NLNL \cite{kim2019nlnl},  APL: NCE+RCE \cite{ma2020normalized}, NCE+AGCE \cite{zhou2021asymmetric}, ANL: ANL-CE  \cite{ye2023active}.

\noindent\textbf{Label Noise Generation.}
Noisy labels for the MNIST and CIFAR-10/100 datasets are generated using standard approaches from previous works \cite{ma2020normalized, ye2023active, wang2019symmetric, zhou2021asymmetric, zhang2018generalized, kim2019nlnl}. 
For symmetric noise, labels within each class are randomly flipped to incorrect labels of other classes.
For asymmetric noise, label flipping occurs within a specified set of classes. Specifically, for MNIST, the label flips are as follows:
$7 \rightarrow 1$, $2 \rightarrow 7$, $5 \leftrightarrow 6$, and $3 \rightarrow 8$ \cite{ye2023active, ma2020normalized}. For CIFAR-10, the flips are TRUCK $\rightarrow$ AUTOMOBILE, BIRD $\rightarrow$AIRPLANE, DEER$\rightarrow$ HORSE, and CAT $\leftrightarrow$ DOG \cite{wang2019symmetric, zhang2018generalized}. For CIFAR-100, the 100 classes are grouped into 20 super-classes, each containing 5 sub-classes. Within each super-class, labels are flipped in a circular manner to the next class.  The noise rate,$\eta$,  is varied as follows: for symmetric noise, $\eta \in \{ 0.2, 0.4, 0.6, 0.8\}$ and for asymmetric noise, $\eta \in \{ 0.2, 0.3, 0.4\}$. Asymmetric noise is kept below 0.50 to prevent flipping to a noisy class.

\noindent\textbf{Experimental Details.} 
We evaluate two Vision Transformer variants (ViT-B/16 and ViT-L/16), both pre-trained on ImageNet-21k \cite{dosovitskiy2020image}.
Following the optimization strategy of Ye et al. \cite{ye2023active}, we use an SGD optimizer 0.90 momentum and a weight decay of $1\times 10^{-3}$ for MNIST, $1\times 10^{-4}$ for CIFAR-10, and $1\times 10^{-5}$ for CIFAR-100. For WebVision, Clothing1M, and Food-101N, we use Nesterov momentum of 0.90 and weight decay of $3\times10^{-5}$ were used. The initial learning rate is set uniformly at 0.001, with a batch size of 256 and gradient norm clipping at 5.0 across all setups \cite{ye2023active}. Baseline method hyperparameters are consistent with those used in the original papers.

\begin{table*}[t!]
\centering
\caption{\textbf{Test accuracy of ViT-L/16 with MLP-3 fine-tuning across three benchmark datasets, demonstrating the impact of explicit entropy minimization on model performance under varying noise levels.} The table reports results on clean data as well as under symmetric noise rates and asymmetric noise rates. Improvements in accuracy due to the proposed entropy loss are highlighted in \textcolor{blue}{blue}. Evaluations are conducted on MNIST, CIFAR-10, and CIFAR-100 using multiple classification methods and noisy label learning (NLL) strategies.}
\renewcommand{\arraystretch}{0.94}
\begin{tabular}{cl|l|llll|lll}
\hline

 \hline
& \multirow{2}{4em}{Method} & \multirow{2}{4em}{Clean} & \multicolumn{4}{c|}{Symm Noise Rate ($\eta$)} & \multicolumn{3}{c}{Asym Noise Rate ($\eta$)}\\
\cline{4-10}
&&& 0.2 & 0.4  & 0.6 & 0.8 & 0.2 & 0.3 & 0.4\\
\hline

 \hline
\parbox[t]{0.01cm}{\multirow{8}{*}{\rotatebox[origin=c]{90}{MNIST}}}&CE+$H_l$&  99.22 \textcolor{blue}{\tiny{($\uparrow$0.40)}}&  98.44 \textcolor{blue}{\tiny{($\uparrow$0.79)}}&  98.04  \textcolor{blue}{\tiny{($\uparrow$1.17)}}&  96.87 \textcolor{blue}{\tiny{($\uparrow$0.39)}}&  92.19 \textcolor{blue}{\tiny{($\uparrow$0.40)}}&  98.82 \textcolor{blue}{\tiny{($\uparrow$1.17)}}&  98.04 \textcolor{blue}{\tiny{($\uparrow$0.39)}}& 98.04 \textcolor{blue}{\tiny{($\uparrow$1.17)}}\\

 &MAE+$H_l$&  97.66 \textcolor{blue}{\tiny{($\uparrow$0.39)}}&  96.09 \textcolor{blue}{\tiny{($\uparrow$1.22)}}&  95.70 \textcolor{blue}{\tiny{($\uparrow$3.13)}}&  86.72  \textcolor{blue}{\tiny{($\uparrow$0.79)}}&  51.95 \textcolor{blue}{\tiny{($\uparrow$0.97)}}&  67.19 \textcolor{blue}{\tiny{($\uparrow$0.40)}}&  67.19 \textcolor{blue}{\tiny{($\uparrow$0.79)}}& 66.8 \textcolor{blue}{\tiny{($\uparrow$0.11)}}\\
 
 &FL+$H_l$&  99.22 \textcolor{blue}{\tiny{($\uparrow$0.39)}}&  98.44 \textcolor{blue}{\tiny{($\uparrow$0.79)}}&  97.65 \textcolor{blue}{\tiny{($\uparrow$1.17)}}&  97.65 \textcolor{blue}{\tiny{($\uparrow$3.12)}}&  91.41 \textcolor{blue}{\tiny{($\uparrow$1.96)}}&  98.04 \textcolor{blue}{\tiny{($\uparrow$0.39)}}&  98.04 \textcolor{blue}{\tiny{($\uparrow$1.17)}}& 96.87 \textcolor{blue}{\tiny{($\uparrow$1.17)}}\\

 &GCE+$H_l$& 98.62 \textcolor{blue}{\tiny{($\uparrow$0.18)}}& 98.44 \textcolor{blue}{\tiny{($\uparrow$0.19)}}& 98.44 \textcolor{blue}{\tiny{($\uparrow$1.57)}}& 97.26 \textcolor{blue}{\tiny{($\uparrow$0.39)}}& 91.41\textcolor{blue}{\tiny{($\uparrow$1.57)}}& 98.05 \textcolor{blue}{\tiny{($\uparrow$0.79)}}& 97.65 \textcolor{blue}{\tiny{($\uparrow$0.39)}}&97.26 \textcolor{blue}{\tiny{($\uparrow$1.57)}}\\

 &SCE+$H_l$& 99.22 \textcolor{blue}{\tiny{($\uparrow$0.40)}}& 98.82 \textcolor{blue}{\tiny{($\uparrow$1.56)}}& 98.82 \textcolor{blue}{\tiny{($\uparrow$1.56)}}& 97.65 \textcolor{blue}{\tiny{($\uparrow$2.73)}}& 96.09 \textcolor{blue}{\tiny{($\uparrow$3.56)}}& 98.82 \textcolor{blue}{\tiny{($\uparrow$0.78)}}& 98.44 \textcolor{blue}{\tiny{($\uparrow$0.40)}}&97.92 \textcolor{blue}{\tiny{($\uparrow$0.27)}}\\

 &NCE+RCE+$H_l$& 98.82 \textcolor{blue}{\tiny{($\uparrow$0.77)}}& 98.04 \textcolor{blue}{\tiny{($\uparrow$1.17)}}& 97.65 \textcolor{blue}{\tiny{($\uparrow$1.56)}}& 96.48 \textcolor{blue}{\tiny{($\uparrow$0.39)}}& 86.33 \textcolor{blue}{\tiny{($\uparrow$13.68)}}& 98.44 \textcolor{blue}{\tiny{($\uparrow$0.40)}}& 98.04 \textcolor{blue}{\tiny{($\uparrow$0.78)}}& 90.67 \textcolor{blue}{\tiny{($\uparrow$1.61)}}\\

 &NCE+AGCE+$H_l$& 88.67 \textcolor{blue}{\tiny{($\uparrow$0.73)}}& 86.71 \textcolor{blue}{\tiny{($\uparrow$1.95)}}& 85.93 \textcolor{blue}{\tiny{($\uparrow$1.56)}}& 82.25 \textcolor{blue}{\tiny{($\uparrow$0.61)}}& 81.25 \textcolor{blue}{\tiny{($\uparrow$7.43)}}& 80.07 \textcolor{blue}{\tiny{($\uparrow$5.07)}}& 67.57 \textcolor{blue}{\tiny{($\uparrow$4.03)}}&66.40 \textcolor{blue}{\tiny{($\uparrow$4.97)}}\\

 &ANL-CE+$H_l$& 97.26 \textcolor{blue}{\tiny{($\uparrow$0.78)}}& 96.09 \textcolor{blue}{\tiny{($\uparrow$0.39)}}& 93.75 \textcolor{blue}{\tiny{($\uparrow$0.78)}}& 87.11 \textcolor{blue}{\tiny{($\uparrow$0.39)}}& 54.39 \textcolor{blue}{\tiny{($\uparrow$0.49)}}& 96.65 \textcolor{blue}{\tiny{($\uparrow$0.17)}}& 96.09 \textcolor{blue}{\tiny{($\uparrow$0.39)}}&92.96 \textcolor{blue}{\tiny{($\uparrow$1.17)}}\\

\hline

\hline
 \parbox[t]{0.01cm}{\multirow{8}{*}{\rotatebox[origin=c]{90}{CIFAR-10}}}&CE+$H_l$& 97.26 \textcolor{blue}{\tiny{($\uparrow$1.17)}}& 96.87  \textcolor{blue}{\tiny{($\uparrow$2.34)}}& 96.48  \textcolor{blue}{\tiny{($\uparrow$15.13)}}& 96.48 \textcolor{blue}{\tiny{($\uparrow$39.25)}}& 95.70 \textcolor{blue}{\tiny{($\uparrow$70.1)}}& 96.48 \textcolor{blue}{\tiny{($\uparrow$3.10)}}& 88.67 \textcolor{blue}{\tiny{($\uparrow$0.31)}}&82.03 \textcolor{blue}{\tiny{($\uparrow$2.84)}}\\

 &MAE+$H_l$& 96.48 \textcolor{blue}{\tiny{($\uparrow$1.17)}}& 96.48 \textcolor{blue}{\tiny{($\uparrow$1.17)}}& 96.48 \textcolor{blue}{\tiny{($\uparrow$1.56)}}& 95.70 \textcolor{blue}{\tiny{($\uparrow$1.56)}}& 95.70 \textcolor{blue}{\tiny{($\uparrow$2.56)}}& 67.96 \textcolor{blue}{\tiny{($\uparrow$2.55)}}& 67.57 \textcolor{blue}{\tiny{($\uparrow$2.26)}}& 61.32 \textcolor{blue}{\tiny{($\uparrow$2.16)}}\\

 &FL+$H_l$& 96.87 \textcolor{blue}{\tiny{($\uparrow$.17)}}& 97.26 \textcolor{blue}{\tiny{($\uparrow$7.03)}}& 96.09 \textcolor{blue}{\tiny{($\uparrow$13.41)}}& 61.71 \textcolor{blue}{\tiny{($\uparrow$3.23)}}& 27.34  \textcolor{blue}{\tiny{($\uparrow$1.02)}}& 96.87 \textcolor{blue}{\tiny{($\uparrow$7.03)}}& 96.48 \textcolor{blue}{\tiny{($\uparrow$8.59)}}&94.92 \textcolor{blue}{\tiny{($\uparrow$13.68)}}\\

 &GCE+$H_l$& 96.48 \textcolor{blue}{\tiny{($\uparrow$0.78)}}& 96.48 \textcolor{blue}{\tiny{($\uparrow$1.17)}}& 96.09 \textcolor{blue}{\tiny{($\uparrow$1.17)}}& 95.31 \textcolor{blue}{\tiny{($\uparrow$0.78)}}& 90.62 \textcolor{blue}{\tiny{($\uparrow$18.75)}}& 96.09 \textcolor{blue}{\tiny{($\uparrow$0.78)}}& 94.53 \textcolor{blue}{\tiny{($\uparrow$2.74)}}&86.71 \textcolor{blue}{\tiny{($\uparrow$3.12)}}\\

 &SCE+$H_l$& 96.09 \textcolor{blue}{\tiny{($\uparrow$0.78)}}& 95.70 \textcolor{blue}{\tiny{($\uparrow$0.39)}}& 95.70 \textcolor{blue}{\tiny{($\uparrow$1.17)}}& 87.10 \textcolor{blue}{\tiny{($\uparrow$2.73)}}& 43.75 \textcolor{blue}{\tiny{($\uparrow$3.98)}}& 94.53 \textcolor{blue}{\tiny{($\uparrow$0.39)}}& 92.18 \textcolor{blue}{\tiny{($\uparrow$2.73)}}&83.98 \textcolor{blue}{\tiny{($\uparrow$0.39)}}\\

 &NCE+RCE+$H_l$& 96.48 \textcolor{blue}{\tiny{($\uparrow$0.78)}}& 96.09 \textcolor{blue}{\tiny{($\uparrow$0.78)}}& 96.09 \textcolor{blue}{\tiny{($\uparrow$0.78)}}& 93.79 \textcolor{blue}{\tiny{($\uparrow$0.04)}}& 90.62 \textcolor{blue}{\tiny{($\uparrow$1.95)}}& 96.48 \textcolor{blue}{\tiny{($\uparrow$0.78)}}& 96.09 \textcolor{blue}{\tiny{($\uparrow$0.78)}}&94.53 \textcolor{blue}{\tiny{($\uparrow$1.18)}}\\

 &NCE+AGCE+$H_l$& 96.48 \textcolor{blue}{\tiny{($\uparrow$1.95)}}& 96.48 \textcolor{blue}{\tiny{($\uparrow$1.95)}}& 96.09 \textcolor{blue}{\tiny{($\uparrow$2.34)}}& 95.70 \textcolor{blue}{\tiny{($\uparrow$2.00)}}& 91.79 \textcolor{blue}{\tiny{($\uparrow$1.17)}}& 96.48 \textcolor{blue}{\tiny{($\uparrow$0.39)}}& 96.10 \textcolor{blue}{\tiny{($\uparrow$0.01)}}&95.70 \textcolor{blue}{\tiny{($\uparrow$0.71)}}\\

 &ANL-CE+$H_l$& 97.26 \textcolor{blue}{\tiny{($\uparrow$1.56)}}& 96.87 \textcolor{blue}{\tiny{($\uparrow$1.30)}}& 96.09 \textcolor{blue}{\tiny{($\uparrow$0.78)}}& 95.70 \textcolor{blue}{\tiny{($\uparrow$0.65)}}& 95.31 \textcolor{blue}{\tiny{($\uparrow$1.56)}}& 97.26 \textcolor{blue}{\tiny{($\uparrow$1.56)}}& 96.87 \textcolor{blue}{\tiny{($\uparrow$2.34)}}&93.35 \textcolor{blue}{\tiny{($\uparrow$1.69)}}\\
\hline

\hline
\parbox[t]{0.01cm}{\multirow{8}{*}{\rotatebox[origin=c]{90}{CIFAR-100}}}&CE+$H_l$& 89.84 \textcolor{blue}{\tiny{($\uparrow$1.44)}}& 88.28 \textcolor{blue}{\tiny{($\uparrow$8.60)}}& 85.54 \textcolor{blue}{\tiny{($\uparrow$18.49)}}& 83.20 \textcolor{blue}{\tiny{($\uparrow$31.26)}}& 78.51 \textcolor{blue}{\tiny{($\uparrow$50.65)}}& 83.20 \textcolor{blue}{\tiny{($\uparrow$5.99)}}& 76.17 \textcolor{blue}{\tiny{($\uparrow$8.34)}}&68.75 \textcolor{blue}{\tiny{($\uparrow$12.51)}}\\

 &MAE+$H_l$& 51.71 \textcolor{blue}{\tiny{($\uparrow$0.55)}}& 48.44 \textcolor{blue}{\tiny{($\uparrow$0.24)}}& 45.31 \textcolor{blue}{\tiny{($\uparrow$3.26)}}& 39.84 \textcolor{blue}{\tiny{($\uparrow$3.52)}}& 28.90 \textcolor{blue}{\tiny{($\uparrow$4.69)}}& 40.62 \textcolor{blue}{\tiny{($\uparrow$4.69)}}& 32.42 \textcolor{blue}{\tiny{($\uparrow$1.57)}}&30.07 \textcolor{blue}{\tiny{($\uparrow$0.78)}}\\

 &FL+$H_l$& 89.84 \textcolor{blue}{\tiny{($\uparrow$2.61)}}& 87.11 \textcolor{blue}{\tiny{($\uparrow$7.30)}}& 71.48 \textcolor{blue}{\tiny{($\uparrow$5.99)}}& 50.78 \textcolor{blue}{\tiny{($\uparrow$1.44)}}& 29.30 \textcolor{blue}{\tiny{($\uparrow$1.70)}}& 81.25 \textcolor{blue}{\tiny{($\uparrow$7.56)}}& 72.65 \textcolor{blue}{\tiny{($\uparrow$6.38)}}&63.28 \textcolor{blue}{\tiny{($\uparrow$5.86)}}\\
 
 &GCE+$H_l$& 89.84 \textcolor{blue}{\tiny{($\uparrow$1.69)}}& 88.28 \textcolor{blue}{\tiny{($\uparrow$0.53)}}& 87.89 \textcolor{blue}{\tiny{($\uparrow$0.27)}}& 88.28 \textcolor{blue}{\tiny{($\uparrow$2.74)}}& 85.54 \textcolor{blue}{\tiny{($\uparrow$6.90)}}& 89.84 \textcolor{blue}{\tiny{($\uparrow$2.61)}}& 84.37 \textcolor{blue}{\tiny{($\uparrow$6.78)}}&75.00 \textcolor{blue}{\tiny{($\uparrow$14.98)}}\\

 &SCE+$H_l$& 90.23 \textcolor{blue}{\tiny{($\uparrow$2.08)}}& 89.06 \textcolor{blue}{\tiny{($\uparrow$6.91)}}& 87.50 \textcolor{blue}{\tiny{($\uparrow$16.02)}}& 87.10 \textcolor{blue}{\tiny{($\uparrow$34.77)}}& 51.17 \textcolor{blue}{\tiny{($\uparrow$22.92)}}& 86.71 \textcolor{blue}{\tiny{($\uparrow$9.24)}}& 78.12 \textcolor{blue}{\tiny{($\uparrow$10.55)}}&67.18 \textcolor{blue}{\tiny{($\uparrow$12.11)}}\\

 &NCE+RCE+$H_l$& 89.84 \textcolor{blue}{\tiny{($\uparrow$2.09)}}& 89.45 \textcolor{blue}{\tiny{($\uparrow$1.96)}}& 89.45 \textcolor{blue}{\tiny{($\uparrow$2.35)}}& 89.06 \textcolor{blue}{\tiny{($\uparrow$3.13)}}& 86.32 \textcolor{blue}{\tiny{($\uparrow$7.16)}}& 90.23 \textcolor{blue}{\tiny{($\uparrow$3.65)}}& 89.84 \textcolor{blue}{\tiny{($\uparrow$11.07)}}&67.96 \textcolor{blue}{\tiny{($\uparrow$5.99)}}\\

 &NCE+AGCE+$H_l$& 91.01 \textcolor{blue}{\tiny{($\uparrow$1.69)}}& 91.01 \textcolor{blue}{\tiny{($\uparrow$2.61)}}& 89.06 \textcolor{blue}{\tiny{($\uparrow$1.18)}}& 87.11 \textcolor{blue}{\tiny{($\uparrow$1.05)}}& 85.93  \textcolor{blue}{\tiny{($\uparrow$2.87)}}& 91.00 \textcolor{blue}{\tiny{($\uparrow$4.68)}}& 89.84 \textcolor{blue}{\tiny{($\uparrow$7.30)}}&85.54 \textcolor{blue}{\tiny{($\uparrow$14.45)}}\\

 &ANL-CE+$H_l$& 91.02 \textcolor{blue}{\tiny{($\uparrow$2.35)}}& 90.23 \textcolor{blue}{\tiny{($\uparrow$1.82)}}& 89.84 \textcolor{blue}{\tiny{($\uparrow$1.44)}}& 88.67 \textcolor{blue}{\tiny{($\uparrow$3.52)}}& 87.89 \textcolor{blue}{\tiny{($\uparrow$4.04)}}& 89.84 \textcolor{blue}{\tiny{($\uparrow$2.35)}}& 88.67 \textcolor{blue}{\tiny{($\uparrow$3.26)}}&83.98 \textcolor{blue}{\tiny{($\uparrow$10.16)}}\\
 \hline

 \hline
 
\end{tabular}
\label{tab:vitl_mlp_explicit_ent_min}
\end{table*}

\begin{table}[t!]
    \centering
    \caption{\textbf{Test accuracy of ViT-B/16 and ViT-L/16 backbones fine-tuned with LP and MLP-3 on real-world noisy datasets, showing the impact of explicit entropy minimization.} The table presents accuracy results on WebVision, Clothing1M, and Food-101N datasets, comparing different classification methods and noisy label learning (NLL) techniques. The improvements attributed to the proposed entropy loss are highlighted in \textcolor{blue}{blue}.}
    \begin{tabular}{ll|cc|cc}
    \hline

    \hline
    &\multirow{2}{4em}{Method}& \multicolumn{2}{c}{ViT-B/16} & \multicolumn{2}{|c}{ViT-L/16} \\ 
    \cline{3-6}
        && LP  & MLP-3 & LP & MLP-3\\
        \hline

        \hline
    \parbox[t]{0.01cm}{\multirow{6}{*}{\rotatebox[origin=c]{90}{WebVision}}}&CE+$H_l$&  88.2 \textcolor{blue}{\tiny{($\uparrow$0.39)}}&  89.4 \textcolor{blue}{\tiny{($\uparrow$0.88)}}&  89.2 \textcolor{blue}{\tiny{($\uparrow$2.45)}}& 89.7 \textcolor{blue}{\tiny{($\uparrow$2.93)}}\\

 &GCE+$H_l$&  89.8 \textcolor{blue}{\tiny{($\uparrow$0.68)}}&  81.7  \textcolor{blue}{\tiny{($\uparrow$4.98)}}&  91.3 \textcolor{blue}{\tiny{($\uparrow$1.17)}}& 85.2 \textcolor{blue}{\tiny{($\uparrow$0.39)}}\\

 &SCE+$H_l$&  86.7 \textcolor{blue}{\tiny{($\uparrow$0.68)}}&  89.0  \textcolor{blue}{\tiny{($\uparrow$1.56)}}&  87.9 \textcolor{blue}{\tiny{($\uparrow$3.03)}}& 90.9 \textcolor{blue}{\tiny{($\uparrow$2.73)}}\\

 &NCE+RCE+$H_l$& 90.3 \textcolor{blue}{\tiny{($\uparrow$1.66)}}& 90.5 \textcolor{blue}{\tiny{($\uparrow$1.95)}}& 90.7 \textcolor{blue}{\tiny{($\uparrow$1.27)}}&90.2 \textcolor{blue}{\tiny{($\uparrow$1.66)}}\\

 &NCE+AGCE+$H_l$& 89.6  \textcolor{blue}{\tiny{($\uparrow$0.39)}}& 90.3 \textcolor{blue}{\tiny{($\uparrow$0.98)}}& 90.7 \textcolor{blue}{\tiny{($\uparrow$0.98)}}&90.9 \textcolor{blue}{\tiny{($\uparrow$2.54)}}\\

 &ANL-CE+$H_l$& 88.9 \textcolor{red}{\tiny{($\downarrow$0.09)}}& 90.3 \textcolor{blue}{\tiny{($\uparrow$1.17)}}& 90.9 \textcolor{blue}{\tiny{($\uparrow$0.1)}}&90.9 \textcolor{blue}{\tiny{($\uparrow$1.85)}}\\
\hline

\hline
\parbox[t]{0.01cm}{\multirow{6}{*}{\rotatebox[origin=c]{90}{Clothing1M}}}
&CE+$H_l$&  65.0 \textcolor{blue}{\tiny{($\uparrow$1.08)}}&  66.4\textcolor{blue}{\tiny{($\uparrow$1.76)}}&  64.8 \textcolor{blue}{\tiny{($\uparrow$0.98)}}& 66.3 \textcolor{blue}{\tiny{($\uparrow$1.27)}}\\

&GCE+$H_l$& 64.6 \textcolor{blue}{\tiny{($\uparrow$2.24)}}	&66.0 \textcolor{blue}{\tiny{($\uparrow$0.60)}}	&65.2 \textcolor{blue}{\tiny{($\uparrow$1.27)}}	&66.4 \textcolor{blue}{\tiny{($\uparrow$0.79)}} \\

&SCE+$H_l$& 64.9 \textcolor{blue}{\tiny{($\uparrow$1.57)}}	&62.5 \textcolor{blue}{\tiny{($\uparrow$0.29)}}	&65.9 \textcolor{blue}{\tiny{($\uparrow$1.86)}}	&65.9  \textcolor{blue}{\tiny{($\uparrow$0.99)}}\\

&NCE+RCE+$H_l$& 63.0	\textcolor{blue}{\tiny{($\uparrow$0.79)}} &65.8	\textcolor{blue}{\tiny{($\uparrow$0.30)}} &65.0 \textcolor{blue}{\tiny{($\uparrow$0.98)}}	&67.5 \textcolor{blue}{\tiny{($\uparrow$2.06)}}\\

&NCE+AGCE+$H_l$& 63.0	\textcolor{blue}{\tiny{($\uparrow$0.49)}} &66.2 \textcolor{blue}{\tiny{($\uparrow$1.57)}}	&65.1	\textcolor{blue}{\tiny{($\uparrow$1.18)}} &66.8 \textcolor{blue}{\tiny{($\uparrow$1.96)}}\\

&ANL-CE+$H_l$& 63.3	\textcolor{blue}{\tiny{($\uparrow$0.49)}}&65.8 \textcolor{blue}{\tiny{($\uparrow$1.47)}}	&64.0	\textcolor{blue}{\tiny{($\uparrow$0.29)}} &67.3 \textcolor{blue}{\tiny{($\uparrow$2.34)}}\\
\hline

\hline
\parbox[t]{0.01cm}{\multirow{6}{*}{\rotatebox[origin=c]{90}{Food-101N}}}
&CE+$H_l$&  75.6 \textcolor{blue}{\tiny{($\uparrow$0.49)}}&75.0 \textcolor{blue}{\tiny{($\uparrow$0.69)}} &81.4 \textcolor{blue}{\tiny{($\uparrow$0.30)}}	&82.3 \textcolor{blue}{\tiny{($\uparrow$0.92)}}\\

&GCE+$H_l$& 76.9 \textcolor{blue}{\tiny{($\uparrow$0.35)}}&73.7	\textcolor{blue}{\tiny{($\uparrow$1.57)}}&82.0	\textcolor{blue}{\tiny{($\uparrow$0.30)}} &80.7 \textcolor{blue}{\tiny{($\uparrow$0.59)}} \\

&SCE+$H_l$&  75.3 \textcolor{blue}{\tiny{($\uparrow$1.27)}}&74.5 \textcolor{blue}{\tiny{($\uparrow$1.57)}}&82.0 \textcolor{blue}{\tiny{($\uparrow$0.88)}}	&79.0 \textcolor{blue}{\tiny{($\uparrow$2.55)}}\\

&NCE+RCE+$H_l$&76.6	\textcolor{blue}{\tiny{($\uparrow$0.39)}} &75.5 \textcolor{blue}{\tiny{($\uparrow$0.59)}} &80.9 	\textcolor{blue}{\tiny{($\uparrow$0.11)}} &82.0 \textcolor{blue}{\tiny{($\uparrow$1.18)}} \\

&NCE+AGCE+$H_l$& 76.6 \textcolor{blue}{\tiny{($\uparrow$0.30)}} &75.3 \textcolor{blue}{\tiny{($\uparrow$0.11)}} 	&81.3 \textcolor{blue}{\tiny{($\uparrow$0.40)}} 	&82.5 \textcolor{blue}{\tiny{($\uparrow$1.37)}} \\

&ANL-CE+$H_l$&70.3 \textcolor{blue}{\tiny{($\uparrow$0.39)}} &73.1 \textcolor{blue}{\tiny{($\uparrow$0.59)}}	&78.2 \textcolor{blue}{\tiny{($\uparrow$0.20)}}	&80.8 \textcolor{blue}{\tiny{($\uparrow$0.59)}} \\
\hline
    \end{tabular}  
    \label{tab:real_world_data_res}
\end{table}

\subsection{Vulnerability of ViT Fine-Tuning to Noisy Labels}\label{sec:finetuning}
We assess five popular fine-tuning techniques-Full-FT, AdaptFormer (AF) \cite{chen2022adaptformer}, VPT \cite{jia2022visual}, MLP-K, and LP \cite{he2020momentum}- for Vision Transformers under noisy label conditions, as illustrated in Fig. \ref{fig:finetune_tech}.  The performance of these techniques on the CIFAR-10 dataset,  under both symmetric and asymmetric noises, is shown in Fig. \ref{fig:finetune_comp}. 
Performance drops for Full-FT, AF, VPT, MLP-K, and LP were  \{72.8\%, 48.0\%, 64.8\%, 64.8\%, 34.1\%\} at 0.80 symmetric noise,  and \{58.0\%,  35.1\%, 25.6\%, 12.2\%, 35.0\%\} at 0.40 asymmetric noise, respectively. 
Although all methods experience significant performance degradation due to noisy labels, Full-FT suffers the largest accuracy decline. This may be attributed to the distortion caused by noisy labels in the learned feature spaces, consistent with previous findings in out-of-distribution \cite{kumar2022fine} and adversarial transfer learning studies \cite{hua2023initialization}.  
On average, LP emerges as the most robust fine-tuning technique, likely because it tunes fewer parameters on noisy labels. The training time comparison in Fig. \ref{fig:finetune_comp}(c) shows that AF, despite fewer parameters, requires 20 times longer to train than LP. This is due to AF's inability to reuse previous computations, whereas LP and MLP-K can be adapted to leverage prior computations, making them the most efficient fine-tuning methods.

\vspace{-3mm}
\subsection{Robustness Comparison of CNNs Vs. ViTs}
We compare the robustness of CNNs and Vision Transformers (ViTs) using CIFAR-10/100 under both symmetric and asymmetric noise settings, as shown in Fig. \ref{fig:cnn_vs_vits}.  
For CNNs, we follow the setup from a recent state-of-the-art method \cite{ye2023active}. For each ViT variant (ViT-B/16 and ViT-L/16), we evaluate two fine-tuning techniques: MLP-3 and LP. Across a wide range of experiments, we observe that ViTs demonstrate significantly higher robustness compared to CNNs. 
\vspace{-3mm}
\subsection{Effectiveness of Existing NLL methods for ViTs}
We assess the performance of five existing noisy label learning NLL loss functions on two ViT variants (ViT-B/16 and ViT-L/16), each using two fine-tuning techniques (MLP-3 and LP) across six datasets: MNIST, CIFAR-10/100, WebVision, Clothing1M, and Food-101N. For the first three clean datasets, we experimented with symmetric noise levels  \{0.2, 0.4, 0.6, 0.8\} and asymmetric noise levels \{0.2, 0.3, 0.4\}. 
Table \ref{tab:LF_vs_NLL_comp}, presents average test accuracy across these datasets. The `Noisy' column reflects the results averaged over all noise levels, while `CLF' denotes the average performance for commonly used loss functions and `NLL' for the five NLL methods. Our findings show that NLL methods generally enhance performance on the CIFAR-10/100 and WebVision datasets, but lead to reduced accuracy on MNIST. For Clothing1M and Food-101N, performance remains similar for both the CLF and NLL methods.
Detailed results for the ViT-B/16 model fine-tuned with MLP-3 on  MNIST and CIFAR-10/100 are provided in Table \ref{tab:vitb_MLP-K_res}, with a summary of results for real-world noisy datasets in Table \ref{tab:real_world_data_res}. Further detailed results on MNIST and CIFAR-10/100 can be found in 
Tables  IX, X, and XI in the supplementary document.
In these tables, we observe that for each noise level, the highest performance across both CLF and NLL loss functions is consistently achieved by an NLL loss function. However, the top-performing NLL function varies under different settings. Therefore, while existing NLL methods originally designed for CNNs do improve the robustness of ViTs against noisy labels, our analysis also indicates that there is substantial room for further improvement in ViT performance under noisy conditions.

\begin{figure}[t!]
    \centering
    \begin{subfigure}{0.45\textwidth}
        \centering
        \includegraphics[width=0.95\textwidth]{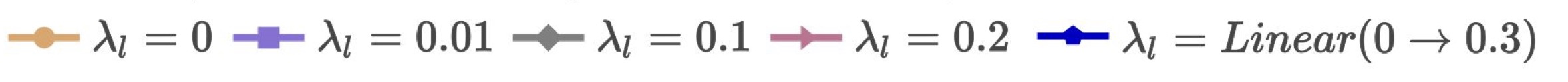}
    \end{subfigure}
    \newline\begin{subfigure}{0.24\textwidth}
        \centering
        \includegraphics[width=0.94\textwidth]{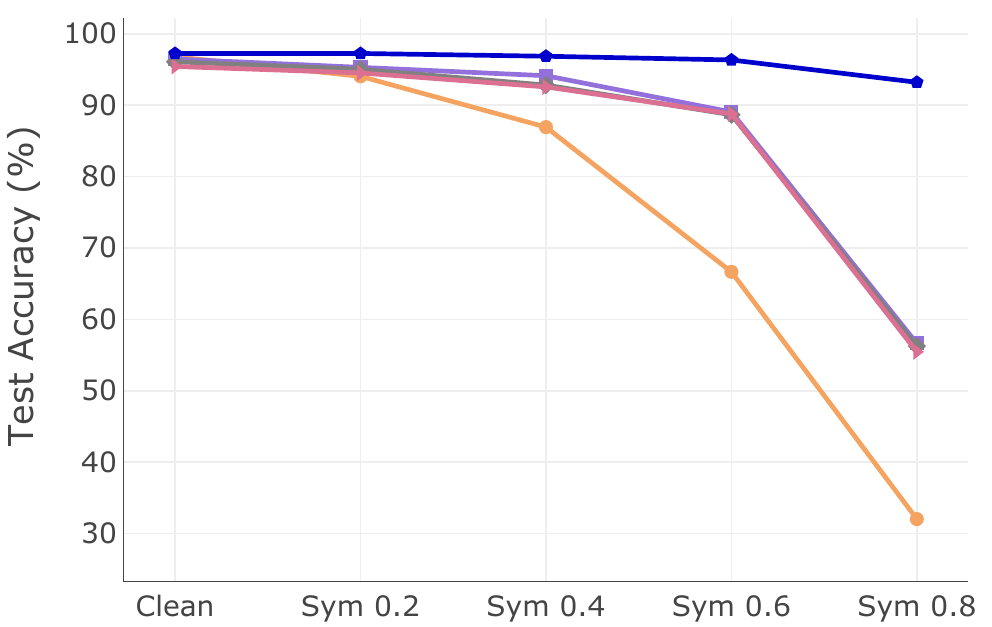}
        \caption{Symmetric Noise}
    \end{subfigure}%
    \begin{subfigure}{0.24\textwidth}
        \centering
        \includegraphics[width=0.94\textwidth]{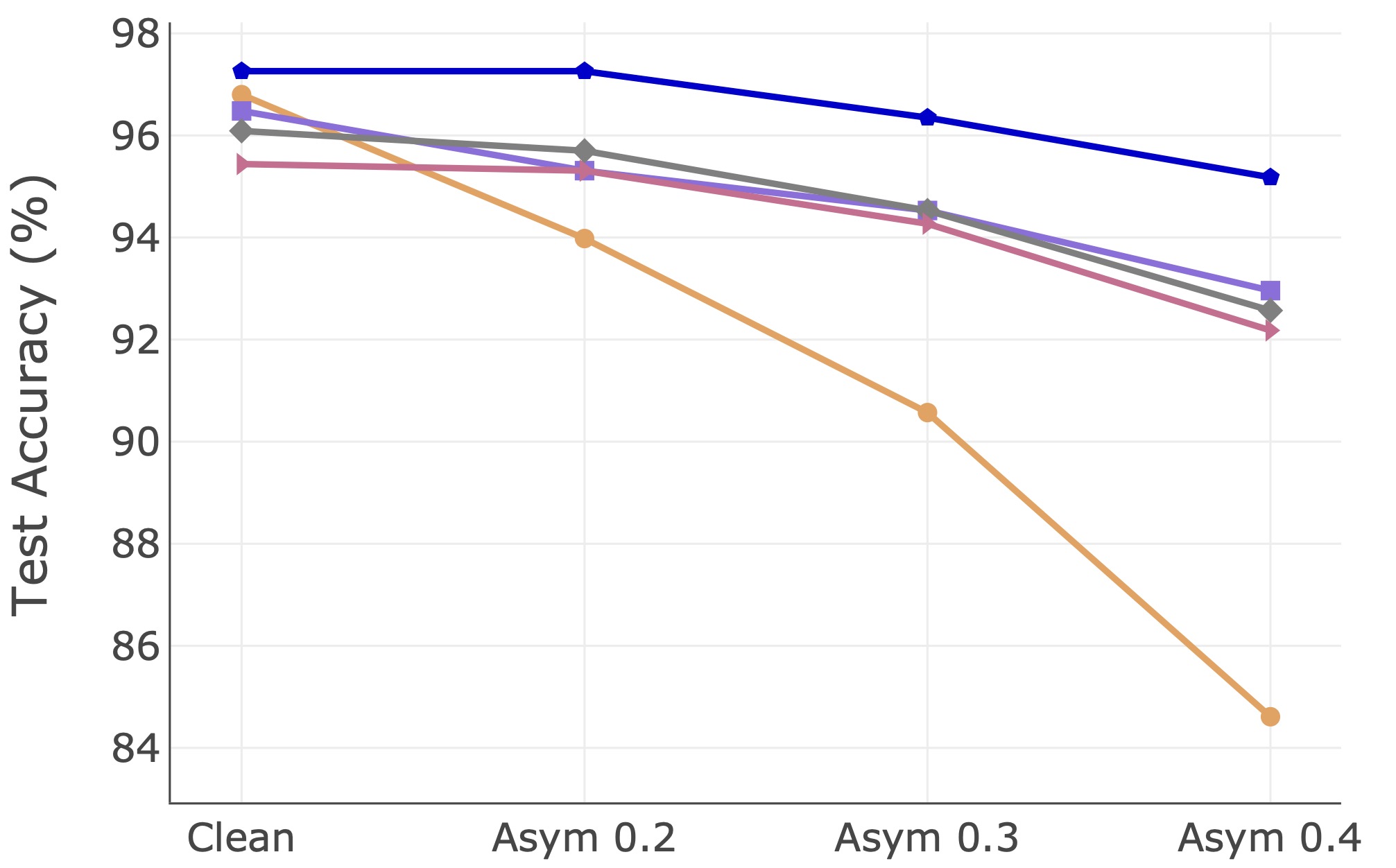}
        \caption{Asymmetric Noise}
    \end{subfigure}%
    \caption{\textbf{Impact of varying $\lambda_l$ on test accuracy for CIFAR-10 using CE+$\lambda_lH_l$ with ViT-B/16+MLP-3 under (a) symmetric noise and (b) asymmetric noise.} The linear scheduling of $\lambda_l$ (Linear(0$\rightarrow$0.3)) achieves the best performance across both noise types.}
    \label{fig:cifar10_lamda_comp}
\end{figure}

\subsection{Implicit Entropy Minimization Relation with Performance}
We examine the relationship between implicit entropy minimization on noisy training data and performance on validation/test data across six datasets. The Cross Entropy (CE) loss function was analyzed alongside five robust NLL loss functions including GCE, SCE, NCE+RCE, NCE+AGCE, and ANL-CE.  
For clean datasets like MNIST and CIFAR-10/100, we applied 0.6 symmetric noise, whereas no additional noise was added to the real-world noisy datasets.
Experiments were conducted using both LP and MLP-3 finetuning methods with ViT-B/16 and ViT-L/16 backbones. The results, summarized in Table \ref{tab:implicit_entropy_comp}, show the CE's performance as a CLF, while for NLL, the best-performing method (in terms of test accuracy) is reported for each dataset.  
On average, NLL methods exhibited more significant entropy reduction compared to CLFs, which likely contributed to their superior performance. This implies that NLL methods implicitly reduce entropy, correlating entropy reduction with performance improvement. A more detailed analysis of entropy minimization is available in Table XX of the supplementary document. Additionally, Fig.\ref{fig:ent_reduction_acc_comp}, shows a consistent decrease in entropy over epochs, with improved validation accuracy as entropy decreases. These findings suggest that robust loss functions implicitly facilitate entropy minimization.

\subsection{Explicit Entropy Minimization Improves Robustness }
In the previous section, we observed that entropy reduction occurs implicitly as networks converge, and this reduction is positively associated with model performance. Building on this observation, we evaluate the performance benefits of explicit entropy reduction, as proposed in Section \ref{sec:ent_analysis}.

\subsubsection{Enhancing ViTs Robustness to Noisy Labels Through Entropy Minimization}
In this experiment, we varied the symmetric noise rates for clean datasets as \{0, 0.2, 0.4, 0.6, 0.8\} and the asymmetric noise rates as \{0.2, 0.3, 0.4\}. The experiment was repeated for common loss functions such as CE, FL, and MAE, as well as for NLL methods, including GCE, SCE, NCE+RCE, NCE+AGCE, and ANL-CE. We used ViT-B/16 and ViT-L/16 backbones with LP and MLP-3 fine-tuning techniques. Table \ref{tab:explicit_ent_min}, shows the average performance for each backbone and fine-tuning technique across six  datasets. Compared to the baseline in Table \ref{tab:LF_vs_NLL_comp}, the performance improvement is shown with ($\uparrow \downarrow$) indicators. For the CIFAR-100 dataset, a 17.36\% average performance improvement was observed on noisy data with ViT-B/16+MLP-3 using CLF.    

Tables \ref{tab:vitb_mlp_explicit_ent_min} and \ref{tab:vitl_mlp_explicit_ent_min} detail the performance for ViT-B/16 and ViT-L/16 backbones, both fine-tuned with MLP-3.
The MNIST dataset saw a maximum improvement of 8.86\%, while CIFAR-10 achieved a 61.19\% improvement using the CE+$H_l$ loss function compared to the standard CE loss. For CIFAR-100, the maximum improvement was 38.67\%. Similar improvements were observed in ViT-L/16 models, where the MNIST dataset improved by 13.68\% with NCE+RCE+$H_l$ loss compared to the NCE+RCE baseline. CIFAR-10/100 datasets saw improvements of 18.75\% and 50.65\%, respectively.
These trends are consistent across all experiments, as evidenced by benchmarks in Tables XII-XVII in the supplementary document, aligning with the results in Tables \ref{tab:vitb_mlp_explicit_ent_min} and \ref{tab:vitl_mlp_explicit_ent_min}. 

For real-world noisy datasets, explicit entropy minimization also resulted in significant improvements, as shown in Table \ref{tab:real_world_data_res}. The ViT-L/16 model achieved 90.92\% accuracy on the WebVision dataset using the ANL-CE+$H_l$ loss function. For the Clothing1M dataset, ViT-L/16+MLP-3 achieved 67.48\% accuracy with the NCE+RCE+$H_l$ loss function, and for the Food-101N dataset, the best performance of 82.03\% was achieved using the GCE+$H_l$ and NCE+RCE+$H_l$ loss functions. Across all real-world noisy datasets, explicit entropy minimization proved highly effective, demonstrating consistent accuracy improvements for both robust and non-robust loss functions.

\begin{table}[t!]
\caption{\textbf{Impact of explicit entropy minimization on CNN robustness to noisy labels:} Test accuracy (\%) on three datasets under different noise conditions. Results for clean data ($\eta$ = 0.0), symmetric noise ($\eta$ = 0.6, 0.8), and asymmetric noise ($\eta$ = 0.3, 0.4) are shown. Extended results can be found in Appendix D, Table XIX.} 
\centering 
 \begin{tabular}{cl|c|cc|cc}
    \hline

    \hline
        &\multirow{2}{4em}{Method}&  Clean &  \multicolumn{2}{c}{Sym Noise ($\eta$)} & \multicolumn{2}{|c}{Asym Noise ($\eta$)}\\
         \cline{3-7}
         &&  ($\eta$=0.0)& 0.6 & 0.8 &0.3 &0.4\\
    \hline

    \hline
         \parbox[t]{0.01cm}{\multirow{4}{*}{\rotatebox[origin=c]{90}{MNIST}}}&CE&  \underline{99.20}  &49.19 &22.51 &88.90&81.79\\
         &ANL-CE & 99.08  & \underline{98.42} &\underline{96.62} & \underline{98.91} & \underline{98.01}\\
         &\textcolor{blue}{CE$+H_l$} & \textbf{99.28}  & 78.05 & 47.85  & 91.30 & 83.90\\
         &\textcolor{blue}{ANL-CE$+H_l$} & 99.13  & \textbf{98.53} &\textbf{96.70}  & \textbf{99.03} &\textbf{98.35} \\
        
         \hline

         \hline
         \parbox[t]{0.01cm}{\multirow{4}{*}{\rotatebox[origin=c]{90}{CIFAR10}}}&CE&  90.38  &38.75 &19.09 & 78.15 & 73.69 \\
        
         &ANL-CE &\underline{91.66}  &\underline{81.12} &\underline{61.27}  &\underline{85.52} & 77.63\\
         
         &\textcolor{blue}{CE$+H_l$} &90.57 & 66.17 &38.75 &81.42 &77.21\\
        &\textcolor{blue}{ANL-CE$+H_l$} &\textbf{91.97} &\textbf{81.86} &\textbf{62.92}  &\textbf{86.79} &\textbf{82.12}\\
         \hline

         \hline
         \parbox[t]{0.01cm}{\multirow{4}{*}{\rotatebox[origin=c]{90}{CIFAR100}}}&CE& \textbf{71.14}  &22.98 &7.55  & 50.30 & 41.53 \\
        
         &ANL-CE &70.68  &\underline{51.52} &\underline{28.07}  & \underline{59.76} & \underline{45.41}\\
         % \cline{2-7}
         &\textcolor{blue}{CE$+H_l$} &\underline{71.04} &34.76&17.28 &49.97 &41.58 \\
         &\textcolor{blue}{ANL-CE$+H_l$}&70.20&\textbf{51.92}& \textbf{28.52}& \textbf{61.70}& \textbf{53.06}\\
         \hline
         
         \hline
    \end{tabular}
\label{tab:CNN_result_short}
\end{table}

\subsubsection{The effect of hyperparameter $\lambda_l$} 
The effect of the hyperparameter $\lambda_l$ on performance was evaluated using ViT-B/16 on the CIFAR-10 dataset, with LP and MLP-3 fine-tuning. The experiments were categorized into two approaches: 1) keeping $\lambda_l$ constant at values {0.01, 0.1, 0.2}, and 2) linearly increasing $\lambda_l$ from 0 to 0.3. Figure 5 compares performance across these different $\lambda_l$ values. A smaller constant $\lambda_l$ may not fully exploit the benefits of entropy regularization, while higher values could negatively impact the training process. A more effective strategy involves gradually increasing $\lambda_l$ from 0 to 0.3, resulting in significant performance improvements across different noise levels and fine-tuning techniques. This approach initially prioritizes the baseline loss for learning task-specific features and then gradually shifts focus towards entropy regularization, leading to enhanced robustness in handling noisy labels.

\subsubsection{Enhancing CNN Robustness to Noisy Labels Through Entropy Minimization}
We evaluate the impact of explicit entropy minimization on the robustness of CNNs to noisy labels using the MNIST and CIFAR-10/100 datasets. Following the experimental settings of \cite{ye2023active} and  \cite{ma2020normalized}, we compare CNN performance with and without explicit entropy minimization under both clean and noisy label conditions. Experiments are conducted with two symmetric noise rates \{0.60, 0.80\} and two asymmetric noise rates \{0.30, 0.40\}. As shown in Table \ref{tab:CNN_result_short}, the best performance for noisy labels is consistently achieved by NLL loss functions with explicit entropy minimization. Detailed results can be found in Table XIX of the supplementary document. These experiments emphasize the effectiveness of explicit entropy minimization.

\section{Conclusion}
In this paper, we examined the vulnerability of Vision Transformers (ViTs) to noisy training labels during fine-tuning. Our empirical results indicate that full fine-tuning is more susceptible to noisy labels than linear probing. In conditions of extreme label noise, ViT fine-tuning performance can significantly degrade. We tested two ViT backbones, ViT-B/16 and ViT-L/16, with linear probing and MLP-K fine-tuning across six datasets: MNIST, CIFAR-10/100, WebVision, Clothing1M, and Food-101N. We also evaluated three commonly used classification losses (CE, FL, and MAE) and six NLL methods (GCE, SCE, NLNL, NCE+RCE, NCE+AGCE, and ANL-CE).
Upon close examination, we found that all existing NLL methods implicitly minimize prediction entropy. Building on this, we proposed explicit entropy minimization as a general strategy to enhance the robustness of ViT fine-tuning against noisy labels. Our experiments demonstrated that introducing explicit entropy regularization improves ViT robustness in the presence of label noise.

% \bibliography{main}

\appendices

\onecolumn
\section{More Datasets Details}\label{A_subsec:dataset}
We evaluated the performance across six datasets: MNIST, CIFAR-10/100, WebVision \cite{li2017webvision}, Clothing1M \cite{xiao2015learning}, and Food-101N \cite{lee2017cleannet}. 
\textbf{The MNIST dataset} contains 70,000 grayscale images of handwritten digits, each measuring $28\times28$ pixels, divided into 10 classes.  We used the standard 60,000/10,000 train/test split and report the results on the test set. 
\textbf{CIFAR-10/100} consists of 60,000 color images of size  $32\times32$ pixels, categorized into 10 and 100 classes, with 6,000 and 600 images per class, respectively. We followed the standard 50,000/10,000 train/test split and report test results for both CIFAR-10 and CIFAR-100. Additionally, we reserved 10\% of the training data as the validation set for MNIST, CIFAR-10, and CIFAR-100.
\textbf{The WebVision dataset} contains over 2.4 million images collected from the web using search queries based on the 1,000 classes of the ILSVRC 2012 benchmark \cite{deng2009imagenet}. For our experiments, we used the "mini" version of WebVision, as proposed by \cite{ye2023active, jiang2018mentornet}, focusing on the first 50 classes from the Google resized image subset.
\textbf{Clothing1M} \cite{xiao2015learning}  is a dataset of images of clothing items collected from online retail websites, divided into 14 classes. It contains one million images with noisy labels, primarily due to automated annotations derived from the surrounding text.
Finally, \textbf{Food-101N} \cite{lee2017cleannet} is a dataset of around 310,009 images, categorized into 101 classes of various food recipes. 

\section{Effectiveness of Existing NLL methods for ViTs}
Table I, reports the average test accuracy of ViT-B/16 and ViT-L/16 models using both linear probing (LP) and MLP-3 fine-tuning across six datasets. The performance of Common Loss Functions (CLF) was averaged over Cross Entropy (CE), Focal Loss (FL), and Mean Absolute Error (MAE). For Noisy Label Learning (NLL) methods, the performance was averaged across Generalized Cross Entropy (GCE), Symmetric Cross Entropy (SCE), Negative Learning for Noisy Labels (NLNL), NCE+RCE, NCE+AGCE, and ANL-CE. For noisy datasets like MNIST, CIFAR-10/100, the performance was averaged over symmetric noise levels \{0.2, 0.4,0.6,0.8\} and asymmetric noise levels \{0.2, 0.3, 0.4\}. Detailed results are presented in Table II of the main paper and in Tables IX, X, and XI in this supplementary document.

%ViT-B/16+LP
\begin{table*}[b!]
\centering
\caption{\textbf{Effectiveness of existing NLL methods for fine-tuning ViTs:} Detailed test accuracy (\%) of the ViT-B/16 backbone with linear probing (LP) across three benchmarks (MNIST, CIFAR-10, CIFAR-100) under varying levels of symmetric and asymmetric noise. The \nth{1} and \nth{2} best results are highlighted in \textbf{bold} and \underline{underlined}.} 
\begin{tabular}{p{0.01cm}p{0.01cm}l|l|cccc|ccc}
\hline

\hline
 & &\multirow{2}{4em}{Method} & \multirow{2}{4em}{Clean} & \multicolumn{4}{c|}{Sym Noise Rate ($\eta$)} & \multicolumn{3}{c}{Asym Noise Rate ($\eta$)}\\
\cline{5-11}
&&&&  0.2 & 0.4 &  0.6 & 0.8 & 0.2 & 0.3 & 0.4\\
\hline

\hline

\parbox[t]{0.01cm}{\multirow{11}{*}{\rotatebox[origin=c]{90}{MNIST}}}&\parbox[t]{0.01cm}{\multirow{3}{*}{\rotatebox[origin=c]{90}{CLF}}}&CE & 96.48$\pm$0.14	&96.09$\pm$0.51	&95.70$\pm$0.18	&94.92$\pm$0.80	&87.89$\pm$0.32& 94.14$\pm$0.18	&90.23$\pm$0.66	&85.54$\pm$0.31 \\

&&MAE &  85.15$\pm$0.04	&76.17$\pm$0.12	&75.39$\pm$0.02	&66.79$\pm$	0.03&53.90$\pm$0.04& 63.73$\pm$0.14	&57.62$\pm$0.001 &57.32$\pm$0.03\\

&&FL &  96.09$\pm$0.03	&95.70$\pm$0.21	&95.97$\pm$0.38	&91.66$\pm$0.37	&87.50$\pm$0.95& 94.53$\pm$0.18	&91.41$\pm$0.55	&86.32$\pm$0.18\\
\cline{2-11}

&\parbox[t]{0.01cm}{\multirow{6}{*}{\rotatebox[origin=c]{90}{NLL}}}&GCE &  95.31$\pm$0.51	&94.92$\pm$0.01	&93.75$\pm$0.01	&91.41$\pm$	0.39 &87.11$\pm$1.77&94.53$\pm$0.04	&93.35$\pm$0.18	&89.84$\pm$0.18\\

&&SCE & 96.48$\pm$0.02	&96.09$\pm$0.11	&96.09$\pm$0.06	&95.31$\pm$0.25	&91.79$\pm$1.01& 95.70$\pm$0.46	&87.89$\pm$0.33	&79.29$\pm$0.28\\

&&NLNL& 87.63$\pm$0.18 &86.45$\pm$0.18 &84.50$\pm$0.97 & 81.24$\pm$0.95 & 69.26$\pm$3.31&86.58$\pm$0.18	&85.93$\pm$	0.31&81.90$\pm$0.48\\

&&NCE+RCE & 97.26$\pm$0.11&  96.48$\pm$0.1&  96.48$\pm$0.49&  96.09$\pm$0.32&  89.06$\pm$3.51 & 96.87$\pm$0.18&  88.67$\pm$0.46& 78.52$\pm$0.41 \\

&&NCE+AGCE & 82.81$\pm$0.94	&75.39$\pm$0.72	&74.22$\pm$0.74	&72.26$\pm$0.04	&56.64$\pm$0.44&67.96$\pm$0.94	&59.38$\pm$0.07	&56.64$\pm$0.23\\

&&ANL-CE &87.34$\pm$0.55	&82.34$\pm$0.36	&68.75$\pm$0.37	&61.58$\pm$1.33	&50.64$\pm$0.55&67.96$\pm$0.31	&53.90$\pm$0.66	&45.70$\pm$0.48\\

\hline

\hline

\parbox[t]{0.01cm}{\multirow{11}{*}{\rotatebox[origin=c]{90}{CIFAR-10}}} &\parbox[t]{0.01cm}{\multirow{3}{*}{\rotatebox[origin=c]{90}{CLF}}}&CE &  \underline{96.55$\pm$0.05}&  95.89$\pm$0.13&  95.08$\pm$0.11&  92.21$\pm$0.85& 68.86$\pm$0.08&  90.13$\pm$0.61&  86.04$\pm$0.66&  80.35$\pm$0.78\\

&&MAE &95.83$\pm$0.18&  92.26$\pm$0.04&  91.92$\pm$0.01&  86.06$\pm$1.47& 66.79$\pm$2.72 &  92.96$\pm$0.62&  87.37$\pm$0.64&  79.94$\pm$0.49\\

&&FL &  96.36$\pm$0.18&  95.87$\pm$0.02&  94.97$\pm$0.03&  92.44$\pm$0.01& 68.15$\pm$0.02&  91.54$\pm$0.18&  84.76$\pm$0.23&  77.18$\pm$0.91\\

\cline{2-11}
&\parbox[t]{0.01cm}{\multirow{6}{*}{\rotatebox[origin=c]{90}{NLL}}}&GCE & 96.26$\pm$0.03&  95.31$\pm$0.31&  95.57$\pm$0.36&  95.55$\pm$0.34& 91.08$\pm$1.81&  90.95$\pm$0.40&  88.11$\pm$0.07&  79.37$\pm$1.91 \\

&&SCE & 96.45$\pm$0.02&  96.17$\pm$0.06&  95.89$\pm$0.13&  95.24$\pm$0.01& 88.5$\pm$0.01&  96.05$\pm$0.10&  95.35$\pm$0.04&  91.71$\pm$0.09\\

&&NLNL&  95.83$\pm$0.66&92.15$\pm$0.48&86.05$\pm$0.36 &33.61$\pm$0.49&20.88$\pm$0.79&90.18$\pm$0.18&84.92$\pm$0.55&79.61$\pm$0.18\\

&&NCE+RCE & 96.27$\pm$0.02& 95.57$\pm$0.18& 95.41$\pm$0.02& 95.18$\pm$0.02&92.70$\pm$0.48& 90.43$\pm$0.14& 89.98$\pm$0.31& 85.34$\pm$0.80\\

&&NCE+AGCE & 96.37$\pm$0.03& 96.29$\pm$0.01& 96.16$\pm$0.02& 95.79$\pm$0.02&92.37$\pm$0.02& 90.67$\pm$0.21& 83.80$\pm$0.50& 81.85$\pm$0.73\\

&&ANL-CE &95.97$\pm$0.36& 95.57$\pm$0.18& 95.31$\pm$0.31& 95.05$\pm$0.18&90.23$\pm$1.14& 95.44$\pm$0.48& 95.18$\pm$0.36& 93.61$\pm$0.18\\

\hline

\hline

\parbox[t]{0.01cm}{\multirow{11}{*}{\rotatebox[origin=c]{90}{CIFAR-100}}}&\parbox[t]{0.01cm}{\multirow{3}{*}{\rotatebox[origin=c]{90}{CLF}}}&CE &  86.12$\pm$0.01&  69.91$\pm$0.07&  68.35$\pm$1.38&  58.07$\pm$1.20& 48.51$\pm$0.09&  65.07$\pm$0.98&  60.48$\pm$0.22&  52.1$\pm$0.30\\

&&MAE & 62.49$\pm$0.87&  60.28$\pm$1.28&  58.06$\pm$1.95&  56.24$\pm$1.77& 47.52$\pm$1.04&  48.43$\pm$0.53&  44.72$\pm$0.37&  33.75$\pm$1.75\\

&&FL &  83.33$\pm$0.36&  81.63$\pm$0.84&  80.73$\pm$1.62&  75.12$\pm$1.33& 54.03$\pm$1.81&  63.54$\pm$0.48&  55.59$\pm$0.49&  49.47$\pm$1.12\\
\cline{2-11}
&\parbox[t]{0.9mm}{\multirow{6}{*}{\rotatebox[origin=c]{90}{NLL}}}&GCE &   85.8$\pm$0.18&  85.28$\pm$0.18&  84.76$\pm$0.31&  82.94$\pm$0.48& 80.85$\pm$0.84&  84.24$\pm$0.84&  83.33$\pm$0.80&  69.91$\pm$0.23\\

&&SCE & 79.94$\pm$0.18&  67.05$\pm$0.39&  66.27$\pm$0.84&  58.72$\pm$0.97& 46.96$\pm$1.1&  58.85$\pm$0.48&  50.38$\pm$0.15&  46.09$\pm$1.65\\

&&NLNL& 74.97$\pm$0.03 & 69.12$\pm$0.48 & 63.06$\pm$0.09 &42.03$\pm$0.39 & 31.28$\pm$0.06&73.49$\pm$0.41&68.09$\pm$0.13&51.43$\pm$0.32\\

&&NCE+RCE& 85.28$\pm$0.18& 85.02$\pm$0.48& 84.76$\pm$0.78& 84.24$\pm$1.02&82.16$\pm$1.02& 83.46$\pm$0.48& 83.98$\pm$0.31& 78.77$\pm$0.11\\

&&NCE+AGCE & 85.54$\pm$0.31& 85.02$\pm$0.36& 84.84$\pm$0.91& 84.76$\pm$0.63 & 83.59$\pm$0.84 & 84.69$\pm$0.48 &  84.24$\pm$0.31& 83.78$\pm$0.31\\

&&ANL-CE &83.2$\pm$0.01& 80.2$\pm$0.91& 73.69$\pm$1.39& 67.57$\pm$1.12&66.82$\pm$0.55& 79.16$\pm$0.09& 73.43$\pm$0.36& 57.81$\pm$0.19\\

\hline

\hline
\end{tabular}
\label{tab:vitb_lp_res}
\end{table*}

%ViT-L/16+MLP-3
\begin{table*}[t!]
\centering
\caption{\textbf{Effectiveness of existing NLL methods for fine-tuning ViTs:} Detailed test accuracy (\%) of the ViT-L/16 backbone with MLP-3 fine-tuning across three benchmarks (MNIST, CIFAR-10, CIFAR-100) under varying levels of symmetric and asymmetric noise. The \nth{1} and \nth{2} best results are highlighted in \textbf{bold} and \underline{underlined}.} 
\begin{tabular}{p{0.01cm}p{0.01cm}l|l|cccc|ccc}
\hline

\hline
 &&\multirow{2}{4em}{Method} & \multirow{2}{4em}{Clean} & \multicolumn{4}{c|}{Sym Noise Rate ($\eta$)} & \multicolumn{3}{c}{Asym Noise Rate ($\eta$)}\\
\cline{5-11}
&&&&  0.2 & 0.4 &  0.6 & 0.8 & 0.2 & 0.3 & 0.4\\
\hline

\hline

\parbox[t]{0.01cm}{\multirow{11}{*}{\rotatebox[origin=c]{90}{MNIST}}}& \parbox[t]{0.01cm}{\multirow{3}{*}{\rotatebox[origin=c]{90}{CLF}}}&CE & 98.82$\pm$0.08& \textbf{97.65$\pm$0.02}&96.87$\pm$0.04& 96.48$\pm$0.09&91.79$\pm$0.15 & 97.65$\pm$0.02& 97.65$\pm$0.04&96.87$\pm$0.09\\

&&MAE & 97.27$\pm$0.17& 94.87$\pm$0.29 &92.57$\pm$0.03& 85.93$\pm$0.04 &50.98$\pm$0.88 & 66.79$\pm$0.74& 66.40$\pm$0.07 &66.69$\pm$0.04 \\

&&FL & \textbf{98.83$\pm$0.01}& \textbf{97.65$\pm$0.04}&96.48$\pm$0.01& 94.53$\pm$0.10&89.45$\pm$0.16 & 97.65$\pm$0.05& 96.87$\pm$0.07&95.70$\pm$0.16\\
\cline{2-11}

& \parbox[t]{0.01cm}{\multirow{6}{*}{\rotatebox[origin=c]{90}{NLL}}}&GCE &  98.44$\pm$0.04& \textbf{97.65$\pm$0.03}&96.87$\pm$0.01& \textbf{96.87$\pm$0.02}&89.84$\pm$0.07 & 97.26$\pm$0.01& 97.26$\pm$0.02&95.70$\pm$0.12\\

&&SCE & 98.82$\pm$0.01& 97.26$\pm$0.04&\textbf{97.26$\pm$0.06}& 94.92$\pm$0.04& \textbf{92.53$\pm$0.17}& \textbf{98.04$\pm$0.05}& \textbf{98.04$\pm$0.02} &\textbf{97.65$\pm$0.12} \\

&&NLNL&95.60$\pm$0.07 & 91.39$\pm$0.02 &86.92$\pm$0.18 &43.73$\pm$0.16 &10.10$\pm$0.23 &94.10$\pm$0.18& 86.82$\pm$0.25&75.78$\pm$0.19\\

&&NCE+RCE &98.05$\pm$0.09& 96.87$\pm$0.10&96.09$\pm$0.15& 96.09$\pm$0.08&72.65$\pm$0.25& \textbf{98.04$\pm$0.08}& 97.26$\pm$0.54&89.06$\pm$0.19 \\

&&NCE+AGCE & 87.94$\pm$0.07& 84.76$\pm$0.04 &84.37$\pm$0.10& 81.64$\pm$1.14&73.82$\pm$2.28& 75.00$\pm$0.91& 63.54$\pm$0.90&61.43$\pm$0.51\\

&&ANL-CE & 96.48$\pm$0.16& 95.70$\pm$0.18&92.97$\pm$0.36& 84.72$\pm$0.18&53.90$\pm$1.57 & 96.48$\pm$0.63& 95.70$\pm$0.66&91.79$\pm$0.95\\
\hline

\hline

\parbox[t]{0.01cm}{\multirow{11}{*}{\rotatebox[origin=c]{90}{CIFAR-10}}}& \parbox[t]{0.01cm}{\multirow{3}{*}{\rotatebox[origin=c]{90}{CLF}}} &CE &  \textbf{96.09$\pm$0.02}& 94.53$\pm$0.05& 81.35$\pm$0.23& 57.23$\pm$0.43&25.60$\pm$0.41& 93.38$\pm$0.11& 88.36$\pm$0.21&79.19$\pm$0.39\\

&&MAE &95.31$\pm$0.04& 95.31$\pm$0.01& 94.92$\pm$0.04& 94.14$\pm$0.03&93.14$\pm$0.04 & 65.41$\pm$0.35& 65.31$\pm$0.13&59.16$\pm$0.02\\

&&FL & 95.70$\pm$0.01& 90.23$\pm$0.17& 82.68$\pm$0.27& 58.48$\pm$0.59&26.32$\pm$0.55& 89.84$\pm$0.09& 87.89$\pm$0.30&81.24$\pm$0.24\\
\cline{2-11}
& \parbox[t]{0.01cm}{\multirow{6}{*}{\rotatebox[origin=c]{90}{NLL}}}&GCE & 95.70$\pm$0.08& 95.31$\pm$0.01& 94.92$\pm$0.01& 94.53$\pm$0.04&71.87$\pm$0.64& 95.31$\pm$0.03& 91.79$\pm$0.13&83.59$\pm$0.98\\

&&SCE & 95.31$\pm$0.06& 95.31$\pm$0.05& 94.53$\pm$0.19& 84.37$\pm$0.25&39.77$\pm$0.35& 94.14$\pm$0.01& 89.45$\pm$0.17&83.59$\pm$0.48\\

&&NLNL& 95.74$\pm$0.13 & 91.73$\pm$0.07 & 80.67$\pm$0.09 & 23.08$\pm$0.12 & 10.04$\pm$0.52 & 92.51$\pm$0.10 & 84.74$\pm$0.12 & 80.63$\pm$0.13\\

&&NCE+RCE & 95.70$\pm$0.06& 95.31$\pm$0.04& \textbf{95.31$\pm$0.08}& 93.75$\pm$0.04&88.67$\pm$0.14& 95.70$\pm$0.04 &95.31$\pm$0.07 &93.35$\pm$0.27\\

&&NCE+AGCE & 94.53$\pm$0.05& 94.53$\pm$0.04& 93.75$\pm$0.06& 93.70$\pm$0.10&90.62$\pm$0.46& \textbf{96.09$\pm$0.09}& \textbf{96.09$\pm$0.07}& \textbf{94.99$\pm$0.41} \\

&&ANL-CE & 95.70$\pm$0.55& \textbf{95.57$\pm$0.54}& \textbf{95.31$\pm$0.37}& \textbf{95.05$\pm$0.18}&\textbf{93.75$\pm$0.58}& 95.70$\pm$0.31& 94.53$\pm$0.32&91.66$\pm$0.76\\
\hline

\hline

\parbox[t]{0.01cm}{\multirow{11}{*}{\rotatebox[origin=c]{90}{CIFAR-100}}}& \parbox[t]{0.01cm}{\multirow{3}{*}{\rotatebox[origin=c]{90}{CLF}}}& CE & 88.40$\pm$0.12&  79.68$\pm$0.63&  67.05$\pm$0.81&  51.94$\pm$0.40& 27.86$\pm$1.31&  77.21$\pm$0.60&  67.83$\pm$0.57& 56.24$\pm$1.27 \\

&&MAE & 51.16$\pm$0.93& 48.20$\pm$0.59& 42.05$\pm$0.60& 36.32$\pm$0.50 &24.21$\pm$1.59 & 35.93$\pm$0.75& 30.85$\pm$0.90&29.29$\pm$1.08\\

&&FL & 87.23$\pm$0.67& 79.81$\pm$0.63& 65.49$\pm$0.94& 49.34$\pm$0.92&27.60$\pm$1.63 & 73.69$\pm$0.43& 66.27$\pm$0.39&57.42$\pm$0.84 \\
\cline{2-11}
& \parbox[t]{0.01cm}{\multirow{6}{*}{\rotatebox[origin=c]{90}{NLL}}}&GCE & 88.15$\pm$0.48& 87.75$\pm$0.66& 87.62$\pm$0.48& 85.54$\pm$0.84 &78.64$\pm$1.28 & 87.23$\pm$0.74& 77.59$\pm$0.29&60.02$\pm$0.63\\

&&SCE & 88.15$\pm$0.48& 82.15$\pm$0.75& 71.48$\pm$0.31& 52.33$\pm$0.30 &28.25$\pm$0.80 & 77.47$\pm$0.91& 67.57$\pm$0.39&55.07$\pm$0.32\\

&&NLNL& 82.24$\pm$0.03 & 77.84$\pm$0.02 & 65.69$\pm$0.14 & 10.38$\pm$0.02 & 10.01$\pm$0.05 &76.47$\pm$0.23 & 67.88$\pm$0.24 &51.16$\pm$1.02\\

&&NCE+RCE& 87.75$\pm$0.49& 87.49$\pm$0.15& 87.10$\pm$0.32& 85.93$\pm$0.09&79.16$\pm$0.48 & 86.58$\pm$0.12& 78.77$\pm$0.55&61.97$\pm$0.73\\

&&NCE+AGCE & \textbf{89.32$\pm$0.80} & 88.40$\pm$0.92& 87.88$\pm$0.88& \textbf{86.06$\pm$0.80} &83.06$\pm$0.63 & 86.32$\pm$0.21& 82.54$\pm$0.47&71.09$\pm$0.27\\

&&ANL-CE & 88.67$\pm$0.31& \textbf{88.41$\pm$0.80}& \textbf{88.40$\pm$0.75} & 85.15$\pm$0.22& \textbf{83.85$\pm$0.91} &  \textbf{87.49$\pm$0.39} & \textbf{85.41$\pm$0.50} & \textbf{73.82$\pm$1.15} \\

\hline

\hline
\end{tabular}
\label{tab:vitl_mlp_res}
\end{table*}

%ViT-L/16+LP
\begin{table*}[t!]
\centering
\caption{\textbf{Effectiveness of existing NLL methods for fine-tuning ViTs:} Detailed test accuracy (\%) of the ViT-L/16 backbone with linear probing (LP) across three benchmarks (MNIST, CIFAR-10, CIFAR-100) under varying levels of symmetric and asymmetric noise. The \nth{1} and \nth{2} best results are highlighted in \textbf{bold} and \underline{underlined}.} 
\begin{tabular}{p{0.01cm}p{0.01cm}l|l|cccc|ccc}
\hline

\hline
 &&\multirow{2}{4em}{Method} & \multirow{2}{4em}{Clean} & \multicolumn{4}{c|}{Sym Noise Rate ($\eta$)} & \multicolumn{3}{c}{Asym Noise Rate ($\eta$)}\\
\cline{5-11}
&&&&  0.2 & 0.4 &  0.6 & 0.8 & 0.2 & 0.3 & 0.4\\
\hline

\hline

\parbox[t]{0.01cm}{\multirow{11}{*}{\rotatebox[origin=c]{90}{MNIST}}}&\parbox[t]{0.01cm}{\multirow{3}{*}{\rotatebox[origin=c]{90}{CLF}}}&CE & 98.04$\pm$0.01&  98.04$\pm$0.02 &96.48$\pm$0.02&  94.92$\pm$0.01& 83.20$\pm$0.14 &  95.70$\pm$0.02&  93.35$\pm$0.05& 88.67$\pm$0.13\\

&&MAE & 86.72$\pm$0.13&  86.14$\pm$0.03 &85.54$\pm$0.48&  83.98$\pm$0.81& 73.43$\pm$0.25&  65.62$\pm$0.78&  57.81$\pm$1.25& 57.42$\pm$0.53\\

&&FL & 97.26$\pm$0.01&  96.09$\pm$0.01&96.87$\pm$0.03&  96.48$\pm$0.01& 86.32$\pm$0.05&  94.53$\pm$0.01&  91.41$\pm$0.02& 84.37$\pm$0.12 \\
\cline{2-11}

&\parbox[t]{0.01cm}{\multirow{6}{*}{\rotatebox[origin=c]{90}{NLL}}}&GCE & 96.87$\pm$0.01& 95.70$\pm$0.02&95.70$\pm$0.01& 94.14$\pm$0.01&92.57$\pm$0.02 &  95.31$\pm$0.08&  94.92$\pm$0.01& 91.41$\pm$0.02  \\

&&SCE & 98.43$\pm$0.02&  98.04$\pm$0.08&97.26$\pm$0.08&  95.31$\pm$0.01& 93.35$\pm$0.03&  97.26$\pm$0.02&  97.26$\pm$0.02& 96.09$\pm$0.06\\

&&NLNL& 94.91$\pm$0.04 & 90.58$\pm$0.05 &86.07$\pm$0.14 & 40.23$\pm$0.95 & 13.12$\pm$0.87 &92.66$\pm$0.15&85.27$\pm$0.20&72.56$\pm$0.20\\

&&NCE+RCE & 98.05$\pm$0.03&  98.04$\pm$0.05 &97.27$\pm$0.02 &  95.32$\pm$0.01 & 85.93$\pm$0.89 &  97.65$\pm$0.05& 89.45$\pm$0.03 &69.14$\pm$0.05\\

&&NCE+AGCE & 83.59$\pm$0.34&  83.20$\pm$0.03&82.03$\pm$0.38&  80.46$\pm$0.76 & 75.78$\pm$0.97& 77.34$\pm$0.13& 63.64$\pm$1.52&57.80$\pm$0.04\\

&&ANL-CE &96.09$\pm$0.18& 93.51$\pm$0.13&92.14$\pm$0.18& 84.46$\pm$0.29&53.90$\pm$1.23 & 94.92$\pm$0.37& 93.35$\pm$0.18&91.79$\pm$1.27\\

\hline

\hline

\parbox[t]{0.01cm}{\multirow{11}{*}{\rotatebox[origin=c]{90}{CIFAR-10}}} &\parbox[t]{0.01cm}{\multirow{3}{*}{\rotatebox[origin=c]{90}{CLF}}}&CE &  96.87$\pm$0.01&  95.70$\pm$0.02&  94.14$\pm$0.02&  89.84$\pm$0.03& 58.59$\pm$0.05 &  93.75$\pm$0.01&  92.18$\pm$0.04&  83.98$\pm$0.09\\

&&MAE & 96.82$\pm$0.01&  96.46$\pm$0.02&  96.09$\pm$0.01&  95.70$\pm$0.03& 95.31$\pm$0.04&  83.98$\pm$0.38&  75.00$\pm$0.35&  62.33$\pm$0.21\\

&&FL & 94.92$\pm$0.02&  94.53$\pm$0.02&  93.35$\pm$0.01&  89.84$\pm$0.01& 60.89$\pm$0.14&  94.53$\pm$0.02&  91.40$\pm$0.07&  82.81$\pm$0.10\\
\cline{2-11}
&\parbox[t]{0.01cm}{\multirow{6}{*}{\rotatebox[origin=c]{90}{NLL}}}&GCE & 95.70$\pm$0.01&  95.70$\pm$0.02&  95.31$\pm$0.04&  94.92$\pm$0.02& 87.89$\pm$0.6 &  96.09$\pm$0.01&  93.75$\pm$0.01&  90.44$\pm$0.05\\

&&SCE & 95.31$\pm$0.02&  94.92$\pm$0.01&  94.92$\pm$0.02&  93.75$\pm$0.02& 74.21$\pm$0.07 &  94.92$\pm$0.02&  93.35$\pm$0.06&  90.45$\pm$0.12\\

&&NLNL& 95.75$\pm$0.07 & 90.72$\pm$0.05 & 81.62$\pm$0.12 & 41.38$\pm$0.05 & 19.69$\pm$0.09 &94.32$\pm$0.05 & 91.50$\pm$0.04& 80.34$\pm$0.12\\

&&NCE+RCE & 95.31$\pm$0.03& 94.92$\pm$0.04& 94.53$\pm$0.06& 94.92$\pm$0.06&91.79$\pm$0.09& 95.70$\pm$0.05& 95.31$\pm$0.01& 94.53$\pm$0.26\\

&&NCE+AGCE & 95.31$\pm$0.02& 94.90$\pm$0.04& 94.53$\pm$0.06& 94.14$\pm$0.06&92.18$\pm$0.09& 95.31$\pm$0.01& 94.53$\pm$0.04& 91.01$\pm$0.24\\

&&ANL-CE & 95.57$\pm$0.18& 95.31$\pm$0.01& 95.57$\pm$0.18& 95.96$\pm$0.18&95.09$\pm$0.55& 95.55$\pm$0.01& 95.44$\pm$0.19& 93.56$\pm$0.05\\

\hline

\hline

\parbox[t]{0.01cm}{\multirow{11}{*}{\rotatebox[origin=c]{90}{CIFAR-100}}}&\parbox[t]{0.01cm}{\multirow{3}{*}{\rotatebox[origin=c]{90}{CLF}}}& CE &85.80$\pm$0.36&  70.56$\pm$0.39&  64.71$\pm$0.43&  58.71$\pm$0.74& 40.23$\pm$1.24 &67.44$\pm$0.12&  59.24$\pm$0.83& 53.12$\pm$1.38 \\

&&MAE &70.13$\pm$0.68&  69.53$\pm$0.26&  64.85$\pm$0.24&  64.45$\pm$1.06& 53.12$\pm$1.26 &  53.90$\pm$0.38&  49.34$\pm$1.02& 45.70$\pm$1.14 \\

&&FL & 86.45$\pm$0.29&  69.01$\pm$0.71&  62.75$\pm$0.44&  57.02$\pm$0.48& 39.05$\pm$1.77 & 66.66$\pm$0.63&  57.93$\pm$0.30& 52.47$\pm$1.11 \\
\cline{2-11}
&\parbox[t]{0.01cm}{\multirow{6}{*}{\rotatebox[origin=c]{90}{NLL}}}&GCE & 89.58$\pm$0.18&  88.10$\pm$0.12&  88.02$\pm$0.36&  87.10$\pm$0.32& 84.24$\pm$0.12 & 88.02$\pm$0.18&  84.76$\pm$0.10& 69.01$\pm$0.67 \\

&&SCE & 83.72$\pm$0.18&  57.93$\pm$0.75&  59.89$\pm$0.48&  55.07$\pm$0.29& 38.92$\pm$0.14 &  61.06$\pm$0.75&  55.33$\pm$0.12& 50.38$\pm$0.55\\

&&NLNL&  81.83$\pm$0.01 & 78.18$\pm$0.25 & 69.62$\pm$0.05 & 50.07$\pm$0.14 & 35.08$\pm$0.51 & 77.52$\pm$0.31&68.60$\pm$0.04&50.16$\pm$0.08\\

&&NCE+RCE& 88.80$\pm$0.36&  88.75$\pm$0.49& 88.50$\pm$0.48&  87.50$\pm$0.36& 86.58$\pm$0.36 &88.80$\pm$0.48&  87.88$\pm$0.64& 70.95$\pm$0.95\\

&&NCE+AGCE & 88.93$\pm$0.18&  88.89$\pm$0.55&  88.19$\pm$0.91&  87.52$\pm$0.46& 86.97$\pm$0.92 &89.45$\pm$0.48&  89.06$\pm$0.61& 83.85$\pm$0.20\\

&&ANL-CE &88.93$\pm$0.18&  88.15$\pm$0.37&  88.28$\pm$0.39&  87.23$\pm$0.20& 75.64$\pm$1.02 &88.54$\pm$0.49&  88.41$\pm$0.18& 87.91$\pm$0.20\\

\hline

\hline
\end{tabular}
\label{tab:vitl_lp_res}
\end{table*}

\begin{table*}[t!]
\centering
\caption{\textbf{Impact of explicit entropy minimization on ViT performance with noisy labels:} Detailed benchmarking of ViT-B/16 with linear probing (LP) and MLP-3 fine-tuning on the MNIST dataset in terms of test accuracy. Performance improvements due to explicit entropy minimization are reported in \textcolor{blue}{blue}.} 
\renewcommand{\arraystretch}{0.94}
\begin{tabular}{cl|l|llll|lll}
\hline

 \hline
& \multirow{2}{4em}{Method} & \multirow{2}{4em}{Clean} & \multicolumn{4}{c|}{Symm Noise Rate ($\eta$)} & \multicolumn{3}{c}{Asym Noise Rate ($\eta$)}\\
\cline{4-10}
&&& 0.2 & 0.4  & 0.6 & 0.8 & 0.2 & 0.3 & 0.4\\
\hline

 \hline
 \parbox[t]{0.01cm}{\multirow{16}{*}{\rotatebox[origin=c]{90}{LINEAR PROBING}}} &CE &  96.48	&96.09	&95.7	&94.92	&87.89	&94.14	&90.23	&85.54\\
 
&CE+$H_l$ &97.27 \textcolor{blue}{\tiny{($\uparrow$0.79)}} &97.26 \textcolor{blue}{\tiny{($\uparrow$1.17)}}	&96.09 \textcolor{blue}{\tiny{($\uparrow$0.39)}}	&95.70 \textcolor{blue}{\tiny{($\uparrow$0.78)}}	&89.45 \textcolor{blue}{\tiny{($\uparrow$1.56)}}	&95.31 \textcolor{blue}{\tiny{($\uparrow$1.17)}}	&92.97 \textcolor{blue}{\tiny{($\uparrow$2.74)}}	&88.28 \textcolor{blue}{\tiny{($\uparrow$2.74)}}\\
\cline{2-10}

&MAE & 85.15	&76.17	&75.39	&66.79	&53.9	&63.73	&57.62	&57.32\\
&MAE+$H_l$ & 85.55 \textcolor{blue}{\tiny{($\uparrow$0.40)}}&76.50 \textcolor{blue}{\tiny{($\uparrow$0.33)}}	&75.78 \textcolor{blue}{\tiny{($\uparrow$0.39)}} &67.08 \textcolor{blue}{\tiny{($\uparrow$0.29)}}	&64.06 \textcolor{blue}{\tiny{($\uparrow$10.16)}}	&69.14 	\textcolor{blue}{\tiny{($\uparrow$5.41)}} &66.79 \textcolor{blue}{\tiny{($\uparrow$9.17)}}&65.20 \textcolor{blue}{\tiny{($\uparrow$7.88)}} \\

\cline{2-10}
&FL &  96.09	&95.7	&94.97	&91.66	&87.5	&94.53	&91.41	&86.32\\
&FL+$H_l$ & 96.48 \textcolor{blue}{\tiny{($\uparrow$0.39)}}	&96.09 \textcolor{blue}{\tiny{($\uparrow$0.39)}}	&95.70 \textcolor{blue}{\tiny{($\uparrow$0.73)}}	&95.70 \textcolor{blue}{\tiny{($\uparrow$1.04)}}	&90.23 \textcolor{blue}{\tiny{($\uparrow$2.73)}}	&94.92 \textcolor{blue}{\tiny{($\uparrow$0.39)}}	&92.97 \textcolor{blue}{\tiny{($\uparrow$1.56)}}	&87.89 \textcolor{blue}{\tiny{($\uparrow$1.57)}}\\

\cline{2-10}
&GCE &  95.31	&94.92	&93.75	&91.41	&87.11	&94.53	&93.35	&89.84\\
&GCE+$H_l$ & 95.65 \textcolor{blue}{\tiny{($\uparrow$0.34)}} &95.09 \textcolor{blue}{\tiny{($\uparrow$0.17)}} &94.53 \textcolor{blue}{\tiny{($\uparrow$0.78)}} &91.82 \textcolor{blue}{\tiny{($\uparrow$0.41)}}	&87.50 \textcolor{blue}{\tiny{($\uparrow$0.39)}}	 &94.92  \textcolor{blue}{\tiny{($\uparrow$0.39)}}&93.75 \textcolor{blue}{\tiny{($\uparrow$0.40)}}&91.01 \textcolor{blue}{\tiny{($\uparrow$1.17)}}\\

\cline{2-10}
&SCE & 96.48	&96.09	&96.09	&95.31	&91.79	&95.7	&87.89	&79.29\\
&SCE+$H_l$ & 96.87  \textcolor{blue}{\tiny{($\uparrow$0.39)}}&96.48\textcolor{blue}{\tiny{($\uparrow$0.39)}}&96.48 \textcolor{blue}{\tiny{($\uparrow$0.39)}}&96.09 \textcolor{blue}{\tiny{($\uparrow$0.78)}}	 &93.57 \textcolor{blue}{\tiny{($\uparrow$1.78)}}	&96.48 \textcolor{blue}{\tiny{($\uparrow$0.78)}}	&95.31 \textcolor{blue}{\tiny{($\uparrow$7.42)}} &79.30 \textcolor{blue}{\tiny{($\uparrow$0.01)}}\\ 
\cline{2-10}

&NCE+RCE & 97.26	&96.48	&96.48	&96.09	&89.06	&96.87	&88.67	&78.52\\
&NCE+RCE+$H_l$ &99.05 \textcolor{blue}{\tiny{($\uparrow$0.79)}} & 97.65 \textcolor{blue}{\tiny{($\uparrow$1.17)}} & 97.27 \textcolor{blue}{\tiny{($\uparrow$0.79)}} 	&97.27 \textcolor{blue}{\tiny{($\uparrow$1.18)}} &93.36 \textcolor{blue}{\tiny{($\uparrow$4.30)}}	&97.27 \textcolor{blue}{\tiny{($\uparrow$0.40)}}	&89.06 \textcolor{blue}{\tiny{($\uparrow$0.39)}}	&78.90 \textcolor{blue}{\tiny{($\uparrow$0.38)}}\\
\cline{2-10}

&NCE+AGCE & 82.81	&75.39	&74.22	&72.26	&56.64	&67.96	&59.38	&56.64\\
&NCE+AGCE+$H_l$ & 83.42 \textcolor{blue}{\tiny{($\uparrow$0.61)}} &82.78 \textcolor{blue}{\tiny{($\uparrow$7.39)}}	&81.25 
\textcolor{blue}{\tiny{($\uparrow$7.03)}}&73.43 
\textcolor{blue}{\tiny{($\uparrow$1.17)}} &59.37 
\textcolor{blue}{\tiny{($\uparrow$2.73)}}	&68.36 
\textcolor{blue}{\tiny{($\uparrow$0.40)}}	&64.06 
\textcolor{blue}{\tiny{($\uparrow$4.68)}} 	&58.40 
\textcolor{blue}{\tiny{($\uparrow$1.76)}} \\

\cline{2-10}
&ANL-CE &87.34	&82.34	&68.75	&61.58	&50.64	&67.96	&53.9	&45.7\\
&ANL-CE+$H_l$ &88.51 \textcolor{blue}{\tiny{($\uparrow$1.17)}} &83.41 \textcolor{blue}{\tiny{($\uparrow$1.07)}}	 &71.09 
\textcolor{blue}{\tiny{($\uparrow$2.34)}}	&70.31	
\textcolor{blue}{\tiny{($\uparrow$8.73)}}&56.4	
\textcolor{blue}{\tiny{($\uparrow$5.76)}}&74.6	
\textcolor{blue}{\tiny{($\uparrow$6.64)}}&62.89 
\textcolor{blue}{\tiny{($\uparrow$8.99)}}	&46.09 
\textcolor{blue}{\tiny{($\uparrow$0.39)}}\\
\hline

\hline
\parbox[t]{0.01cm}{\multirow{16}{*}{\rotatebox[origin=c]{90}{MLP-3}}}&CE & 96.09 & 95.33 & 94.79  & 94.13 & 85.41 & 96.01 & 95.44 & 92.49 \\
&CE+$H_l$ & 99.22 \textcolor{blue}{\tiny{($\uparrow$0.39)}} & 98.05 \textcolor{blue}{\tiny{($\uparrow$0.40)}}  & 98.05 \textcolor{blue}{\tiny{($\uparrow$0.79)}} & 96.48 \textcolor{blue}{\tiny{($\uparrow$1.95)}}& 92.08 \textcolor{blue}{\tiny{($\uparrow$0.28)}} & 98.04 \textcolor{blue}{\tiny{($\uparrow$0.39)}}& 97.65 \textcolor{blue}{\tiny{($\uparrow$0.80)}} & 95.70 \textcolor{blue}{\tiny{($\uparrow$0.39)}} \\

\cline{2-10}
&MAE & 78.12	&77.92	&76.04		&68.09	&26.8	&67.57	&59.01	&57.03\\
&MAE+$H_l$ & 87.50 \textcolor{blue}{\tiny{($\uparrow$0.39)}}&79.68 \textcolor{blue}{\tiny{($\uparrow$1.76)}} &78.51 \textcolor{blue}{\tiny{($\uparrow$2.47)}} &76.95 \textcolor{blue}{\tiny{($\uparrow$8.86)}} &29.29 \textcolor{blue}{\tiny{($\uparrow$2.49)}} &67.96	\textcolor{blue}{\tiny{($\uparrow$0.39)}} &67.07	\textcolor{blue}{\tiny{($\uparrow$8.06)}} &57.81 \textcolor{blue}{\tiny{($\uparrow$0.78)}}\\

\cline{2-10}
&FL & 96.61 & 94.92 & 94.79 & 93.74 & 84.23 & 96.04 & 94.82 & 91.72 \\
&FL+$H_l$ & 98.44 \textcolor{blue}{\tiny{($\uparrow$0.79)}}& 98.05 \textcolor{blue}{\tiny{($\uparrow$0.39)}}& 96.48 \textcolor{blue}{\tiny{($\uparrow$0.39)}}& 96.48 \textcolor{blue}{\tiny{($\uparrow$1.39)}}& 89.28  \textcolor{blue}{\tiny{($\uparrow$0.61)}}& 98.05 \textcolor{blue}{\tiny{($\uparrow$0.40)}} & 97.26 \textcolor{blue}{\tiny{($\uparrow$1.56)}} & 95.31 \textcolor{blue}{\tiny{($\uparrow$0.39)}} \\

\cline{2-10}
&GCE & 95.31 & 94.92 & 94.14  & 92.18 & 46.74 & 94.79 & 94.4 & 90.52 \\
&GCE+$H_l$& 97.65  \textcolor{blue}{\tiny{($\uparrow$0.38)}}& 97.27 \textcolor{blue}{\tiny{($\uparrow$0.40)}}& 97.09 \textcolor{blue}{\tiny{($\uparrow$0.22)}} & 96.09 \textcolor{blue}{\tiny{($\uparrow$1.17)}} & 66.79 \textcolor{blue}{\tiny{($\uparrow$2.73)}} & 96.87 \textcolor{blue}{\tiny{($\uparrow$0.78)}} & 96.09 \textcolor{blue}{\tiny{($\uparrow$0.78)}} & 94.92  \textcolor{blue}{\tiny{($\uparrow$3.13)}}\\

\cline{2-10}
&SCE & 96.48& 95.57 & 95.45  & 94.93 & 79.68 & 96.01 & 95.61 & 93.3 \\
&SCE+$H_l$ & 98.05 \textcolor{blue}{\tiny{($\uparrow$0.79)}}& 98.05 \textcolor{blue}{\tiny{($\uparrow$0.79)}} & 97.26 \textcolor{blue}{\tiny{($\uparrow$0.78)}}& 96.88  \textcolor{blue}{\tiny{($\uparrow$1.01)}}& 95.70 \textcolor{blue}{\tiny{($\uparrow$1.17)}}& 97.26 \textcolor{blue}{\tiny{($\uparrow$0.78)}}& 96.87 \textcolor{blue}{\tiny{($\uparrow$0.39)}} & 96.48  \textcolor{blue}{\tiny{($\uparrow$0.39)}}\\

\cline{2-10}
&NCE+RCE & 95.70 & 94.92 & 94.66 & 83.59 & 23.82 & 87.5 & 86.16 & 68.75 \\
&NCE+RCE+$H_l$ & 98.04 \textcolor{blue}{\tiny{($\uparrow$0.77)}}& 97.27 \textcolor{blue}{\tiny{($\uparrow$0.40)}}& 96.88 \textcolor{blue}{\tiny{($\uparrow$0.40)}}& 96.48 \textcolor{blue}{\tiny{($\uparrow$0.39)}} & 75.94 \textcolor{blue}{\tiny{($\uparrow$2.51)}} & 96.48 \textcolor{blue}{\tiny{($\uparrow$0.39)}}& 88.45 \textcolor{blue}{\tiny{($\uparrow$0.39)}}& 81.11 \textcolor{blue}{\tiny{($\uparrow$1.03)}}\\

\cline{2-10}
&NCE+AGCE & 73.34 & 64.84 & 45.7  & 22.65 & 14.06 & 51.17 & 49.6 & 48.82 \\
&NCE+AGCE+$H_l$ & 89.59 \textcolor{blue}{\tiny{($\uparrow$0.69)}}& 81.64 \textcolor{blue}{\tiny{($\uparrow$0.79)}}& 75.92 \textcolor{blue}{\tiny{($\uparrow$1.31)}}& 61.66 \textcolor{blue}{\tiny{($\uparrow$1.12)}} & 47.66  \textcolor{blue}{\tiny{($\uparrow$0.79)}}& 68.81 \textcolor{blue}{\tiny{($\uparrow$1.63)}}& 57.91 \textcolor{blue}{\tiny{($\uparrow$0.10)}} & 57.62  \textcolor{blue}{\tiny{($\uparrow$0.20)}}\\

\cline{2-10}
&ANL-CE & 87.5 & 82.03 & 76.95  & 49.6 & 36.32 & 81.25 & 79.29 & 64.84 \\

&ANL-CE +$H_l$ & 94.14 \textcolor{blue}{\tiny{($\uparrow$1.56)}}& 91.80 \textcolor{blue}{\tiny{($\uparrow$0.79)}} & 85.20 \textcolor{blue}{\tiny{($\uparrow$1.22)}}& 71.48 \textcolor{blue}{\tiny{($\uparrow$2.34)}} & 68.43 \textcolor{blue}{\tiny{($\uparrow$2.03)}}& 91.65 \textcolor{blue}{\tiny{($\uparrow$0.63)}}& 86.33 \textcolor{blue}{\tiny{($\uparrow$0.79)}}& 71.88  \textcolor{blue}{\tiny{($\uparrow$1.57)}}\\

\hline

\hline

\end{tabular}
\label{tab:mnist_vitb_ent_min}
\end{table*}

\section{Explicit Entropy Minimization Improves ViTs}
The average test accuracy for ViT-B/16 and ViT-L/16 using linear probing (LP) and MLP-3 fine-tuning across six datasets is presented in Table IV of the main paper. This performance was averaged across three Common Loss Functions (CLF) and six Noisy Label Learning (NLL) methods. For noisy datasets like MNIST and CIFAR-10/100, the results were averaged over symmetric noise levels \{0.2, 0.4, 0.6, 0.8\} and asymmetric noise levels \{0.2, 0.3, 0.4\}. Detailed results are available in Tables V and VI of the main paper, while additional results are provided in Tables XII through XVIII in this supplementary document. Specifically, Tables XII, XIV, and XVI present detailed results for ViT-B/16 using LP and MLP-3 fine-tuning on the MNIST and CIFAR-10/100 datasets, while Tables XIII, XV, and XVII provide the corresponding results for ViT-L/16. Table XVIII includes detailed results for explicit entropy minimization for both backbones on the WebVision, Clothing1M, and Food-101N datasets. Across all datasets, employing explicit entropy minimization consistently improved overall performance compared to baseline methods.

\begin{table*}
\centering
\caption{\textbf{Impact of explicit entropy minimization on ViT performance with noisy labels:} Detailed benchmarking of ViT-L/16 with linear probing (LP) and MLP-3 fine-tuning on the MNIST dataset in terms of test accuracy. Improvements due to the proposed explicit entropy minimization loss are highlighted in \textcolor{blue}{blue}.}
\renewcommand{\arraystretch}{0.94}
\begin{tabular}{cl|l|llll|lll}
\hline

 \hline
& \multirow{2}{4em}{Method} & \multirow{2}{4em}{Clean} & \multicolumn{4}{c|}{Symm Noise Rate ($\eta$)} & \multicolumn{3}{c}{Asym Noise Rate ($\eta$)}\\
\cline{4-10}
&&& 0.2 & 0.4  & 0.6 & 0.8 & 0.2 & 0.3 & 0.4\\
\hline

 \hline

 \parbox[t]{0.01cm}{\multirow{16}{*}{\rotatebox[origin=c]{90}{LINEAR PROBING}}}&CE&  98.04	&98.04	&96.48	&94.92	&83.2	&95.7	&93.35	&88.67\\
 &CE+$H_l$&  98.44 \textcolor{blue}{\tiny{($\uparrow$0.40)}}&  98.43 \textcolor{blue}{\tiny{($\uparrow$0.39)}}&  98.04  \textcolor{blue}{\tiny{($\uparrow$1.56)}}&  96.53 \textcolor{blue}{\tiny{($\uparrow$1.61)}}&  87.11 \textcolor{blue}{\tiny{($\uparrow$3.91)}}&  97.48 \textcolor{blue}{\tiny{($\uparrow$0.78)}}&  95.31 \textcolor{blue}{\tiny{($\uparrow$1.96)}}& 90.23 \textcolor{blue}{\tiny{($\uparrow$1.56)}}\\
 \cline{2-10}
 &MAE& 86.72	&86.14	&85.54	&83.98	&73.43	&65.62	&57.81	&57.42\\

 &MAE+$H_l$&  88.28 \textcolor{blue}{\tiny{($\uparrow$1.56)}}&  87.89 \textcolor{blue}{\tiny{($\uparrow$1.75)}}&  87.11 \textcolor{blue}{\tiny{($\uparrow$1.57)}}&  84.15  \textcolor{blue}{\tiny{($\uparrow$0.17)}}&  75.66 \textcolor{blue}{\tiny{($\uparrow$2.23)}}&  87.50 \textcolor{blue}{\tiny{($\uparrow$21.88)}}&  67.57 \textcolor{blue}{\tiny{($\uparrow$9.76)}}& 58.59 \textcolor{blue}{\tiny{($\uparrow$1.17)}}\\
 
\cline{2-10}
 &FL&  97.26	&96.09	&96.87	&96.48	&86.32	&94.53	&91.41	&84.37\\
 
 &FL+$H_l$&  98.43 \textcolor{blue}{\tiny{($\uparrow$1.17)}}&  98.05 \textcolor{blue}{\tiny{($\uparrow$1.96)}}&  97.27 \textcolor{blue}{\tiny{($\uparrow$0.40)}}&  96.87 \textcolor{blue}{\tiny{($\uparrow$0.39)}}&  89.45 \textcolor{blue}{\tiny{($\uparrow$3.13)}}&  98.04 \textcolor{blue}{\tiny{($\uparrow$3.51)}}&  96.09 \textcolor{blue}{\tiny{($\uparrow$4.68)}}& 95.70 \textcolor{blue}{\tiny{($\uparrow$11.33)}}\\
\hline

\hline
\end{tabular}
\label{tab:mnist_vitl_ent_min}
\end{table*}

\begin{table*}
\centering
\caption{\textbf{Impact of explicit entropy minimization on ViT performance with noisy labels:} Detailed benchmarking of ViT-B/16 with linear probing (LP) and MLP-3 fine-tuning on the CIFAR-10 dataset in terms of test accuracy. Performance improvements due to explicit entropy minimization are highlighted in \textcolor{blue}{blue}.} 

\renewcommand{\arraystretch}{0.94}
\begin{tabular}{cl|l|llll|lll}
\hline

 \hline
& \multirow{2}{4em}{Method} & \multirow{2}{4em}{Clean} & \multicolumn{4}{c|}{Symm Noise Rate ($\eta$)} & \multicolumn{3}{c}{Asym Noise Rate ($\eta$)}\\
\cline{4-10}
&&& 0.2 & 0.4  & 0.6 & 0.8 & 0.2 & 0.3 & 0.4\\
\hline

 \hline
 \parbox[t]{0.01cm}{\multirow{16}{*}{\rotatebox[origin=c]{90}{LINEAR PROBING}}}&CE &  96.55	&95.89	&95.08	&92.21	&68.86	&90.13	&86.04	&80.35\\

&CE+$H_l$ &97.26 \textcolor{blue}{\tiny{($\uparrow$0.71)}}&96.48 \textcolor{blue}{\tiny{($\uparrow$0.59)}} &96.35 \textcolor{blue}{\tiny{($\uparrow$1.27)}} &95.18 \textcolor{blue}{\tiny{($\uparrow$2.97)}} 	&87.36 \textcolor{blue}{\tiny{($\uparrow$18.5)}}	&96.35 \textcolor{blue}{\tiny{($\uparrow$6.22)}} &94.66 \textcolor{blue}{\tiny{($\uparrow$8.62)}} &91.27 \textcolor{blue}{\tiny{($\uparrow$10.92)}} \\

\cline{2-10}
&MAE &95.83	&92.26	&91.92	&86.06	&66.79	&92.96	&87.37&	79.94\\
&MAE+$H_l$ & 96.48 \textcolor{blue}{\tiny{($\uparrow$0.65)}} &96.09 \textcolor{blue}{\tiny{($\uparrow$3.83)}}&95.7 \textcolor{blue}{\tiny{($\uparrow$3.78)}}&	94.92 \textcolor{blue}{\tiny{($\uparrow$8.86)}} &67.57 \textcolor{blue}{\tiny{($\uparrow$0.78)}} &94.56 \textcolor{blue}{\tiny{($\uparrow$1.60)}}&89.96 \textcolor{blue}{\tiny{($\uparrow$2.59)}} &81.76 \textcolor{blue}{\tiny{($\uparrow$1.82)}}\\

\cline{2-10}
&FL &  96.36	&95.87	&94.97	&92.44	&68.15	&91.54	&84.76	&77.18\\
&FL+$H_l$ & 96.87 \textcolor{blue}{\tiny{($\uparrow$0.52)}}&96.87 \textcolor{blue}{\tiny{($\uparrow$1.00)}}  &95.31 \textcolor{blue}{\tiny{($\uparrow$0.34)}}  &95.31 \textcolor{blue}{\tiny{($\uparrow$2.87)}} &88.67 \textcolor{blue}{\tiny{($\uparrow$20.52)}}  &95.31 \textcolor{blue}{\tiny{($\uparrow$3.77)}}   &94.53 \textcolor{blue}{\tiny{($\uparrow$9.77)}} 	&87.89\textcolor{blue}{\tiny{($\uparrow$10.71)}} \\
\hline

\hline
\end{tabular}
\label{tab:cifar10_vitb_ent_min}
\end{table*}

\begin{table*}
\centering
\caption{\textbf{Impact of explicit entropy minimization on ViT performance with noisy labels:} Detailed benchmarking of ViT-L/16 with linear probing (LP) and MLP-3 fine-tuning on the CIFAR-10 dataset in terms of test accuracy. Performance improvements due to explicit entropy minimization are highlighted in \textcolor{blue}{blue}.} 
\renewcommand{\arraystretch}{0.94}
\begin{tabular}{cl|l|llll|lll}
\hline

 \hline
& \multirow{2}{4em}{Method} & \multirow{2}{4em}{Clean} & \multicolumn{4}{c|}{Symm Noise Rate ($\eta$)} & \multicolumn{3}{c}{Asym Noise Rate ($\eta$)}\\
\cline{4-10}
&&& 0.2 & 0.4  & 0.6 & 0.8 & 0.2 & 0.3 & 0.4\\
\hline

 \hline
 \parbox[t]{0.01cm}{\multirow{16}{*}{\rotatebox[origin=c]{90}{LINEAR PROBING}}}&CE&  96.87&  95.7&  94.14&  89.84&  58.59&  93.75&  92.18& 83.98\\
 &CE+$H_l$&  96.88 \textcolor{blue}{\tiny{($\uparrow$0.01)}}&  96.48 \textcolor{blue}{\tiny{($\uparrow$0.78)}}&  94.14 \textcolor{blue}{\tiny{($\uparrow$0.00)}}&  91.01 \textcolor{blue}{\tiny{($\uparrow$1.17)}}&  62.11 \textcolor{blue}{\tiny{($\uparrow$3.52)}}&  96.48  \textcolor{blue}{\tiny{($\uparrow$2.73)}}&  94.92 \textcolor{blue}{\tiny{($\uparrow$2.74)}}& 89.45 \textcolor{blue}{\tiny{($\uparrow$5.47)}}\\
 \cline{2-10}
 
 &MAE&  96.82&  96.46&  96.09&  95.7&  95.31&  83.98&  75& 62.33\\
 &MAE+$H_l$&  96.87 \textcolor{blue}{\tiny{($\uparrow$0.05)}}&  96.48 \textcolor{blue}{\tiny{($\uparrow$0.02)}}&  96.48 \textcolor{blue}{\tiny{($\uparrow$0.39)}}&  96.09 \textcolor{blue}{\tiny{($\uparrow$0.39)}}&  95.7  \textcolor{blue}{\tiny{($\uparrow$0.39)}}&  85.15  \textcolor{blue}{\tiny{($\uparrow$1.17)}}&  75.78  \textcolor{blue}{\tiny{($\uparrow$0.78)}}& 67.18 \textcolor{blue}{\tiny{($\uparrow$4.85)}}\\
 \cline{2-10}
 &FL&  94.92&  94.53&  93.35&  89.84&  60.89&  94.53&  91.4& 82.81\\
 &FL+$H_l$&  96.87 \textcolor{blue}{\tiny{($\uparrow$1.95)}}&  96.09 \textcolor{blue}{\tiny{($\uparrow$1.56)}}&  95.7 \textcolor{blue}{\tiny{($\uparrow$2.35)}}&  95.31 \textcolor{blue}{\tiny{($\uparrow$5.47)}}&  76.95  \textcolor{blue}{\tiny{($\uparrow$16.06)}}&  96.09 \textcolor{blue}{\tiny{($\uparrow$1.56)}}&  96.09 \textcolor{blue}{\tiny{($\uparrow$4.69)}}& 85.93 \textcolor{blue}{\tiny{($\uparrow$3.12)}}\\
\hline

\hline
\end{tabular}
\label{tab:cifar10_vitl_ent_min}
\end{table*}

\begin{table*}
\centering
\caption{
\textbf{Impact of explicit entropy minimization on ViT performance with noisy labels:} Detailed benchmarking of ViT-B/16 with linear probing (LP) and MLP-3 fine-tuning on the CIFAR-100 dataset in terms of test accuracy. Performance improvements due to the proposed explicit entropy minimization loss are highlighted in \textcolor{blue}{blue}.
} 
\renewcommand{\arraystretch}{0.94}
\begin{tabular}{cl|l|llll|lll}
\hline

 \hline
& \multirow{2}{4em}{Method} & \multirow{2}{4em}{Clean} & \multicolumn{4}{c|}{Symm Noise Rate ($\eta$)} & \multicolumn{3}{c}{Asym Noise Rate ($\eta$)}\\
\cline{4-10}
&&& 0.2 & 0.4  & 0.6 & 0.8 & 0.2 & 0.3 & 0.4\\
\hline

 \hline

 \parbox[t]{0.01cm}{\multirow{16}{*}{\rotatebox[origin=c]{90}{LINEAR PROBING}}}&CE &  86.12	&69.91	&68.35	&58.07	&48.51	&65.07	&60.48	&52.1\\
&CE+$H_l$ &86.32\textcolor{blue}{\tiny{($\uparrow$0.2)}} &84.24\textcolor{blue}{\tiny{($\uparrow$14.33)}}	&83.59 \textcolor{blue}{\tiny{($\uparrow$15.24)}} 	&82.68 \textcolor{blue}{\tiny{($\uparrow$24.61)}}	&75.39 \textcolor{blue}{\tiny{($\uparrow$26.88)}}&79.29 \textcolor{blue}{\tiny{($\uparrow$14.22)}} &70.95 \textcolor{blue}{\tiny{($\uparrow$10.47)}}	 &56.82 \textcolor{blue}{\tiny{($\uparrow$4.72)}}\\

\cline{2-10}
&MAE & 62.49	&60.28	&58.06	&56.24	&47.52	&48.43	&44.72	&33.75\\
&MAE+$H_l$ & 63.67 \textcolor{blue}{\tiny{($\uparrow$1.18)}}& 63.67 \textcolor{blue}{\tiny{($\uparrow$3.39)}} &59.76 \textcolor{blue}{\tiny{($\uparrow$1.70)}} &58.2 \textcolor{blue}{\tiny{($\uparrow$1.96)}}	&48.82 \textcolor{blue}{\tiny{($\uparrow$1.30)}} &57.42 \textcolor{blue}{\tiny{($\uparrow$8.99)}}	&48.04 \textcolor{blue}{\tiny{($\uparrow$3.32)}}	& 34.37 \textcolor{blue}{\tiny{($\uparrow$0.62)}}\\

\cline{2-10}
&FL &  83.33	&81.63	&80.73	&75.12	&54.03	&63.54	&55.59	&49.47\\
&FL+$H_l$ & 85.93 \textcolor{blue}{\tiny{($\uparrow$2.6)}} &83.98 	\textcolor{blue}{\tiny{($\uparrow$2.35)}}&83.59 \textcolor{blue}{\tiny{($\uparrow$2.86)}}&80.46 \textcolor{blue}{\tiny{($\uparrow$5.34)}} &77.73 \textcolor{blue}{\tiny{($\uparrow$23.7)}}&75.39 \textcolor{blue}{\tiny{($\uparrow$11.85)}}	&67.18\textcolor{blue}{\tiny{($\uparrow$11.59)}} &52.34 \textcolor{blue}{\tiny{($\uparrow$2.87)}} \\
\hline

\hline
\end{tabular}
\label{tab:cifar100_vitb_ent_min}
\end{table*}

\begin{table*}
\centering
\caption{
\textbf{Impact of explicit entropy minimization on ViT performance with noisy labels:} Detailed benchmarking of ViT-L/16 with linear probing (LP) and MLP-3 fine-tuning on the CIFAR-100 dataset in terms of test accuracy. Improvements due to the proposed explicit entropy minimization loss are highlighted in \textcolor{blue}{blue}.}
\label{tab:cifar100_vitl_ent_min}
\renewcommand{\arraystretch}{0.94}
\begin{tabular}{cl|l|llll|lll}
\hline

 \hline
& \multirow{2}{4em}{Method} & \multirow{2}{4em}{Clean} & \multicolumn{4}{c|}{Symm Noise Rate ($\eta$)} & \multicolumn{3}{c}{Asym Noise Rate ($\eta$)}\\
\cline{4-10}
&&& 0.2 & 0.4  & 0.6 & 0.8 & 0.2 & 0.3 & 0.4\\
\hline

 \hline
 \parbox[t]{0.01cm}{\multirow{16}{*}{\rotatebox[origin=c]{90}{LINEAR PROBING}}}&CE&  85.8&  70.56&  64.71&  58.71&  40.23&  67.44&  59.24& 53.12\\
 &CE+$H_l$&  89.84 \textcolor{blue}{\tiny{($\uparrow$4.04)}}&  85.93 \textcolor{blue}{\tiny{($\uparrow$15.37)}}&  79.68 \textcolor{blue}{\tiny{($\uparrow$14.97)}}&  70.7 \textcolor{blue}{\tiny{($\uparrow$11.99)}}&  57.42 \textcolor{blue}{\tiny{($\uparrow$17.19)}}&  77.34 \textcolor{blue}{\tiny{($\uparrow$9.9)}}&  69.92 \textcolor{blue}{\tiny{($\uparrow$10.68)}}& 59.76  \textcolor{blue}{\tiny{($\uparrow$6.64)}}\\
 \cline{2-10}
 &MAE&  70.13&  69.53&  64.85&  64.45&  53.12&  53.9&  46.48& 45.31\\
 &MAE+$H_l$&  73.82 \textcolor{blue}{\tiny{($\uparrow$3.69)}}&  73.02 \textcolor{blue}{\tiny{($\uparrow$3.49)}}&  68.35 \textcolor{blue}{\tiny{($\uparrow$3.5)}}&  65.23 \textcolor{blue}{\tiny{($\uparrow$0.78)}}&  58.98 \textcolor{blue}{\tiny{($\uparrow$5.86)}}&  54.68  \textcolor{blue}{\tiny{($\uparrow$0.78)}}&  47.65 \textcolor{blue}{\tiny{($\uparrow$1.17)}}& 44.14 \textcolor{red}{\tiny{($\downarrow$1.17)}}\\
 \cline{2-10}
 &FL&  86.45&  69.01&  62.75&  57.02&  39.05&  66.66&  57.93& 52.47\\
 &FL+$H_l$&  89.06 \textcolor{blue}{\tiny{($\uparrow$2.61)}}&  85.15 \textcolor{blue}{\tiny{($\uparrow$16.14)}}&  79.29 \textcolor{blue}{\tiny{($\uparrow$16.54)}}&  75.39 \textcolor{blue}{\tiny{($\uparrow$18.37)}}&  58.59 \textcolor{blue}{\tiny{($\uparrow$19.54)}}&  73.82  \textcolor{blue}{\tiny{($\uparrow$7.16)}}&  66.01 \textcolor{blue}{\tiny{($\uparrow$8.08)}}& 55.46 \textcolor{blue}{\tiny{($\uparrow$2.99)}}\\
\hline

 \hline
    \end{tabular}

\end{table*}

\begin{table*}[t!]
\centering
\caption{
\textbf{Impact of explicit entropy minimization on ViT performance with noisy labels:} Detailed benchmarking of ViT-B/16+LP, ViT-B/16+MLP-3, ViT-L/16+LP, and ViT-L/16+MLP-3 fine-tuning on the WebVision, Clothing1M, and Food-101N datasets in terms of test accuracy. Improvements using the proposed explicit entropy loss are highlighted in \textcolor{blue}{blue}.} 

\begin{tabular}{ll|ll|ll}
\hline

\hline
  &\multirow{2}{4em}{Method}& \multicolumn{2}{c|}{ViT-B/16}& \multicolumn{2}{c}{ViT-L/16} \\
\cline{3-6}
 &&  LP&  MLP-3&  LP& MLP-3\\
 \hline

 \hline
 \parbox[t]{0.01cm}{\multirow{12}{*}{\rotatebox[origin=c]{90}{WebVision}}}&CE&  87.79&  88.47&  86.71& 86.81\\
 &CE+$H_l$&  88.18 \textcolor{blue}{\tiny{($\uparrow$0.39)}}&  89.35 \textcolor{blue}{\tiny{($\uparrow$0.88)}}&  89.16 \textcolor{blue}{\tiny{($\uparrow$2.45)}}& 89.74 \textcolor{blue}{\tiny{($\uparrow$2.93)}}\\
\cline{2-6}
 &GCE&  89.16&  76.75&  90.13& 84.76\\
 &GCE+$H_l$&  89.84 \textcolor{blue}{\tiny{($\uparrow$0.68)}}&  81.73  \textcolor{blue}{\tiny{($\uparrow$4.98)}}&  91.3 \textcolor{blue}{\tiny{($\uparrow$1.17)}}& 85.15 \textcolor{blue}{\tiny{($\uparrow$0.39)}}\\
\cline{2-6}
 &SCE&  86.03&  87.4&  84.86& 88.18\\
 &SCE+$H_l$&  86.71 \textcolor{blue}{\tiny{($\uparrow$0.68)}}&  88.96  \textcolor{blue}{\tiny{($\uparrow$1.56)}}&  87.89 \textcolor{blue}{\tiny{($\uparrow$3.03)}}& 90.91 \textcolor{blue}{\tiny{($\uparrow$2.73)}}\\
\hline

\end{tabular}

\label{tab:real_noisy_ent_min}
\end{table*}

\begin{table*}[t!]
\caption{\textbf{Explicit entropy minimization improves CNN performance with noisy labels:} Test accuracy (\%) of CNN models using various loss functions on the MNIST, CIFAR-10, and CIFAR-100 datasets. The best results are highlighted in \textbf{bold}, and the second-best results are \underline{underlined}.} 
\centering 
 \begin{tabular}{ll|c|p{1cm}p{1cm}p{1cm}p{1cm}|p{1cm}p{1cm}p{1cm}}
    \hline

    \hline
        &\multirow{2}{4em}{Method}&  \multirow{2}{4em}{Clean ($\eta$=0.0)}&  \multicolumn{4}{c}{Sym Noise Rate ($\eta$)} & \multicolumn{3}{|c}{Asym Noise Rate ($\eta$)}\\
         \cline{3-10}
         &&  &  0.2&  0.4&  0.6&  0.8 & 0.2 & 0.3 & 0.4\\
    \hline

    \hline
         \parbox[t]{0.01cm}{\multirow{11}{*}{\rotatebox[origin=c]{90}{MNIST}}}&CE&  99.20 & 91.40 & 74.46 &49.19 &22.51 & 94.02&88.90&81.79\\
         &MAE&  99.16 &\underline{99.03} &98.80 &97.69 &70.35 &\textbf{99.11}&98.42&87.40\\
         &FL& 99.16 &91.66&75.42 &50.58 &22.93&94.02 &88.90 &81.79 \\
         \cline{2-10}
         &GCE & 99.18 &98.84 &96.81 &80.86 &33.59 &96.59 &88.99 &81.91\\
         &SCE &  \underline{99.30} &98.91 &97.48 &88.35 &48.28 &97.95 &94.00 &84.54\\
         &NLNL& 98.61 &98.02 &97.17 &95.42 &86.34 &98.35 &97.51 &95.84\\
\hline
         
         \hline
    \end{tabular}
\label{tab:CNN_result}
\end{table*}

\section{Explicit Entropy Minimization Improves CNN}\label{A_subsec:extended_CNN_res}
In Table XIX, detailed results and comparisons for CNN models are presented. Across all noise levels, the proposed explicit entropy minimization loss consistently led to performance improvements for CNNs.

\begin{table}[t!]
    \centering
    \caption{\textbf{Detailed comparison of implicit entropy reduction ($\Delta H$) between the \nth{1} and last training epochs, alongside \% test accuracy (Acc.) for six datasets.} Common loss functions (CLF) and NLL methods are evaluated with a 0.60 symmetric noise rate.}
    \begin{tabular}{ccc|cc|cc|cc|cc|cc|cc}
    \hline
    
    \hline 
         & &\multirow{2}{4em}{Method}&  \multicolumn{2}{c|}{MNIST}&   \multicolumn{2}{c|}{CIFAR-10}&  \multicolumn{2}{c|}{CIFAR-100}&   \multicolumn{2}{c|}{WebVision}&   \multicolumn{2}{c|}{Clothing1M}&  \multicolumn{2}{c}{Food-101N}\\
\cline{4-15}
 & && $\Delta H$& Acc.& $\Delta H$& Acc.& $\Delta H$& Acc.& $\Delta H$& Acc.& $\Delta H$& Acc.& $\Delta H$&Acc.\\
 \hline
 
 \hline
 \parbox[t]{0.1cm}{\multirow{8}{*}{\rotatebox[origin=c]{90}{Vit-B/16 + LP}}}&\multirow{3}{*}{CLF}&CE& 0.174& 94.92& 0.419& 92.21& 0.409& 58.07& 0.1202& 87.79& 0.018& 63.96& 0.46&75.09\\
 &&MAE&  0.48& 66.79& 0.371& 86.06& 0.478& 56.24& -&- &- & -& -&-\\
 &&FL&  0.382& 91.66& 0.425& 92.44& 0.548& 75.12& -&- &- & -& -&-\\
 \cline{2-15}
 &\multirow{5}{*}{NLL}&GCE&  0.41& 91.41& 0.901& 95.55& 0.952& 82.94& 0.298& 89.16& 0.026& 62.4& 0.25&76.6\\
 &&SCE&  0.186& 95.31& 0.482& 95.24& 0.452& 58.72& 0.271& 86.03& 0.028& 63.37& 0.21&74.02\\
 &&NCE+RCE&  0.068& 96.09& 0.462& 95.18& 0.963& 84.24& 0.205& 88.67& 0.016& 62.3& 0.26&76.17\\
 &&NCE+AGCE&  0.49& 72.26& 0.913& 95.79& 0.967& 84.76& 0.212& 89.25& 0.01& 62.5& 0.26&76.26\\
 &&ANL-CE&  0.784& 61.58& 0.891& 95.05& 0.823& 67.57& 0.468& 88.96& 0.04& 62.79& 0.654&69.92\\
 \hline
 
 \hline

 \parbox[t]{0.1cm}{\multirow{8}{*}{\rotatebox[origin=c]{90}{Vit-B/16 + MLP-3}}}&\multirow{3}{*}{CLF}&CE&    0.082& 94.53& 0.153& 66.66& 0.34& 41.66& 0.345& 88.47& 0.204& 64.64& 0.42&74.31\\
 &&MAE&    0.85& 68.09& 0.143& 75.82& 0.319& 33.06& -&- &- & -& -&-\\
 &&FL&    0.404& 95.09& 0.512& 70.81& 0.402& 42.44& -&- &- & -& -&-\\
 \cline{2-15}
 &\multirow{5}{*}{NLL}&GCE&    0.48& 94.92& 0.903& 95.63& 0.945& 79.29& 0.226& 76.75& 0.419& 65.42& 0.412&72.16\\
 &&SCE&    0.143& 95.87& 0.412& 89.58& 0.401& 47.26& 0.1327& 87.4& 0.103& 62.21& 0.182&72.94\\
 &&NCE+RCE&    0.068& 96.09& 0.456& 95.12& 0.951& 80.07& 0.198& 88.57& 0.015& 65.52& 0.21&74.9\\
 &&NCE+AGCE&    0.486& 60.54& 0.482& 94.53& 0.95& 81.37& 0.196& 89.35& 0.011& 64.64& 0.013&75.18\\
 &&ANL-CE&    0.802& 69.14& 0.985& 94.27& 0.988& 81.5& 0.286& 89.16& 0.018& 64.35& 0.528&72.55\\
 \hline

 \hline
 
 \parbox[t]{0.1cm}{\multirow{8}{*}{\rotatebox[origin=c]{90}{Vit-L/16 + LP}}}&\multirow{3}{*}{CLF}&CE&    0.174& 94.92& 0.39& 89.84& 0.412& 58.71& 0.101& 86.71& 0.079& 63.86& 0.521&81.05\\
 &&MAE&    0.86& 83.98& 0.456& 95.7& 0.568& 64.45& -&- &- & -& -&-\\
 &&FL&    0.421& 96.48& 0.392& 89.84& 0.493& 57.02& -&- &- & -& -&-\\
 \cline{2-15}
&\multirow{5}{*}{NLL}&GCE&    0.416& 94.14& 0.883& 94.92& 0.9666& 87.1& 0.512& 90.13& 0.004& 63.96& 0.534&81.73\\
 &&SCE&    0.19& 95.31& 0.457& 93.75& 0.437& 55.07& 0.248& 84.86& 0.182& 64.06& 0.377&81.15\\
 &&NCE+RCE&    0.068& 95.32& 0.448& 94.92& 0.966& 87.5& 0.301& 89.45& 0.082& 64.06& 0.316&80.85\\
 &&NCE+AGCE&    0.5& 80.46& 0.473& 94.14& 0.974& 87.52& 0.297& 89.74& 0.004& 63.96& 0.326&80.85\\
 &&ANL-CE&    0.81& 63.67& 0.99& 95.96& 0.988& 87.23& 0.556& 90.82& 0.021& 63.67& 0.39&78.02\\

 \hline

 \hline
 \parbox[t]{0.1cm}{\multirow{8}{*}{\rotatebox[origin=c]{90}{Vit-L/16 + MLP-3}}}&\multirow{3}{*}{CLF}&CE&   0.188& 96.48& 0.103& 57.23& 0.38& 51.94& 0.062& 86.81& 0.26& 65.03& 0.538&81.34\\
 &&MAE&  0.86& 85.93& 0.979& 94.14& 0.356& 36.32& -&- &- & -& -&-\\
 &&FL&      0.418& 94.53& 0.11& 58.48& 0.456& 49.34&-&- &- & -& -&-\\
 \cline{2-15}
 &\multirow{5}{*}{NLL}&GCE& 0.445& 96.87& 0.872& 94.53& 0.958& 85.54& 0.327& 84.76& 0.434&65.62& 0.495&80.07\\
 &&SCE&      0.187& 94.92& 0.413& 84.37& 0.415& 52.33& 0.324& 88.18& 0.195& 64.94& 0.21&76.46\\
 &&NCE+RCE&      0.064& 96.09& 0.404& 93.75& 0.965& 85.93& 0.122& 88.57& 0.092& 65.42& 0.31&80.85\\
 &&NCE+AGCE&      0.608& 81.64& 0.418& 93.75& 0.969& 86.06& 0.213& 88.37& 0.08& 64.84& 0.35&81.15\\
 &&ANL-CE&      0.85& 86.72& 0.985& 95.05& 0.982& 85.15& 0.4& 89.06& 0.04& 64.94& 0.402&80.17\\
 \hline

 \hline
    \end{tabular}
    
    \label{tab:detailed_entropy_min}
\end{table}
\newpage

 

% \end{IEEEbiography}



\end{document}